\documentclass[11pt]{article}

\usepackage[preprint]{acl}

\usepackage{times}
\usepackage{latexsym}

\usepackage[T1]{fontenc}

\usepackage[utf8]{inputenc}

\usepackage{microtype}

\usepackage{inconsolata}

\usepackage{graphicx}
\usepackage{tikz}
\usepackage[edges]{forest}
\usepackage{xcolor}
\usepackage{listings}
\usepackage[size=tiny]{todonotes}
\usepackage{array}
\usepackage{enumitem}
\usepackage{booktabs}
\usepackage{multirow}
\usepackage{multicol}
\usepackage{tabularx}
\usepackage{xltabular} 
\usepackage{graphicx}
\usepackage{supertabular} 
\usepackage{amsmath}
\usepackage{subcaption}

\usepackage[skip=2pt]{caption}

\lstdefinestyle{promptstyle}{
    basicstyle=\scriptsize\ttfamily,
    breaklines=true,
    breakatwhitespace=true,
    frame=single,
    backgroundcolor=\color{gray!5}, 
    showstringspaces=false,
    columns=fullflexible,
    keepspaces=true
}

\newcommand{\senew}[1]{\textcolor{black}{#1}}
\newcommand{\se}[1]{\textcolor{black}{#1}}

\newcommand{\la}[1]{\textcolor{black}{#1}}
\newcommand{\gb}[1]{\textcolor{black}{#1}}
\newcommand{\mk}[1]{\textcolor{black}{#1}}
\newcommand{\crc}[1]{\textcolor{black}{#1}}
\newcommand{\abc}[1]{\textcolor{black}{#1}}
\usepackage[normalem]{ulem}

\newcommand{\dataset}{\textsc{Annotated}}
\newcommand{\datasetsub}[1]{\dataset$_{\textsc{#1}}$}
\newcommand{\metric}{\textsc{Reportage Score}}

\title{Who Annotates in NLP? A Large-scale Assessment of \\\ Human Annotation Reporting between 2018 and 2025}

\author{
  \textbf{Maria Kunilovskaya}\textsuperscript{1},
  \textbf{Gagan Bhatia}\textsuperscript{1},
  \textbf{Lisa Sophie Albertelli}\textsuperscript{1},\\
  \textbf{Yanran Chen}\textsuperscript{1},
    \textbf{Christian Greisinger}\textsuperscript{1},
  \textbf{Lotta Kiefer}\textsuperscript{1}, \\
  \textbf{Christoph Leiter}\textsuperscript{1},
  \textbf{Subhadeep Roy}\textsuperscript{1}, 
  \textbf{Tewodros Achamaleh}\textsuperscript{2}, \\ 
    \textbf{Muhammad Arslan Manzoor}\textsuperscript{2},
      \textbf{Sebastian Pohl}\textsuperscript{2},
       \textbf{Yufang Hou}\textsuperscript{2},
  \textbf{Steffen Eger}\textsuperscript{1} \\
  \textsuperscript{1}NLLG Lab University of Technology Nuremberg,\\ \textsuperscript{2} IT:U Interdisciplinary Transformation University Austria \\
  \texttt{\{\abc{maria.kunilovskaya}, gagan.bhatia, steffen.eger\}@utn.de}
}

\begin{document}

\maketitle
\begin{abstract}

\gb{Human annotation is the empirical foundation of much NLP research, from dataset construction to model evaluation, but papers often leave it unclear who produced the annotations and how the annotation process was controlled. 
We provide the first large-scale, task-level audit of human annotation reporting across major NLP venues, asking which annotation details are documented, which are missing, and how reporting varies across time, topic, venue, and intended use of human judgment. 
We introduce a unified taxonomy of annotation-reporting practices 
and validate an LLM-assisted extraction pipeline against \datasetsub{gold}, a human-adjudicated gold standard of 41 papers and 72 annotation tasks. 
Our best LLM model reaches human-comparable agreement with adjudicated labels (Krippendorff's $\alpha=0.606$ vs.\ $0.585$ for human--human agreement). 
Using this pipeline, we construct \datasetsub{llm}, a dataset \mk{covering ACL-venue papers} from 2018--2025, with 2,667 extracted annotation tasks from 1,603 papers.  
We find that papers frequently report operational details such as recruitment \mk{strategies}, annotator expertise, and annotation volume, but often omit details needed to assess annotation validity, including training, language proficiency, compensation, socio-demographics, adjudication, and agreement values, especially in model-evaluation studies. 
Our results show that annotation reporting in NLP has improved \senew{over time} but remains uneven. Based on these findings, we propose a scalable framework and bare-minimum reporting recommendations for making human annotation more reliable, reproducible, and interpretable.\abc{\footnote{Code and data are available at \url{https://github.com/NL2G/who-are-the-annotators}.}} 
}
\end{abstract}

\section{\label{sec:Introduction}Introduction}

Human annotation is one of the \se{core} foundations of empirical NLP. When evaluating a machine translation system, building an argument-mining dataset or comparing the outputs of large language models (LLMs), we often rely on human judgments as ground truth.
As a result, 
many NLP claims depend not only on models and metrics, but also on who the annotators \se{are}, what they \se{are} asked to do, how they are trained, how disagreements are handled~\citep{artstein2008survey,van-der-lee-etal-2019-best, popovic-belz-2021-reproduction}. 
\senew{Unqualified or biased human annotation may undermine the validity of \mk{findings in NLP research}: for example, a study evaluating 
\senew{AI-generated} poetry may critically depend on annotators' language proficiency\abc{,} and a study of political stance, sexism, or hate speech may yield different outcomes based on annotators' social and ideological positions.}

\senew{Worryingly, there is extremely little empirical evidence \mk{about} the annotators in the NLP community: does the community use \mk{crowdworkers} or experts (\mk{and in what proportions}), are the annotators adequate for the task, and how well are they compensated?}
\senew{Further, there 
is a lack of quantitative evidence on} 
whether initiatives such as the ACL Responsible NLP Checklist \cite{aclrollingreview_responsiblenlp}, 
\senew{introduced to improve the consistency, transparency, and ethics of AI research,} 
correspond to measurable improvements in annotation reporting quality, how these reporting practices differ across intended uses of human annotation, and which NLP areas systematically underreport information critical for interpreting annotation outcomes.
\gb{We address this gap with the first large-scale, task-level audit of human annotation reporting across major NLP venues.\footnote{Our task-level perspective is essential: a single paper may contain multiple annotation tasks with different annotators, instructions, purposes, and quality-control \abc{measures}. \mk{Treating papers as the unit of analysis would therefore obscure task-level variation in annotators, procedures, and quality control that are crucial for evaluating annotation setups.}}}

\gb{Existing work has examined annotation quality, reproducibility, annotator effects, and selected aspects of reporting practices \citep{bayerl-paul-2011-determines,belz2023missing,klie-etal-2024-analyzing,Beck2023}. However, prior studies typically focus on specific annotation problems or isolated reporting dimensions, such as inter-annotator agreement \citep{artstein2008survey,bayerl-paul-2011-determines}, reproducibility of human evaluation \citep{belz2023missing,belz20242024}, dataset quality management \citep{klie-etal-2024-analyzing}, or annotator demographics and data documentation \citep{bender-friedman-2018-data,pei-jurgens-2023-annotator}. In contrast, we provide the first large-scale, task-level, cross-topic analysis of annotation reporting across major NLP venues, using a unified taxonomy that covers not only agreement and workload, but also recruitment, qualifications, compensation, socio-demographics, and quality control.}
\senew{We also present the first large-scale empirical evidence on how reporting has changed \abc{over nearly a decade}, starting from 2018, \abc{which allows us to examine how reporting practices in the NLP community have changed over time.}} 

Overall\abc{,} we make the following contributions:

  \noindent  \textbf{(1) A task-level reporting dataset}. We introduce \dataset{}, a dataset for auditing human annotation reporting in NLP. It treats each annotation task as a separate unit, capturing within-paper differences in annotators, instructions, purposes, and quality-control procedures.
     
\noindent     \textbf{(2) A human-adjudicated gold standard}. We manually annotate \datasetsub{gold}, a gold-standard set of 41 papers comprising 72 human-annotation tasks, and finalize labels through adjudication. \gb{
   This compact set reflects the substantial cost of task-level expert annotation, estimated at approximately €6,300  in personnel time (Appendix~\ref{appx:demogr}), while providing a human-validated benchmark for evaluating automated methods that extract annotation-reporting information. 
    } 
 
  \noindent  \textbf{(3) An LLM-assisted extraction pipeline}. We evaluate six LLMs as structured extractors of annotation-reporting information. 
  \crc{The strongest model, Gemini-3.1-Pro, reaches Krippendorff's $\alpha=0.606$ against the adjudicated gold standard; a separate comparison against individual pre-adjudication human annotations further supports human-comparable extraction performance.}
    
 \noindent    \textbf{(4) The first large-scale audit of annotation reporting in ACL-venue papers}. We apply the validated pipeline to construct \datasetsub{LLM}, \mk{a corpus of ACL-venue papers} published between 2018 and 2025 \mk{and extract 2,667 annotation tasks from 1,603 papers}. \crc{We analyze reporting practices over time, venues, NLP topics, and uses of human judgment, including resource creation, model evaluation, and human-performance benchmarking.}
 

\noindent \textbf{(5) A scalable framework for more reliable annotation reporting}. Motivated by our large-scale audit of annotation-reporting practices, we propose a scalable framework and bare-minimum reporting recommendations for making human annotation more reliable, reproducible, and interpretable (\S\ref{sec:conclusion}). Since NLP models, benchmarks, and datasets often rely on human annotation for validation, improving annotation reporting has foundational implications for the interpretation of empirical claims in NLP. 




\section{\label{sec:related_work}Related Work}


\paragraph{Annotator effects, annotation quality, and reporting.}
Human annotation has long been recognized as central to NLP evaluation and dataset construction, with prior work emphasizing the importance of agreement measurement, evaluation design, and reproducibility \citep{artstein2008survey,van-der-lee-etal-2019-best,popovic-belz-2021-reproduction}. A growing body of research further shows that annotators are not interchangeable: their demographic backgrounds, language competence, cultural context, and social or political attitudes can systematically affect annotation outcomes. For instance, \citet{pei-jurgens-2023-annotator} show that demographic variables influence offensiveness judgments, while \citet{jiang-etal-2024-examining} find that annotators' social and political attitudes affect sexism and misogyny annotations. Related work highlights the need to document demographic and linguistic context in NLP datasets and evaluations \citep{bender-friedman-2018-data,joshi-etal-2020-state}. At the same time, work on reproducibility has shown that many human evaluations cannot be fully reproduced because papers omit essential information about annotators, materials, instructions, and procedures \citep{belz2023missing,belz20242024}. 
Earlier meta-analytic work examined annotation and annotator characteristics across 96 studies \citep{bayerl-paul-2011-determines}, annotated-data quality dimensions \citep{Beck2023}, and annotation quality management in dataset papers \citep{klie-etal-2024-analyzing}.

\paragraph{AI4Science and meta-scientific analysis.}
AI4Science studies how AI systems can support or transform scientific work, including literature analysis, discovery, experimentation, writing, figure generation, and evaluation \citep{pramanick-etal-2025-nature,chen2025ai4researchsurveyartificialintelligence,eger2026transformingsciencelargelanguage,xie2025bridgingaiscienceimplications}. Recent work has explored LLM-based support for peer review \citep{tyser2024aidrivenreviewsystemsevaluating}, abstract generation \citep{10.1002/pra2.1323}, and scientific graphics generation \citep{belouadi2025tikzerozeroshottextguidedgraphics,greisinger2026tikzillascalingtexttotikzhighquality}. In NLP, LLMs have also increasingly been used as annotators, evaluators, or assistants for scientific and empirical analysis, enabled by the general capabilities of LLMs \citep{brown2020languagemodelsfewshotlearners,grattafiori2024llama3herdmodels,singh2026openaigpt5card}. For example, LLMs have been used to support annotation in social-scientific analysis \citep{kostikova-etal-2024-fine, ziems-etal-2024-large} 
\abc{to} interpret latent semantic spaces learnt in pretrained models \citep{mousi-etal-2023-llms}, and \abc{to} identify ethical concerns in ACL papers \citep{karamolegkou-etal-2025-ethical}. Our work follows this emerging use of LLMs as tools for analysis, but applies them to the meta-scientific study of research practice itself: we use LLMs to extract structured information \mk{on} how human annotation is reported across NLP papers.


\section{\label{sec:human_annotation}Human Annotation Setup}
The annotation unit in this study is an \emph{individual human annotation task} rather than a paper, since papers may contain multiple studies with different reporting levels. A task is defined as a shared annotation setup in which annotators follow the same instructions and annotate the same data. 
\crc{For example, sentiment and stance labeling over the same tweets constitute separate annotation tasks when they use different instructions. We distinguish three intended uses: \emph{resource creation}, such as constructing a new NER corpus; \emph{human performance}, such as establishing a human QA baseline; and \emph{model-output evaluation}, such as rating the fluency of machine translations.}
In our experience, identifying the number of human tasks within papers was a non-trivial preliminary step. In cases of disagreement, a third annotation was commissioned. Papers with unresolved disagreements after the third annotation, as well as structurally complex papers judged by at least two annotators as containing more than three annotation tasks, were excluded from the final manual dataset.

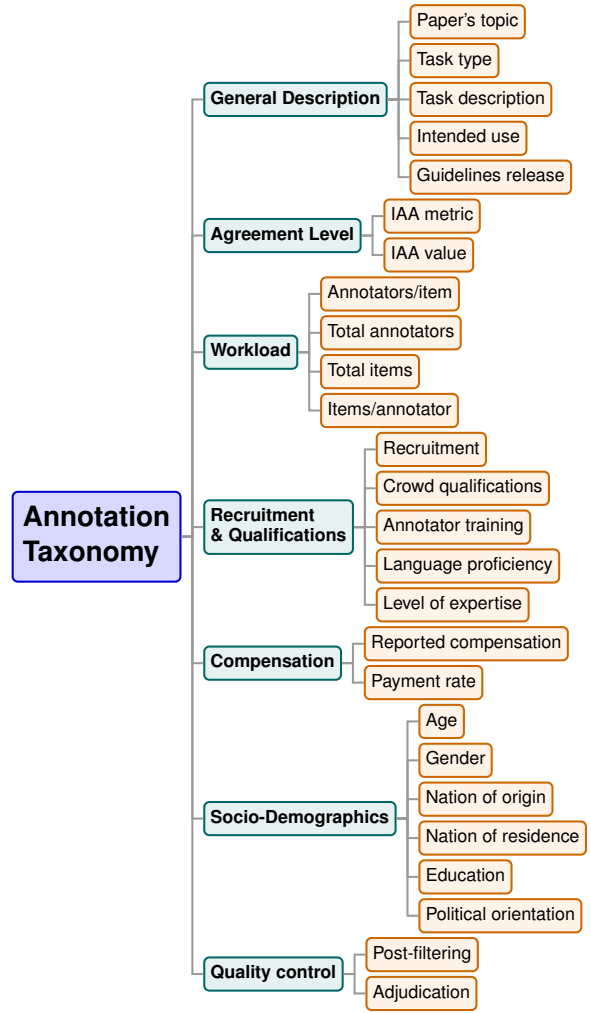
\begin{figure}[ht]
    \centering
    \resizebox{\columnwidth}{!}{
    \begin{forest}
      forked edges,
      for tree={
        grow'=0,                 
        parent anchor=east,      
        child anchor=west,       
        anchor=west,             
        align=left,              
        font=\sffamily\scriptsize, 
        draw,
        thick,
        rounded corners=2pt,
        inner sep=2.5pt,         
        l sep=3mm,               
        fork sep=1.5mm,          
        s sep=0.5mm,             
        edge={thick, draw=gray!80}, 
      },
      root/.style={fill=blue!15, draw=blue!80!black, font=\sffamily\bfseries\normalsize, inner sep=4pt},
      aspect/.style={fill=teal!10, draw=teal!80!black, font=\sffamily\bfseries\scriptsize},
      category/.style={fill=orange!10, draw=orange!80!black, font=\sffamily\scriptsize}
      [Annotation \\ Taxonomy, root 
        [General Description, aspect
          [Paper's topic, category]
          [Task type, category]
          [Task description, category]
          [Intended use, category]
          [Guidelines release, category]
        ]
        [Agreement Level, aspect
          [IAA metric, category]
          [IAA value, category]
        ]
        [Workload, aspect
          [Annotators/item, category]
          [Total annotators, category]
          [Total items, category]
          [Items/annotator, category]
        ]
        [Recruitment \\ \& Qualifications, aspect
          [Recruitment, category]
          [Crowd qualifications, category]
          [Annotator training, category]
          [Language proficiency, category]
          [Level of expertise, category]
        ]
        [Compensation, aspect
          [Reported compensation, category]
          [Payment rate, category]
        ]
        [Socio-Demographics, aspect
          [Age, category]
          [Gender, category]
          [Nation of origin, category]
          [Nation of residence, category]
          [Education, category]
          [Political orientation, category]
        ]
        [Quality control, aspect
          [Post-filtering, category]
          [Adjudication, category]
        ]
      ]
    \end{forest}}
    \caption{\textbf{Task-level taxonomy for annotation-reporting analysis.} The taxonomy groups 26 reporting categories into seven aspects: general task description, agreement, workload, recruitment and qualifications, compensation, socio-demographics, and quality control.
}
    \label{fig:annotation_taxonomy}
\end{figure}

Annotations were performed using a structured interface with integrated instructions that cover category rationale, definitions, examples, and coding rules. The process began with pilot annotations and extensive group discussions to refine the taxonomy and calibrate annotator interpretations. Closed-list selection was preferred over open-ended extraction to ensure consistent coding and easier aggregation of results. For some categories, annotators relied only on explicit statements in papers, while other tasks required informed judgment. The final taxonomy captures seven aspects of human annotation reporting across 26 categories. 
The full description of categories, permissible values, and instructions \abc{are} provided in Appendix~\ref{appx:humanno} (Table~\ref{tab:anno_guide}). A top-level overview of the taxonomy is shown in Figure~\ref{fig:annotation_taxonomy}. 
This taxonomy provides a unified framework for mapping heterogeneous human annotation tasks in ACL papers. 
Each paper was annotated by at least two independent annotators, including co-authors of this paper. 

In total, 12 annotators contributed to the annotation process: 2 professors, 2 postdoctoral researchers, 6 PhD students, and 2 master’s students. All annotators were proficient in reading academic English.
\gb{Human annotations were finalized through a two-stage adjudication process involving three annotators from the same pool \mk{of annotators}: two independent annotations of each identified task were compared, disagreements were discussed to reach consensus, and a third annotator adjudicated unresolved cases when necessary. The resulting consensus labels form the gold standard used to evaluate LLM extraction quality; we refer to this subset as \datasetsub{gold}. Figures~\ref{fig:paper_exc1}, \ref{fig:paper_exc2}, and \ref{fig:paper_exc3} in Appendix~\ref{appx:analysis} show an annotated paper excerpt and \abc{information extracted from it.}}

\section{\dataset~Dataset}
\label{sec:data}
\begin{table}[t]
\centering
\small
\resizebox{\columnwidth}{!}{
\begin{tabular}{@{} l ccc cc @{}}
\toprule
& \multicolumn{3}{c}{\textbf{Counts}} & \multicolumn{2}{c}{\textbf{Ratios}} \\
\cmidrule(lr){2-4} \cmidrule(l){5-6}
\textbf{Subset} & \textbf{Input} & \textbf{Annotated} & \textbf{Tasks} & \textbf{Ann.\ / Input} & \textbf{Tasks / Paper} \\
\midrule
\datasetsub{gold} & 61 & 41 & 72 & 0.672 & 1.756 \\
\datasetsub{LLM} & 1,995 & 1,603 & 2,667 & 0.820 & 1.664 \\
\bottomrule
\end{tabular}
}
\caption{\dataset{} subsets. Input papers are candidate papers before filtering; \mk{annotated} papers are papers with annotatable human annotation tasks; annotation tasks are task-level annotation records.}
\label{tab:coverage}
\end{table}

We began by retrieving papers from ACL Anthology publications appearing between 2018 and 2025 (three years before and after the introduction of the ACL checklist) in the main and Findings proceedings of ACL, EMNLP, NAACL, TACL, EACL, and AACL. Candidate papers were identified by matching 34 curated annotation-related keywords against titles and abstracts, including terms such as \textit{manual annotation} and \textit{human evaluation}. The keyword list (see Appendix \ref{appx:data}) was iteratively refined during pilot annotation sessions. 


\begin{table}[t]
\centering
\small
\resizebox{\columnwidth}{!}{
\begin{tabular}{lrr}
\toprule
 & \datasetsub{LLM} & \datasetsub{gold} \\
\midrule
Papers                    & 1{,}603 & 41 \\
Tasks                     & 2{,}667 & 72 \\
Tasks/paper               & 1.66 $\pm$ 0.82 & 1.76 $\pm$ 0.80 \\
Pages/paper               & 17.1 $\pm$ 6.9 & 17.0 $\pm$ 6.1 \\
\midrule
\multicolumn{3}{l}{\textit{Selected topics (\% of papers)}} \\
Open-ended generation     & 44.2 & 9.8 \\
Semantics/inference       & 25.1 & 2.4 \\
Subjective/social         & 21.5 & 26.8 \\
Linguistics               & 4.1 & 24.4 \\
\bottomrule
\end{tabular}}
\caption{\crc{Comparison of \datasetsub{LLM} and
\datasetsub{gold}. The subsets are similar in document length and
annotation-task density but differ in topic composition. Topic
labels are multi-label; selected topics show the largest or most
relevant distributional differences.}}
\label{tab:data}
\end{table}

\gb{\noindent\textbf{\datasetsub{gold}} is the manually annotated \mk{and} adjudicated gold-standard set used to evaluate LLM extraction quality. It consists of 41 papers with reliably annotatable content out of 61 initially \abc{selected} \mk{to construct a manual benchmark}. The papers were excluded when the number of annotation tasks could not be resolved after a third annotation, or when at least two annotators identified the paper as containing more than three difficult-to-annotate human annotation tasks. The retained papers contain 72 distinct human annotation tasks. The resulting adjudicated \mk{dataset is used} exclusively for evaluating LLM performance. The size of \datasetsub{gold} reflects the cost of expert annotation: the manual annotation and adjudication effort is estimated at approximately $6,300$ Euros in personnel time, with details provided in Appendix~\ref{appx:demogr}.}

\gb{\noindent\textbf{\datasetsub{LLM}} is the large-scale LLM-extracted set used for the main analysis. The keyword-based sampling yielded 1,995 candidate papers. After filtering papers without annotatable human annotation content, \datasetsub{LLM} contains 1,603 papers and 2,667 extracted annotation tasks. \datasetsub{LLM} strictly excludes all 61 papers sampled for \datasetsub{gold}, ensuring that the gold-standard evaluation set remains separate from the large-scale analysis. Table~\ref{tab:coverage} summarizes the number of input papers, retained papers, and annotation tasks in each subset.}

\crc{This venue restriction applies to the large-scale
\datasetsub{LLM} analysis. A small number of papers from
non-ACL venues were included only during construction of
\datasetsub{gold} and early taxonomy development, allowing us
to test the taxonomy on a broader range of AI and multimodal annotation settings. These papers are not part of the large-scale analysis. We exclude LREC from the present analysis because our temporal study targets the ACL venue ecosystem in which the Responsible NLP Checklist was introduced; LREC has a distinct, strongly resource-oriented research and reporting culture\abc{,  and is a possible direction for future work}.}

\crc{
Table~\ref{tab:data} compares \datasetsub{gold} with the
large-scale \datasetsub{LLM} corpus. The two subsets are highly
similar in document length (17.0 vs.\ 17.1 pages on average) and
annotation density (1.76 vs.\ 1.66 tasks per paper). Their topic composition differs more noticeably (JSD $=0.161$; Cram\'er's $V=0.18$), with \datasetsub{gold} containing relatively more linguistics papers and fewer open-ended generation papers.}

\paragraph{Sampling validation.}
\mk{\datasetsub{LLM} is constructed through keyword-based retrieval, which may introduce sampling bias relative to all ACL-venue papers. To assess the effect of this sampling strategy, we compare it with a stratified random sample from the same venue-year scope. Although the comparison reveals statistically significant differences in frequency distributions for some category-value pairs, the effects are generally modest: the average absolute difference across 67 \abc{observed value frequencies} does not exceed 5.2 percentage points, with effects lacking directional consistency (see detailed analysis in Appendix~\ref{appx:random_validation}). 
We interpret these deviations as sufficiently limited for the purposes of LLM extraction, particularly given the computational, financial, and environmental costs associated with processing the full ACL venue-year population.
Accordingly, we treat \datasetsub{LLM} as a high-recall, annotation-focused corpus suitable for studying reporting practices in papers likely to involve human annotation, 
without claiming it represents all ACL-venue papers.}

\section{Experiment Setup}
\label{sec:experiment}

\paragraph{Extraction.}
We evaluate six LLMs under a unified extraction protocol: three proprietary (\texttt{Gemini-3.1-Pro} \cite{Gemini3}, \texttt{Gemini-3.1-Flash-Lite} \cite{Gemini3}, \texttt{GPT-4.1} \cite{GPT-5}) and three open-weight (\texttt{Qwen3.6-27B} \cite{qwen3.6-27b}, \texttt{gemma-4-31B-it} \cite{gemma4}, \texttt{gpt-oss-120b} \cite{openai2025gptoss120bgptoss20bmodel}). We include both proprietary and open-weight models because prior work shows that closed models often provide strong performance, while open models offer advantages in cost and transparency \citep{oketch-etal-2025-bridging}. Each model is prompted, with the paper converted to plain text, to populate the same \crc{26-category} taxonomy used in human annotation, outputting one JSON record per distinct annotation experiment identified in the paper. The prompt (reproduced in full in Appendix~\ref{app:llm_extraction_prompt}) encodes the complete taxonomy with exact allowed values, field-level decision rules, interdependency constraints, and a self-audit checklist that instructs the model to continue scanning beyond the first annotation section found: a common failure mode that causes multi-experiment papers to be under-extracted. Model output is constrained to a JSON-defined schema via structured-output APIs where available, eliminating free-form generation and enforcing value types at decoding time. To handle long documents within the context constraints of open-weight models, we apply annotation-section chunking: sections likely to contain annotation metadata (e.g., \emph{Annotation Procedure}, \emph{Human Evaluation}, \emph{Data Collection}) are identified via regex heuristics and passed to the model within an 8{,}000-token budget; papers with no detectable annotation sections are truncated to 20{,}000 tokens and processed in full. Proprietary models with million-token context windows receive the full paper text without truncation. Papers for which the model identifies no human annotation experiments are flagged and excluded. 
\crc{The final \datasetsub{LLM} dataset is
constructed with Gemini-3.1-Pro, which receives the complete paper text. The chunking and truncation procedures described above apply only to the open-weight models included in the model-comparison experiment.}

\paragraph{Evaluation.}
LLM-extracted values are compared to gold-standard labels at the level of identified tasks. 
\gb{For papers with multiple tasks, we first align each gold task with the corresponding LLM-extracted task using the short task description recorded during annotation.}
For example, a gold task described as \emph{sentiment labeling of tweets} can be matched to a model-extracted task described as \emph{tweet sentiment annotation}. Before comparison, all values are normalized to a canonical form: IAA metric surface variants (e.g., \emph{Cohen kappa}, $\kappa$, \emph{Cohen's K}) are collapsed to a single canonical label per metric; numeric fields accept shared integer tokens as agreement (allowing, e.g., \emph{27.9K} and \emph{27900} to match); list-valued fields (\texttt{iaa\_metric\_name}, \texttt{papers\_topic}, \texttt{intended\_use}) are compared via set intersection, with partial overlap counting as agreement. We report per-\mk{value} and per-category exact-match agreement rates alongside Krippendorff's~$\alpha$. 


\paragraph{\metric.}
\abc{
To assess the level of reporting associated with human annotation in NLP papers, we compute a \emph{\metric} for each identified annotation task. We start from the assumption that all annotation attributes in our inventory are desirable to report when they are applicable to a given annotation setup. The score therefore measures the proportion of applicable annotation attributes that are explicitly reported.}

Formally, for an annotation task $t$, let $A_t$ denote the set of applicable reporting attributes and let $R_t \subseteq A_t$ denote the subset of applicable attributes that are reported. The \metric~is defined as:
\[
\mathrm{\metric}(t) = \frac{|R_t|}{|A_t|}.
\]

\abc{The denominator $A_t$ is task-sensitive because not every reporting attribute is applicable to every annotation setup. Some attributes have no applicability restrictions and are therefore expected for all tasks: \textit{guideline availability, recruitment strategy, annotator training, expertise, language proficiency, education, number of annotators, number of annotated items, compensation, and post-annotation filtering}. Other attributes are excluded whenever the corresponding annotation setup makes them irrelevant. For example, inter-annotator agreement and adjudication are excluded for single-annotator tasks, including text production assignments; crowdworker screening is excluded when crowdsourcing is not used; and redundant workload fields are excluded when they encode information already captured by other categories. For tasks in \textit{Subjective language \& social meaning}, five socio-demographic attributes are added to the denominator because annotator background is particularly relevant for interpreting subjective or socially grounded judgments. Thus, failing to report an applicable attribute lowers the score, whereas an attribute that is not applicable to the task does not affect it. Full scoring rules are provided in Appendix~\ref{appx:score}.}

\section{\label{sec:results}Results}


In \S\ref{ssec:iaa}, we first analyze annotation agreement, including both human--human and human--LLM agreement. We then examine broader reporting trends in \S\ref{ssec:trends}, using \datasetsub{LLM} and addressing our research questions on annotation intensity, reporting variation across NLP areas, \mk{the impact of ACL Responsible NLP Checklist}, and the intended use of human judgment.
\subsection{\label{ssec:iaa}Agreement Analysis} 
\paragraph{\se{Human--Human agreement}} 
Overall\se{,} human--human exact-match agreement reache\se{s} 79.2\%, with a macro-averaged Krippendorff's $\alpha$ of 0.585, \se{which is commonly regarded as decent agreement.}  
\se{There is substantial variation across annotation dimensions.} 
Fields corresponding to clearly defined and explicitly reported information, including demographic attributes, compensation, and inter-annotator agreement metrics, consistently achieve high reliability ($\alpha > 0.8$). In contrast, lower agreement is concentrated in categories that require interpretation or inference from less formalized descriptions (e.g., guideline availability, annotator expertise, recruitment, and quality control procedures). This pattern is expected given the nature of the task: disagreement arises primarily in cases where the annotated papers provide incomplete or ambiguous information. 
Importantly, all annotations were subsequently adjudicated through extensive discussion, yielding a consensus dataset that resolves individual disagreements. Such adjudication \abc{can be expected to}
to produce labels that more closely approximate the underlying ground truth than independent annotations and provides a sufficiently robust benchmark for evaluating LLMs\mk{~\cite{klie-etal-2024-analyzing}}. 

\paragraph{\se{Human--LLM agreement}} 
Table~\ref{tab:model_evaluation} presents the overall performance of \se{our} six tested LLMs (three open-source and three proprietary). 
Gemini-3.1-Pro shows the strongest results. 
\begin{table}[t]
    \centering
    \resizebox{\columnwidth}{!}{
    \begin{tabular}{lccc}
        \toprule
        \textbf{Model Name} & \textbf{Context Len} & \textbf{Agree\%} & \textbf{Kripp $\alpha$} \\
        \midrule
        Gemini-3.1-Flash-Lite & 1M   & 72.4 & 0.50 \\
        Gemini-3.1-Pro      & 1M   & \textbf{79.9 }& \textbf{0.61} \\
        GPT-4.1             & 1M   & 75.2 & 0.50 \\
        Qwen3.6-27B         & 262k & 63.4 & 0.42 \\
        gemma-4-31B-it      & 256k & 65.8 & 0.45 \\
        gpt-oss-120b        & 131k & 66.9 & 0.47 \\
        \midrule
        Human               & --   & \underline{79.2} & \underline{0.59} \\
        \bottomrule
    \end{tabular}}
    \caption{Evaluation results comparing various models against human baseline performance.}
    \label{tab:model_evaluation}
\end{table}
Overall, the \se{best} LLM achieves slightly higher macro-level agreement than human--human annotation (79.9\% vs. 79.2\%) and a marginally higher Krippendorff’s $\alpha$ (0.606 vs.\ 0.585), indicating that model-assisted annotation is not only consistent with human judgments but, in aggregate, slightly more stable against the gold standard.
Individual per-category human--human and human--LLM agreement values are reported in Table~\ref{tab:iaa_full} \mk{(Appendix~\ref{appx:humanno})}.
\gb{Several categories exhibit a mismatch between high observed percentage agreement and moderate or low $\alpha$, driven by class imbalance and skewed label distributions. Two extreme cases illustrate this effect: for `political orientation', the absence of positive instances yields perfect agreement but limited interpretability, whereas for `nation of origin', scarcity of positive cases makes agreement estimates unstable. For rare categories, $\alpha$ can become negative when observed agreement is lower than the agreement expected by chance under the label distribution. Thus, the negative value for `nation of origin' \mk{in human--human IAA} should not be interpreted as substantive \abc{disagreement}, but as an instability caused by very sparse positive cases.}
\crc{To obtain a symmetric comparison, we additionally compare
Gemini-3.1-Pro against the pooled individual pre-adjudication human
annotations over 126 matched subtasks. The model reaches 74.8\%
exact-match agreement and a macro-averaged Krippendorff's
$\alpha=0.602$. Together with the independent
human--human baseline of 79.2\% agreement and $\alpha=0.585$, this
provides a more directly comparable assessment of model and human
annotation consistency.}
\subsection{\label{ssec:trends}Trends in Reporting} 
\abc{After evaluating all six models on \datasetsub{gold}, we select the best-performing model, Gemini-3.1-Pro, for the large-scale extraction. This model is applied to \datasetsub{LLM} to produce the final dataset used in the downstream analysis,} which cost a total of \abc{$8,300$ Euros}. To analyze temporal trends in methodological reporting and its association with factors such as paper topic, intended use of human judgment, and venue, we use the \emph{\metric}, 
see 
\S\ref{sec:experiment}. %

\paragraph{RQ1: Which NLP areas are more human-annotation intensive, and which aspects of annotation practice are more consistently reported?} 

\begin{figure*}[!ht]
    \centering
    \includegraphics[width=\textwidth]{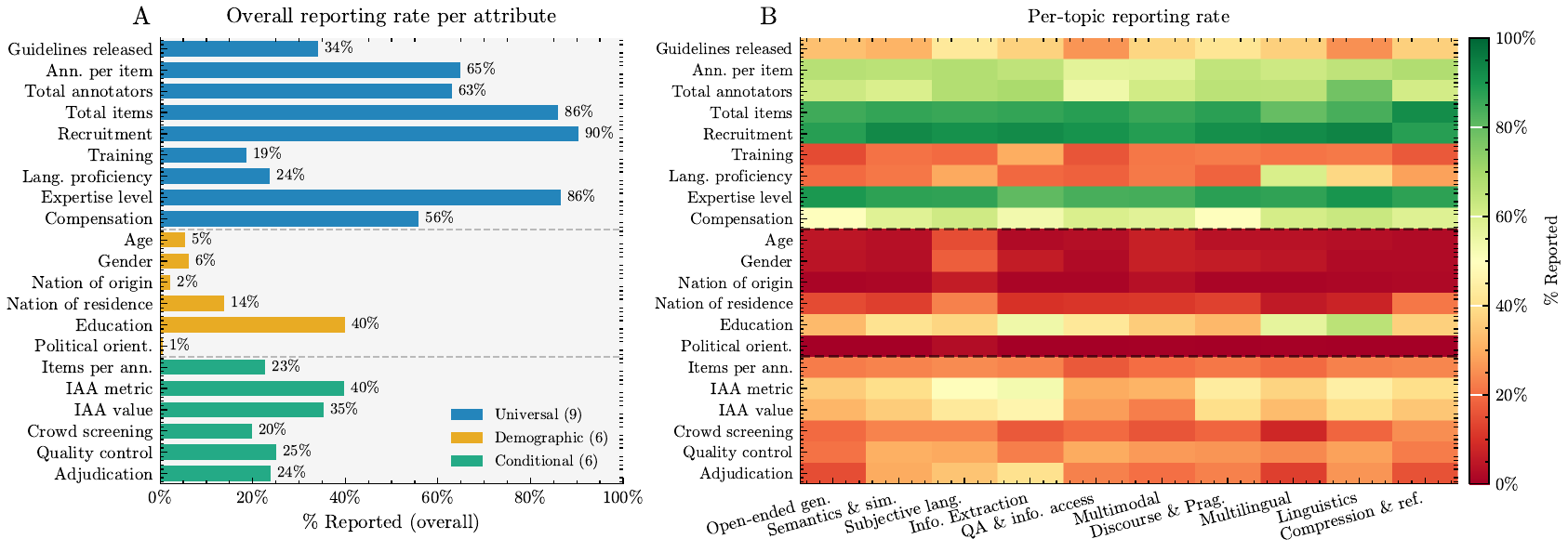}
    \caption{
    \textbf{Reporting patterns across annotation attributes.} Panel~A shows the overall percentage of annotation tasks for which each attribute is reported. Panel~B shows the corresponding reporting rates by NLP topic. Darker green cells indicate higher reporting rates within a topic; orange and red cells indicate lower reporting rates. 
    }
    \label{fig:reportage_importance}
\end{figure*}

\abc{Figure~\ref{fig:reportage_importance} shows reporting rates for individual annotation attributes and their variation across NLP topics.}
\gb{Panel~A shows that papers most often report operational details: recruitment (90.4\%), annotator expertise (86.5\%), and total number of annotated items (86.0\%). In contrast, reporting is much weaker for attributes that help readers assess the annotation process itself, including annotator training (18.7\%), language proficiency (24.0\%), and released annotation guidelines (34.1\%).} \mk{Panel~B shows that this imbalance is broadly consistent across NLP areas. Language proficiency is reported somewhat more frequently in papers involving multilingual tasks and linguistic annotation. 
As expected, annotator demographics are reported somewhat more often in papers dealing with subjective judgments, preferences, and socially constructed meaning (\textit{Subjective language and social meaning} topics), i.e. tasks that investigate social and cultural phenomena through language, such as hate speech, bias, stance, sentiment, and humor annotation.} 
We examine \textit{Subjective language and social meaning} category separately because annotator background, language proficiency, and social positioning can directly affect judgments in these tasks. However, compared with the rest of the dataset, these papers do not show systematically stronger reporting overall (Figure~\ref{fig:ss_vs_rest}). There are some distinctive tendencies: recruitment information and annotator language proficiency are reported slightly more often, crowdworkers and mixed annotator pools are used more frequently, and authors are less often used as annotators (Figures~\ref{fig:recruit_ss} and~\ref{fig:nativeness}, Appendix~\ref{appx:analysis}). These results suggest that socially oriented NLP papers are slightly more attentive to annotator sourcing and language background, but still do not consistently report the broader methodological details needed to assess annotation validity.

\begin{figure}[!ht]
    \centering
    \resizebox{\columnwidth}{!}{%
        \includegraphics{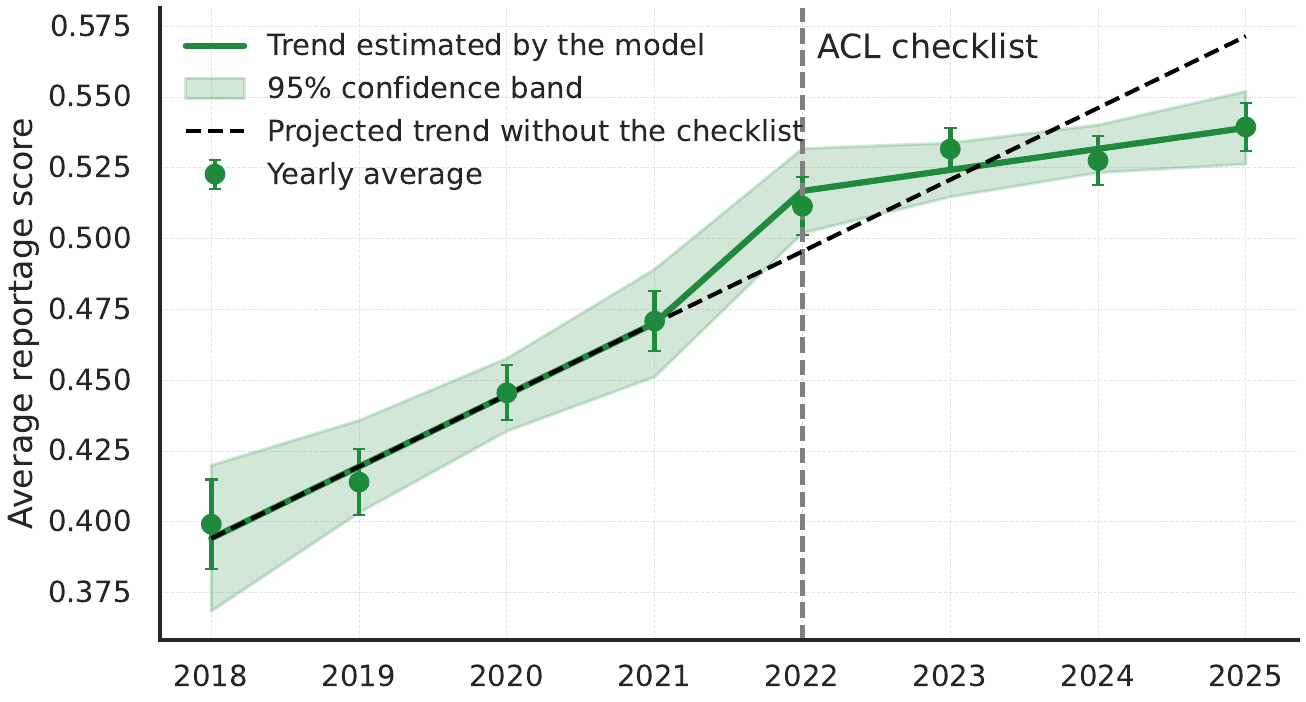}
    }
    \caption{\label{fig:checklist}
    \textbf{Interrupted time-series analysis of \metric{} before and after the ACL Responsible NLP Checklist.} Points show observed yearly mean \metric{} values, error bars indicate standard errors, the solid line shows the fitted trend, and the dashed line shows the counterfactual continuation of the pre-2022 trend. The vertical dashed line marks the checklist introduction.
}  
\vspace{-8pt}   
\end{figure}

\paragraph{RQ2: What is the impact of introducing ACL responsible checklist (2022)?}
\gb{The ACL Responsible NLP Checklist \cite{aclrollingreview_responsiblenlp} asks authors to reflect on data, annotation, and ethical considerations. \mk{Accordingly,} we test whether its introduction around 2022 coincides with a measurable change in human-annotation reporting. We fit an interrupted time-series regression model that estimates both an immediate post-2022 level shift and a change in the post-2022 trend.}
Figure~\ref{fig:checklist} shows the fitted interrupted time series regression line with confidence intervals, alongside the observed yearly reportage scores (green dots with error bars) and the counterfactual continuation of the pre-2022 trend. Reportage scores increase steadily between 2018 and 2021, followed by a modest upward level shift in 2022 and a visibly flatter post-2022 trend relative to the counterfactual continuation of the pre-checklist trajectory.
Our analysis finds no evidence that the introduction of the ACL checklist in 2022 produced a substantial immediate improvement in reporting scores. 
Although reportage quality continues to improve after 2022, the rate of improvement appears lower than in the preceding years, suggesting a flattening of the earlier upward trend.
We can offer two explanations for this finding. First, aggregate averages may obscure heterogeneous developments across different categories of annotation studies, potentially masking opposing trends in reporting practices. Second, the upward trend already visible before 2022 suggests that the checklist may have formalized practices that were already emerging in the field. Moreover, because the ACL Responsible NLP Checklist (introduced at NAACL 2022) was based on the NeurIPS 2021 checklist, awareness of annotation reporting standards may have begun increasing earlier, which aligns with the sharper rise in reporting levels observed around 2021 for some types of annotation studies. A venue-specific interrupted time series analysis for ACL, EMNLP, and NAACL \abc{and methodology are} provided in Appendix~\ref{appx:analysis}. Table~\ref{tab:binned} in Appendix~\ref{appx:analysis} offers a detailed per-category comparison for pre- and post-2022 periods.

\begin{figure}[t]
    \centering
    \resizebox{\columnwidth}{!}{%
        \abc{\includegraphics{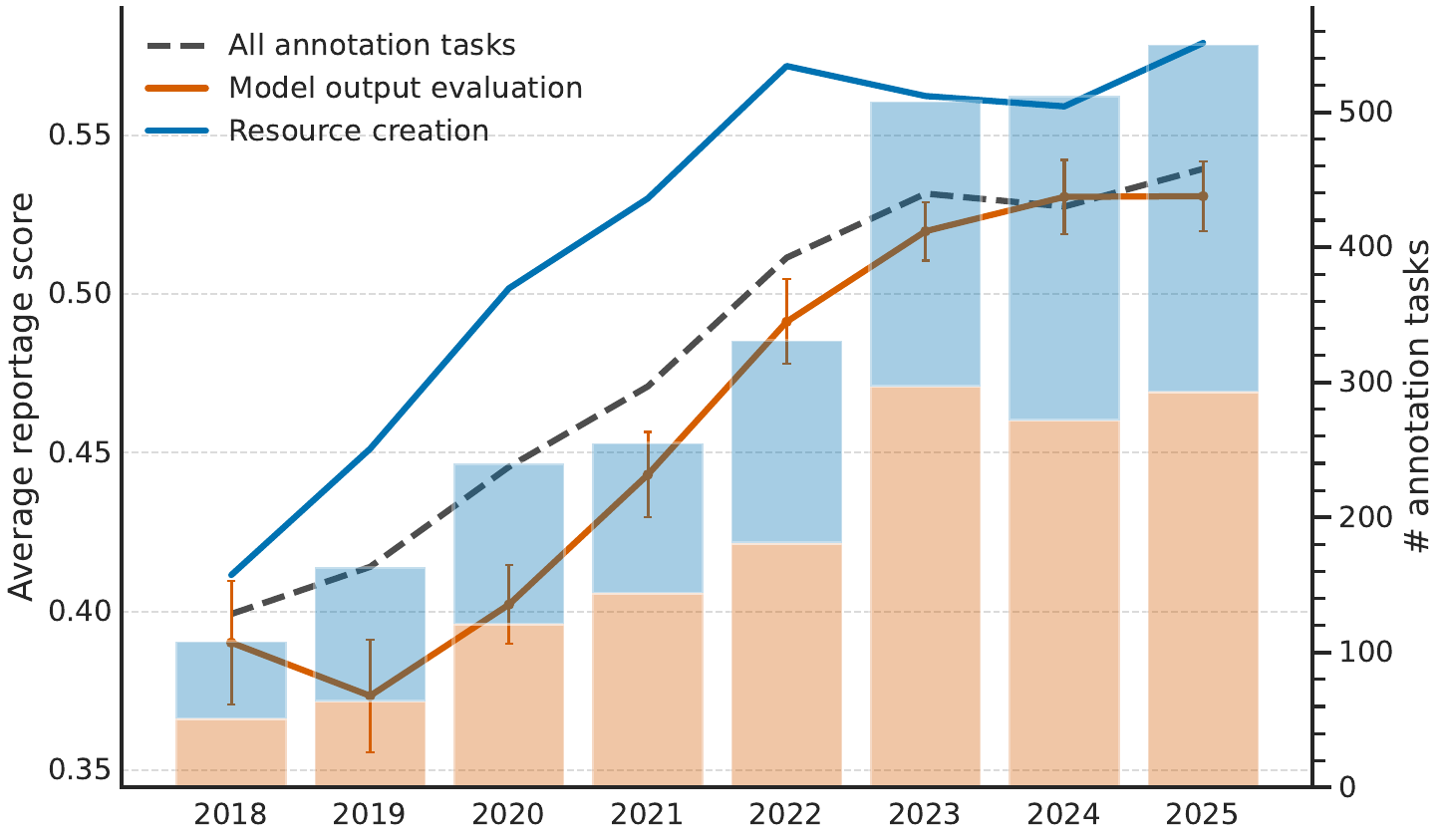}}
    }
    \caption{\label{fig:score_by_uses}
   \textbf{Mean \metric{} by intended use of human annotation.} Lines show yearly mean \metric{} values for all annotation tasks, model-output evaluation tasks, and resource-creation tasks; error bars indicate standard errors. \abc{Stacked bars in the background show the annual number of annotation tasks in the two intended-use categories, illustrating changes in task composition over time.}}
\end{figure}

\paragraph{RQ3: Does the intended use of human judgment affect reportage quality?}
\gb{We use the \emph{intended use} labels from our taxonomy to compare annotation tasks whose primary purpose is resource creation, model output evaluation, or human-performance benchmarking. These labels are assigned during the human and LLM annotation process evaluated in \S\ref{ssec:iaa}. Because human-performance benchmarking is comparatively rare, Figure~\ref{fig:score_by_uses} focuses on the two most frequent uses: resource creation and model output evaluation.}

While overall reportage quality improves over time, resource creation studies consistently report substantially more methodological detail than model evaluation studies, with the gap persisting throughout the observation period. 
In particular, evaluation-focused work more frequently omits information about annotator recruitment, compensation, training, and quality-control procedures.
To further examine this pattern, we fit a logistic regression model predicting the reporting of quality-control-related categories in our annotation scheme, including adjudication, post-filtering, annotator training, IAA metrics, and recruitment source (full results in Appendix~\ref{appx:analysis}). Resource creation studies are substantially more likely to report such measures ($p < 0.001$), while publication year shows only a weak positive association after controlling for annotation use type. The same overall pattern holds for recruitment strategy and compensation reporting, suggesting that model evaluation studies systematically \abc{underreport} key aspects of annotation methodology.

\abc{The trajectories by intended use provide evidence against a purely compositional explanation of the flatter post-2022 reportage trend observed in RQ2. Such an explanation could arise from the post-2022 growth of model-output evaluation tasks, which \abc{tend to} report less methodological detail. Despite sharply diverging task volumes after 2022 (stacked bars in Figure~\ref{fig:score_by_uses}), reporting scores for both resource-creation and model-output-evaluation tasks flatten at approximately the same time. In particular, the earlier increase in reporting for resource-creation tasks also levels off after 2022, even as their share of the two task types decreases with the growth of model-output evaluation. Thus, the aggregate post-2022 flattening cannot be explained solely by a shift in task composition, although the observational design does not establish a causal effect of the checklist.}

\section{Conclusion and Recommendations}
\label{sec:conclusion}

\gb{Human annotation remains a foundation of NLP research, but our results show that the field still lacks consistent standards for documenting who annotators are, how annotation is organized, and how annotation quality is controlled. We provide the first large-scale, task-level audit of human annotation reporting across major NLP venues. 
Our findings show clear progress, but also persistent blind spots: NLP papers often report who annotators are in broad terms, especially recruitment source, expertise, and annotation scale, but they provide a much less complete picture of whether annotators are trained, qualified, compensated, demographically located, or involved in systematic disagreement resolution. In particular, compensation, socio-demographics, training, adjudication, and agreement values remain substantially underreported. These omissions matter: without such information, 
readers cannot assess annotator suitability, potential background effects, disagreement handling, or whether the labels support the paper’s claims.
} \senew{Our findings may also help explain why human annotations are often difficult to reproduce in 
NLP~\citep{belz2023missing}.}

\gb{We draw three recommendations.} 

\abc{First, papers should \textbf{report all relevant annotation details for every human annotation task} they contain. This includes attributes applicable across annotation setups without restrictions: annotator recruitment, number of annotators, number of annotated items, annotators per item, training, language proficiency, expertise, compensation, quality control, and access to annotation guidelines, as well as any additional attributes relevant to the particular task and annotation procedure.}


\abc{Second, \textbf{papers on socially grounded and subjective language tasks should report relevant socio-demographic and background information about annotators}, such as age, gender, nation of residence, nation of origin, and political orientation, as these characteristics may shape the judgments being collected and their interpretation.}

\gb{Third, \textbf{model-evaluation studies require substantially more annotation details}. Our analyses show that when annotation is used for model evaluation, it is documented less than when annotation is used for resource creation, even though such evaluations often support central empirical claims.}

These details are not optional metadata; they are necessary for readers to assess whether the annotation procedure is reliable, interpretable, and reproducible. 

\crc{As LLMs are used as annotators and evaluators,
analogous reporting standards are needed for machine annotation,
including the exact model and version, prompts and decision rules,
context handling, decoding or sampling configuration, and
procedures for validation and evidence tracing. Developing such
standards is a natural extension of the framework proposed here.}

\gb{Overall, our study reframes annotation reporting as a measurable methodological practice. The proposed taxonomy 
and LLM-assisted pipeline provide a scalable basis for monitoring whether NLP papers document human annotation in ways that support reliability and reproducibility.}

\section*{Limitations}

A first limitation concerns the design and reliability of the annotation taxonomy. Some categories require fine-grained interpretation of incomplete or inconsistent paper descriptions, making perfect agreement difficult. We retained these categories because they capture information important for assessing the quality and transparency of the annotation process. The moderate agreement observed for some attributes should therefore be read not only as a limitation of our annotation scheme; it is also evidence that NLP papers often describe annotation procedures in ways that are difficult to interpret consistently.
A second limitation is that the large-scale analysis relies on LLM-based extraction. Although we validate the extraction pipeline against a human-adjudicated gold standard, \mk{extraction} errors remain possible and may affect individual category estimates. Moreover, \datasetsub{gold} is necessarily limited in size because task-level expert annotation is costly; the manual annotation and adjudication effort is estimated at approximately \abc{6,300 Euros} in personnel time, as detailed in Appendix~\ref{appx:demogr}. We therefore interpret our results as aggregate reporting patterns rather than definitive judgments about individual papers. Finally, \datasetsub{LLM} is an annotation-focused corpus constructed through keyword retrieval and should not be interpreted as a fully representative sample of all ACL-venue papers.

\crc{A third set of limitations concerns the granularity of our
taxonomy and released extractions. First, while we record annotator
language proficiency and whether a paper concerns multilingual or
translation research, we do not represent the specific language
identity of the annotated dataset as a structured field. Reporting
patterns may differ systematically across languages, particularly
in low-resource settings, and future versions of the taxonomy
should therefore record target-language identity explicitly.
Second, \textsc{AnnotatedLLM} does not provide calibrated
per-instance confidence scores. We instead report category-level
human--human and human--LLM reliability and will release these
values in machine-readable form, together with verbatim evidence
for selected extraction fields; nevertheless, users should exercise
additional caution for low-agreement categories. Finally, because
keyword retrieval may favor papers that foreground human
annotation, our absolute reporting rates may be optimistic relative
to the broader ACL literature and should not be interpreted as
population estimates.}

\section*{Ethics Statement}

This work analyzes publicly available scientific papers and does not introduce new experiments involving external human subjects. The manual annotation was conducted by the authors and collaborators on published research articles. Nevertheless, the study raises ethical questions about annotation labor and research transparency. In particular, our findings show that compensation, annotator demographics, training, and quality-control procedures are often underreported in NLP papers. We do not interpret missing reporting as evidence that these practices were absent; rather, we treat it as a transparency gap that limits readers' ability to assess annotation validity, fairness, and reproducibility.
Our LLM-assisted extraction pipeline also introduces ethical considerations. Automated extraction may misclassify individual papers or categories, especially when papers describe annotation procedures ambiguously. For this reason, we validate the pipeline against a human-adjudicated gold standard and report aggregate trends rather than using the system to rank or criticize individual papers. \abc{Released dataset and code} should be used to study community-level reporting practices and to support better documentation standards, not to assign blame to individual authors.

\section*{Broader Impact}

This work aims to improve the transparency and reproducibility of human annotation in NLP. By providing a task-level taxonomy, a reporting-completeness metric, and an LLM-assisted extraction pipeline, the study offers tools for monitoring how the field documents annotation labor, quality control, and annotator characteristics. The most direct positive impact is methodological: clearer reporting can help reviewers, readers, and future researchers assess whether annotation-based claims are reliable, comparable, and reproducible. It can also support future checklist design by identifying which annotation details are most often missing. \mk{Importantly, reporting requirements may encourage researchers to reflect more carefully on annotation procedures before data collection begins.}
At the same time, reporting standards must be applied carefully. Calls for more demographic information should not pressure authors to collect sensitive personal data when it is unnecessary, unsafe, or ethically inappropriate. Instead, task-sensitive reporting should mean documenting information that is relevant to interpreting annotation outcomes, while respecting annotator privacy and consent. \abc{More broadly, our goal is to promote more deliberate consideration and transparent reporting of the attributes relevant to each annotation setup.}

\section*{Acknowledgments}
We are grateful to the four anonymous reviewers for their support and constructive criticism, which helped us to improve the paper.
The project behind this paper began during a joint retreat of two research groups: the NLLG Lab at UTN, Germany, and the NLP Lab at IT:U Interdisciplinary Transformation University Austria.
The retreat took place at Speinshart Monastery in Bavaria in February 2026. We are very grateful to the Speinshart Scientific Center for AI and SuperTech for hosting us during this one-week Connect@Speinshart retreat. 
The NLLG Lab further gratefully acknowledges support from the German Research Foundation (DFG) through the Heisenberg Grant EG 375/5-1.

\bibliography{speinshart}

@article{artstein2008survey,
  title={Survey article: Inter-coder agreement for computational linguistics},
  author={Artstein, Ron and Poesio, Massimo},
  journal={Computational linguistics},
  volume={34},
  number={4},
  pages={555--596},
  year={2008}
}

@inproceedings{pramanick-etal-2025-nature,
    title = "The Nature of {NLP}: Analyzing Contributions in {NLP} Papers",
    author = "Pramanick, Aniket  and
      Hou, Yufang  and
      Mohammad, Saif M.  and
      Gurevych, Iryna",
    editor = "Che, Wanxiang  and
      Nabende, Joyce  and
      Shutova, Ekaterina  and
      Pilehvar, Mohammad Taher",
    booktitle = "Proceedings of the 63rd Annual Meeting of the Association for Computational Linguistics (Volume 1: Long Papers)",
    month = jul,
    year = "2025",
    address = "Vienna, Austria",
    publisher = "Association for Computational Linguistics",
    url = "https://aclanthology.org/2025.acl-long.1224/",
    doi = "10.18653/v1/2025.acl-long.1224",
    pages = "25169--25191",
    ISBN = "979-8-89176-251-0"
}

@misc{brown2020languagemodelsfewshotlearners,
      title={Language Models are Few-Shot Learners}, 
      author={Tom B. Brown and Benjamin Mann and Nick Ryder and Melanie Subbiah and Jared Kaplan and Prafulla Dhariwal and Arvind Neelakantan and Pranav Shyam and Girish Sastry and Amanda Askell and Sandhini Agarwal and Ariel Herbert-Voss and Gretchen Krueger and Tom Henighan and Rewon Child and Aditya Ramesh and Daniel M. Ziegler and Jeffrey Wu and Clemens Winter and Christopher Hesse and Mark Chen and Eric Sigler and Mateusz Litwin and Scott Gray and Benjamin Chess and Jack Clark and Christopher Berner and Sam McCandlish and Alec Radford and Ilya Sutskever and Dario Amodei},
      year={2020},
      eprint={2005.14165},
      archivePrefix={arXiv},
      primaryClass={cs.CL},
      url={https://arxiv.org/abs/2005.14165}, 
}

@misc{singh2026openaigpt5card,
      title={OpenAI GPT-5 System Card}, 
      author={Aaditya Singh and Adam Fry and Adam Perelman and Adam Tart and Adi Ganesh and Ahmed El-Kishky and Aidan McLaughlin and Aiden Low and AJ Ostrow and Akhila Ananthram and Akshay Nathan and Alan Luo and Alec Helyar and Aleksander Madry and Aleksandr Efremov and Aleksandra Spyra and Alex Baker-Whitcomb and Alex Beutel and Alex Karpenko and Alex Makelov and Alex Neitz and Alex Wei and Alexandra Barr and Alexandre Kirchmeyer and Alexey Ivanov and Alexi Christakis and Alistair Gillespie and Allison Tam and Ally Bennett and Alvin Wan and Alyssa Huang and Amy McDonald Sandjideh and Amy Yang and Ananya Kumar and Andre Saraiva and Andrea Vallone and Andrei Gheorghe and Andres Garcia Garcia and Andrew Braunstein and Andrew Liu and Andrew Schmidt and Andrey Mereskin and Andrey Mishchenko and Andy Applebaum and Andy Rogerson and Ann Rajan and Annie Wei and Anoop Kotha and Anubha Srivastava and Anushree Agrawal and Arun Vijayvergiya and Ashley Tyra and Ashvin Nair and Avi Nayak and Ben Eggers and Bessie Ji and Beth Hoover and Bill Chen and Blair Chen and Boaz Barak and Borys Minaiev and Botao Hao and Bowen Baker and Brad Lightcap and Brandon McKinzie and Brandon Wang and Brendan Quinn and Brian Fioca and Brian Hsu and Brian Yang and Brian Yu and Brian Zhang and Brittany Brenner and Callie Riggins Zetino and Cameron Raymond and Camillo Lugaresi and Carolina Paz and Cary Hudson and Cedric Whitney and Chak Li and Charles Chen and Charlotte Cole and Chelsea Voss and Chen Ding and Chen Shen and Chengdu Huang and Chris Colby and Chris Hallacy and Chris Koch and Chris Lu and Christina Kaplan and Christina Kim and CJ Minott-Henriques and Cliff Frey and Cody Yu and Coley Czarnecki and Colin Reid and Colin Wei and Cory Decareaux and Cristina Scheau and Cyril Zhang and Cyrus Forbes and Da Tang and Dakota Goldberg and Dan Roberts and Dana Palmie and Daniel Kappler and Daniel Levine and Daniel Wright and Dave Leo and David Lin and David Robinson and Declan Grabb and Derek Chen and Derek Lim and Derek Salama and Dibya Bhattacharjee and Dimitris Tsipras and Dinghua Li and Dingli Yu and DJ Strouse and Drew Williams and Dylan Hunn and Ed Bayes and Edwin Arbus and Ekin Akyurek and Elaine Ya Le and Elana Widmann and Eli Yani and Elizabeth Proehl and Enis Sert and Enoch Cheung and Eri Schwartz and Eric Han and Eric Jiang and Eric Mitchell and Eric Sigler and Eric Wallace and Erik Ritter and Erin Kavanaugh and Evan Mays and Evgenii Nikishin and Fangyuan Li and Felipe Petroski Such and Filipe de Avila Belbute Peres and Filippo Raso and Florent Bekerman and Foivos Tsimpourlas and Fotis Chantzis and Francis Song and Francis Zhang and Gaby Raila and Garrett McGrath and Gary Briggs and Gary Yang and Giambattista Parascandolo and Gildas Chabot and Grace Kim and Grace Zhao and Gregory Valiant and Guillaume Leclerc and Hadi Salman and Hanson Wang and Hao Sheng and Haoming Jiang and Haoyu Wang and Haozhun Jin and Harshit Sikchi and Heather Schmidt and Henry Aspegren and Honglin Chen and Huida Qiu and Hunter Lightman and Ian Covert and Ian Kivlichan and Ian Silber and Ian Sohl and Ibrahim Hammoud and Ignasi Clavera and Ikai Lan and Ilge Akkaya and Ilya Kostrikov and Irina Kofman and Isak Etinger and Ishaan Singal and Jackie Hehir and Jacob Huh and Jacqueline Pan and Jake Wilczynski and Jakub Pachocki and James Lee and James Quinn and Jamie Kiros and Janvi Kalra and Jasmyn Samaroo and Jason Wang and Jason Wolfe and Jay Chen and Jay Wang and Jean Harb and Jeffrey Han and Jeffrey Wang and Jennifer Zhao and Jeremy Chen and Jerene Yang and Jerry Tworek and Jesse Chand and Jessica Landon and Jessica Liang and Ji Lin and Jiancheng Liu and Jianfeng Wang and Jie Tang and Jihan Yin and Joanne Jang and Joel Morris and Joey Flynn and Johannes Ferstad and Johannes Heidecke and John Fishbein and John Hallman and Jonah Grant and Jonathan Chien and Jonathan Gordon and Jongsoo Park and Jordan Liss and Jos Kraaijeveld and Joseph Guay and Joseph Mo and Josh Lawson and Josh McGrath and Joshua Vendrow and Joy Jiao and Julian Lee and Julie Steele and Julie Wang and Junhua Mao and Kai Chen and Kai Hayashi and Kai Xiao and Kamyar Salahi and Kan Wu and Karan Sekhri and Karan Sharma and Karan Singhal and Karen Li and Kenny Nguyen and Keren Gu-Lemberg and Kevin King and Kevin Liu and Kevin Stone and Kevin Yu and Kristen Ying and Kristian Georgiev and Kristie Lim and Kushal Tirumala and Kyle Miller and Lama Ahmad and Larry Lv and Laura Clare and Laurance Fauconnet and Lauren Itow and Lauren Yang and Laurentia Romaniuk and Leah Anise and Lee Byron and Leher Pathak and Leon Maksin and Leyan Lo and Leyton Ho and Li Jing and Liang Wu and Liang Xiong and Lien Mamitsuka and Lin Yang and Lindsay McCallum and Lindsey Held and Liz Bourgeois and Logan Engstrom and Lorenz Kuhn and Louis Feuvrier and Lu Zhang and Lucas Switzer and Lukas Kondraciuk and Lukasz Kaiser and Manas Joglekar and Mandeep Singh and Mandip Shah and Manuka Stratta and Marcus Williams and Mark Chen and Mark Sun and Marselus Cayton and Martin Li and Marvin Zhang and Marwan Aljubeh and Matt Nichols and Matthew Haines and Max Schwarzer and Mayank Gupta and Meghan Shah and Melody Y. Guan and Melody Huang and Meng Dong and Mengqing Wang and Mia Glaese and Micah Carroll and Michael Lampe and Michael Malek and Michael Sharman and Michael Zhang and Michele Wang and Michelle Pokrass and Mihai Florian and Mikhail Pavlov and Miles Wang and Ming Chen and Mingxuan Wang and Minnia Feng and Mo Bavarian and Molly Lin and Moose Abdool and Mostafa Rohaninejad and Nacho Soto and Natalie Staudacher and Natan LaFontaine and Nathan Marwell and Nelson Liu and Nick Preston and Nick Turley and Nicklas Ansman and Nicole Blades and Nikil Pancha and Nikita Mikhaylin and Niko Felix and Nikunj Handa and Nishant Rai and Nitish Keskar and Noam Brown and Ofir Nachum and Oleg Boiko and Oleg Murk and Olivia Watkins and Oona Gleeson and Pamela Mishkin and Patryk Lesiewicz and Paul Baltescu and Pavel Belov and Peter Zhokhov and Philip Pronin and Phillip Guo and Phoebe Thacker and Qi Liu and Qiming Yuan and Qinghua Liu and Rachel Dias and Rachel Puckett and Rahul Arora and Ravi Teja Mullapudi and Raz Gaon and Reah Miyara and Rennie Song and Rishabh Aggarwal and RJ Marsan and Robel Yemiru and Robert Xiong and Rohan Kshirsagar and Rohan Nuttall and Roman Tsiupa and Ronen Eldan and Rose Wang and Roshan James and Roy Ziv and Rui Shu and Ruslan Nigmatullin and Saachi Jain and Saam Talaie and Sam Altman and Sam Arnesen and Sam Toizer and Sam Toyer and Samuel Miserendino and Sandhini Agarwal and Sarah Yoo and Savannah Heon and Scott Ethersmith and Sean Grove and Sean Taylor and Sebastien Bubeck and Sever Banesiu and Shaokyi Amdo and Shengjia Zhao and Sherwin Wu and Shibani Santurkar and Shiyu Zhao and Shraman Ray Chaudhuri and Shreyas Krishnaswamy and Shuaiqi and Xia and Shuyang Cheng and Shyamal Anadkat and Simón Posada Fishman and Simon Tobin and Siyuan Fu and Somay Jain and Song Mei and Sonya Egoian and Spencer Kim and Spug Golden and SQ Mah and Steph Lin and Stephen Imm and Steve Sharpe and Steve Yadlowsky and Sulman Choudhry and Sungwon Eum and Suvansh Sanjeev and Tabarak Khan and Tal Stramer and Tao Wang and Tao Xin and Tarun Gogineni and Taya Christianson and Ted Sanders and Tejal Patwardhan and Thomas Degry and Thomas Shadwell and Tianfu Fu and Tianshi Gao and Timur Garipov and Tina Sriskandarajah and Toki Sherbakov and Tomek Korbak and Tomer Kaftan and Tomo Hiratsuka and Tongzhou Wang and Tony Song and Tony Zhao and Troy Peterson and Val Kharitonov and Victoria Chernova and Vineet Kosaraju and Vishal Kuo and Vitchyr Pong and Vivek Verma and Vlad Petrov and Wanning Jiang and Weixing Zhang and Wenda Zhou and Wenlei Xie and Wenting Zhan and Wes McCabe and Will DePue and Will Ellsworth and Wulfie Bain and Wyatt Thompson and Xiangning Chen and Xiangyu Qi and Xin Xiang and Xinwei Shi and Yann Dubois and Yaodong Yu and Yara Khakbaz and Yifan Wu and Yilei Qian and Yin Tat Lee and Yinbo Chen and Yizhen Zhang and Yizhong Xiong and Yonglong Tian and Young Cha and Yu Bai and Yu Yang and Yuan Yuan and Yuanzhi Li and Yufeng Zhang and Yuguang Yang and Yujia Jin and Yun Jiang and Yunyun Wang and Yushi Wang and Yutian Liu and Zach Stubenvoll and Zehao Dou and Zheng Wu and Zhigang Wang},
      year={2026},
      eprint={2601.03267},
      archivePrefix={arXiv},
      primaryClass={cs.CL},
      url={https://arxiv.org/abs/2601.03267}, 
}

@misc{grattafiori2024llama3herdmodels,
      title={The Llama 3 Herd of Models}, 
      author={Aaron Grattafiori and Abhimanyu Dubey and Abhinav Jauhri and Abhinav Pandey and Abhishek Kadian and Ahmad Al-Dahle and Aiesha Letman and Akhil Mathur and Alan Schelten and Alex Vaughan and Amy Yang and Angela Fan and Anirudh Goyal and Anthony Hartshorn and Aobo Yang and Archi Mitra and Archie Sravankumar and Artem Korenev and Arthur Hinsvark and Arun Rao and Aston Zhang and Aurelien Rodriguez and Austen Gregerson and Ava Spataru and Baptiste Roziere and Bethany Biron and Binh Tang and Bobbie Chern and Charlotte Caucheteux and Chaya Nayak and Chloe Bi and Chris Marra and Chris McConnell and Christian Keller and Christophe Touret and Chunyang Wu and Corinne Wong and Cristian Canton Ferrer and Cyrus Nikolaidis and Damien Allonsius and Daniel Song and Danielle Pintz and Danny Livshits and Danny Wyatt and David Esiobu and Dhruv Choudhary and Dhruv Mahajan and Diego Garcia-Olano and Diego Perino and Dieuwke Hupkes and Egor Lakomkin and Ehab AlBadawy and Elina Lobanova and Emily Dinan and Eric Michael Smith and Filip Radenovic and Francisco Guzmán and Frank Zhang and Gabriel Synnaeve and Gabrielle Lee and Georgia Lewis Anderson and Govind Thattai and Graeme Nail and Gregoire Mialon and Guan Pang and Guillem Cucurell and Hailey Nguyen and Hannah Korevaar and Hu Xu and Hugo Touvron and Iliyan Zarov and Imanol Arrieta Ibarra and Isabel Kloumann and Ishan Misra and Ivan Evtimov and Jack Zhang and Jade Copet and Jaewon Lee and Jan Geffert and Jana Vranes and Jason Park and Jay Mahadeokar and Jeet Shah and Jelmer van der Linde and Jennifer Billock and Jenny Hong and Jenya Lee and Jeremy Fu and Jianfeng Chi and Jianyu Huang and Jiawen Liu and Jie Wang and Jiecao Yu and Joanna Bitton and Joe Spisak and Jongsoo Park and Joseph Rocca and Joshua Johnstun and Joshua Saxe and Junteng Jia and Kalyan Vasuden Alwala and Karthik Prasad and Kartikeya Upasani and Kate Plawiak and Ke Li and Kenneth Heafield and Kevin Stone and Khalid El-Arini and Krithika Iyer and Kshitiz Malik and Kuenley Chiu and Kunal Bhalla and Kushal Lakhotia and Lauren Rantala-Yeary and Laurens van der Maaten and Lawrence Chen and Liang Tan and Liz Jenkins and Louis Martin and Lovish Madaan and Lubo Malo and Lukas Blecher and Lukas Landzaat and Luke de Oliveira and Madeline Muzzi and Mahesh Pasupuleti and Mannat Singh and Manohar Paluri and Marcin Kardas and Maria Tsimpoukelli and Mathew Oldham and Mathieu Rita and Maya Pavlova and Melanie Kambadur and Mike Lewis and Min Si and Mitesh Kumar Singh and Mona Hassan and Naman Goyal and Narjes Torabi and Nikolay Bashlykov and Nikolay Bogoychev and Niladri Chatterji and Ning Zhang and Olivier Duchenne and Onur Çelebi and Patrick Alrassy and Pengchuan Zhang and Pengwei Li and Petar Vasic and Peter Weng and Prajjwal Bhargava and Pratik Dubal and Praveen Krishnan and Punit Singh Koura and Puxin Xu and Qing He and Qingxiao Dong and Ragavan Srinivasan and Raj Ganapathy and Ramon Calderer and Ricardo Silveira Cabral and Robert Stojnic and Roberta Raileanu and Rohan Maheswari and Rohit Girdhar and Rohit Patel and Romain Sauvestre and Ronnie Polidoro and Roshan Sumbaly and Ross Taylor and Ruan Silva and Rui Hou and Rui Wang and Saghar Hosseini and Sahana Chennabasappa and Sanjay Singh and Sean Bell and Seohyun Sonia Kim and Sergey Edunov and Shaoliang Nie and Sharan Narang and Sharath Raparthy and Sheng Shen and Shengye Wan and Shruti Bhosale and Shun Zhang and Simon Vandenhende and Soumya Batra and Spencer Whitman and Sten Sootla and Stephane Collot and Suchin Gururangan and Sydney Borodinsky and Tamar Herman and Tara Fowler and Tarek Sheasha and Thomas Georgiou and Thomas Scialom and Tobias Speckbacher and Todor Mihaylov and Tong Xiao and Ujjwal Karn and Vedanuj Goswami and Vibhor Gupta and Vignesh Ramanathan and Viktor Kerkez and Vincent Gonguet and Virginie Do and Vish Vogeti and Vítor Albiero and Vladan Petrovic and Weiwei Chu and Wenhan Xiong and Wenyin Fu and Whitney Meers and Xavier Martinet and Xiaodong Wang and Xiaofang Wang and Xiaoqing Ellen Tan and Xide Xia and Xinfeng Xie and Xuchao Jia and Xuewei Wang and Yaelle Goldschlag and Yashesh Gaur and Yasmine Babaei and Yi Wen and Yiwen Song and Yuchen Zhang and Yue Li and Yuning Mao and Zacharie Delpierre Coudert and Zheng Yan and Zhengxing Chen and Zoe Papakipos and Aaditya Singh and Aayushi Srivastava and Abha Jain and Adam Kelsey and Adam Shajnfeld and Adithya Gangidi and Adolfo Victoria and Ahuva Goldstand and Ajay Menon and Ajay Sharma and Alex Boesenberg and Alexei Baevski and Allie Feinstein and Amanda Kallet and Amit Sangani and Amos Teo and Anam Yunus and Andrei Lupu and Andres Alvarado and Andrew Caples and Andrew Gu and Andrew Ho and Andrew Poulton and Andrew Ryan and Ankit Ramchandani and Annie Dong and Annie Franco and Anuj Goyal and Aparajita Saraf and Arkabandhu Chowdhury and Ashley Gabriel and Ashwin Bharambe and Assaf Eisenman and Azadeh Yazdan and Beau James and Ben Maurer and Benjamin Leonhardi and Bernie Huang and Beth Loyd and Beto De Paola and Bhargavi Paranjape and Bing Liu and Bo Wu and Boyu Ni and Braden Hancock and Bram Wasti and Brandon Spence and Brani Stojkovic and Brian Gamido and Britt Montalvo and Carl Parker and Carly Burton and Catalina Mejia and Ce Liu and Changhan Wang and Changkyu Kim and Chao Zhou and Chester Hu and Ching-Hsiang Chu and Chris Cai and Chris Tindal and Christoph Feichtenhofer and Cynthia Gao and Damon Civin and Dana Beaty and Daniel Kreymer and Daniel Li and David Adkins and David Xu and Davide Testuggine and Delia David and Devi Parikh and Diana Liskovich and Didem Foss and Dingkang Wang and Duc Le and Dustin Holland and Edward Dowling and Eissa Jamil and Elaine Montgomery and Eleonora Presani and Emily Hahn and Emily Wood and Eric-Tuan Le and Erik Brinkman and Esteban Arcaute and Evan Dunbar and Evan Smothers and Fei Sun and Felix Kreuk and Feng Tian and Filippos Kokkinos and Firat Ozgenel and Francesco Caggioni and Frank Kanayet and Frank Seide and Gabriela Medina Florez and Gabriella Schwarz and Gada Badeer and Georgia Swee and Gil Halpern and Grant Herman and Grigory Sizov and Guangyi and Zhang and Guna Lakshminarayanan and Hakan Inan and Hamid Shojanazeri and Han Zou and Hannah Wang and Hanwen Zha and Haroun Habeeb and Harrison Rudolph and Helen Suk and Henry Aspegren and Hunter Goldman and Hongyuan Zhan and Ibrahim Damlaj and Igor Molybog and Igor Tufanov and Ilias Leontiadis and Irina-Elena Veliche and Itai Gat and Jake Weissman and James Geboski and James Kohli and Janice Lam and Japhet Asher and Jean-Baptiste Gaya and Jeff Marcus and Jeff Tang and Jennifer Chan and Jenny Zhen and Jeremy Reizenstein and Jeremy Teboul and Jessica Zhong and Jian Jin and Jingyi Yang and Joe Cummings and Jon Carvill and Jon Shepard and Jonathan McPhie and Jonathan Torres and Josh Ginsburg and Junjie Wang and Kai Wu and Kam Hou U and Karan Saxena and Kartikay Khandelwal and Katayoun Zand and Kathy Matosich and Kaushik Veeraraghavan and Kelly Michelena and Keqian Li and Kiran Jagadeesh and Kun Huang and Kunal Chawla and Kyle Huang and Lailin Chen and Lakshya Garg and Lavender A and Leandro Silva and Lee Bell and Lei Zhang and Liangpeng Guo and Licheng Yu and Liron Moshkovich and Luca Wehrstedt and Madian Khabsa and Manav Avalani and Manish Bhatt and Martynas Mankus and Matan Hasson and Matthew Lennie and Matthias Reso and Maxim Groshev and Maxim Naumov and Maya Lathi and Meghan Keneally and Miao Liu and Michael L. Seltzer and Michal Valko and Michelle Restrepo and Mihir Patel and Mik Vyatskov and Mikayel Samvelyan and Mike Clark and Mike Macey and Mike Wang and Miquel Jubert Hermoso and Mo Metanat and Mohammad Rastegari and Munish Bansal and Nandhini Santhanam and Natascha Parks and Natasha White and Navyata Bawa and Nayan Singhal and Nick Egebo and Nicolas Usunier and Nikhil Mehta and Nikolay Pavlovich Laptev and Ning Dong and Norman Cheng and Oleg Chernoguz and Olivia Hart and Omkar Salpekar and Ozlem Kalinli and Parkin Kent and Parth Parekh and Paul Saab and Pavan Balaji and Pedro Rittner and Philip Bontrager and Pierre Roux and Piotr Dollar and Polina Zvyagina and Prashant Ratanchandani and Pritish Yuvraj and Qian Liang and Rachad Alao and Rachel Rodriguez and Rafi Ayub and Raghotham Murthy and Raghu Nayani and Rahul Mitra and Rangaprabhu Parthasarathy and Raymond Li and Rebekkah Hogan and Robin Battey and Rocky Wang and Russ Howes and Ruty Rinott and Sachin Mehta and Sachin Siby and Sai Jayesh Bondu and Samyak Datta and Sara Chugh and Sara Hunt and Sargun Dhillon and Sasha Sidorov and Satadru Pan and Saurabh Mahajan and Saurabh Verma and Seiji Yamamoto and Sharadh Ramaswamy and Shaun Lindsay and Shaun Lindsay and Sheng Feng and Shenghao Lin and Shengxin Cindy Zha and Shishir Patil and Shiva Shankar and Shuqiang Zhang and Shuqiang Zhang and Sinong Wang and Sneha Agarwal and Soji Sajuyigbe and Soumith Chintala and Stephanie Max and Stephen Chen and Steve Kehoe and Steve Satterfield and Sudarshan Govindaprasad and Sumit Gupta and Summer Deng and Sungmin Cho and Sunny Virk and Suraj Subramanian and Sy Choudhury and Sydney Goldman and Tal Remez and Tamar Glaser and Tamara Best and Thilo Koehler and Thomas Robinson and Tianhe Li and Tianjun Zhang and Tim Matthews and Timothy Chou and Tzook Shaked and Varun Vontimitta and Victoria Ajayi and Victoria Montanez and Vijai Mohan and Vinay Satish Kumar and Vishal Mangla and Vlad Ionescu and Vlad Poenaru and Vlad Tiberiu Mihailescu and Vladimir Ivanov and Wei Li and Wenchen Wang and Wenwen Jiang and Wes Bouaziz and Will Constable and Xiaocheng Tang and Xiaojian Wu and Xiaolan Wang and Xilun Wu and Xinbo Gao and Yaniv Kleinman and Yanjun Chen and Ye Hu and Ye Jia and Ye Qi and Yenda Li and Yilin Zhang and Ying Zhang and Yossi Adi and Youngjin Nam and Yu and Wang and Yu Zhao and Yuchen Hao and Yundi Qian and Yunlu Li and Yuzi He and Zach Rait and Zachary DeVito and Zef Rosnbrick and Zhaoduo Wen and Zhenyu Yang and Zhiwei Zhao and Zhiyu Ma},
      year={2024},
      eprint={2407.21783},
      archivePrefix={arXiv},
      primaryClass={cs.AI},
      url={https://arxiv.org/abs/2407.21783}, 
}

@inproceedings{belz20242024,
  title={The 2024 repronlp shared task on reproducibility of evaluations in nlp: Overview and results},
  author={Belz, Anja and Thomson, Craig},
  booktitle={Proceedings of the Fourth Workshop on Human Evaluation of NLP Systems (HumEval)@ LREC-COLING 2024},
  pages={91--105},
  year={2024}
}

@inproceedings{belz2023missing,
  title={Missing information, unresponsive authors, experimental flaws: The impossibility of assessing the reproducibility of previous human evaluations in NLP},
  author={Belz, Anja and Thomson, Craig and Reiter, Ehud and Abercrombie, Gavin and Alonso-Moral, Jose M and Arvan, Mohammad and Braggaar, Anouck and Cieliebak, Mark and Clark, Elizabeth and Van Deemter, Kees and others},
  booktitle={Proceedings of the Fourth Workshop on Insights from Negative Results in NLP},
  pages={1--10},
  year={2023}
}

@article{bayerl-paul-2011-determines,
    title = "What Determines Inter-Coder Agreement in Manual Annotations? A Meta-Analytic Investigation",
    author = "Bayerl, Petra Saskia  and
      Paul, Karsten Ingmar",
    journal = "Computational Linguistics",
    volume = "37",
    number = "4",
    month = dec,
    year = "2011",
    address = "Cambridge, MA",
    publisher = "MIT Press",
    url = "https://aclanthology.org/J11-4004/",
    doi = "10.1162/COLI_a_00074",
    pages = "699--725"
}

@article{klie-etal-2024-analyzing,
    title = "Analyzing Dataset Annotation Quality Management in the Wild",
    author = "Klie, Jan-Christoph  and
      Eckart de Castilho, Richard  and
      Gurevych, Iryna",
    journal = "Computational Linguistics",
    volume = "50",
    number = "3",
    month = sep,
    year = "2024",
    address = "Cambridge, MA",
    publisher = "MIT Press",
    url = "https://aclanthology.org/2024.cl-3.1/",
    doi = "10.1162/coli_a_00516",
    pages = "817--866"
}

@inproceedings{pei-jurgens-2023-annotator,
    title = "When Do Annotator Demographics Matter? Measuring the Influence of Annotator Demographics with the {POPQUORN} Dataset",
    author = "Pei, Jiaxin  and
      Jurgens, David",
    editor = "Prange, Jakob  and
      Friedrich, Annemarie",
    booktitle = "Proceedings of the 17th Linguistic Annotation Workshop (LAW-XVII)",
    month = jul,
    year = "2023",
    address = "Toronto, Canada",
    publisher = "Association for Computational Linguistics",
    url = "https://aclanthology.org/2023.law-1.25/",
    doi = "10.18653/v1/2023.law-1.25",
    pages = "252--265"
}

@article{Beck2023,
  author    = {Jacob Beck},
  title     = {Quality aspects of annotated data},
  journal   = {AStA Wirtschafts- und Sozialstatistisches Archiv},
  year      = {2023},
  volume    = {17},
  number    = {3},
  pages     = {331--353},
  doi       = {10.1007/s11943-023-00332-y},
  url       = {https://doi.org/10.1007/s11943-023-00332-y},
  issn      = {1863-8163},
  month     = dec
}

@misc{tyser2024aidrivenreviewsystemsevaluating,
      title={AI-Driven Review Systems: Evaluating LLMs in Scalable and Bias-Aware Academic Reviews}, 
      author={Keith Tyser and Ben Segev and Gaston Longhitano and Xin-Yu Zhang and Zachary Meeks and Jason Lee and Uday Garg and Nicholas Belsten and Avi Shporer and Madeleine Udell and Dov Te'eni and Iddo Drori},
      year={2024},
      eprint={2408.10365},
      archivePrefix={arXiv},
      primaryClass={cs.AI},
      url={https://arxiv.org/abs/2408.10365}, 
}

@article{10.1002/pra2.1323,
author = {Kim, Yumi and Lee, Jongwook and Yang, Seungwon},
title = {Exploring LLM AI in Automatic Generation of Abstracts for Research Publications},
year = {2025},
issue_date = {October 2025},
publisher = {John Wiley \& Sons, Inc.},
address = {USA},
volume = {62},
number = {1},
url = {https://doi.org/10.1002/pra2.1323},
doi = {10.1002/pra2.1323},
journal = {Proceedings of the Association for Information Science and Technology},
month = oct,
pages = {967–971},
numpages = {5}
}

@misc{chen2025ai4researchsurveyartificialintelligence,
      title={AI4Research: A Survey of Artificial Intelligence for Scientific Research}, 
      author={Qiguang Chen and Mingda Yang and Libo Qin and Jinhao Liu and Zheng Yan and Jiannan Guan and Dengyun Peng and Yiyan Ji and Hanjing Li and Mengkang Hu and Yimeng Zhang and Yihao Liang and Yuhang Zhou and Jiaqi Wang and Zhi Chen and Wanxiang Che},
      year={2025},
      eprint={2507.01903},
      archivePrefix={arXiv},
      primaryClass={cs.CL},
      url={https://arxiv.org/abs/2507.01903}, 
}

@inproceedings{jiang-etal-2024-examining,
    title = "Re-examining Sexism and Misogyny Classification with Annotator Attitudes",
    author = "Jiang, Aiqi  and
      Vitsakis, Nikolas  and
      Dinkar, Tanvi  and
      Abercrombie, Gavin  and
      Konstas, Ioannis",
    editor = "Al-Onaizan, Yaser  and
      Bansal, Mohit  and
      Chen, Yun-Nung",
    booktitle = "Findings of the Association for Computational Linguistics: EMNLP 2024",
    month = nov,
    year = "2024",
    address = "Miami, Florida, USA",
    publisher = "Association for Computational Linguistics",
    url = "https://aclanthology.org/2024.findings-emnlp.887/",
    doi = "10.18653/v1/2024.findings-emnlp.887",
    pages = "15103--15125"
}

@misc{Gemini3,
    title = {A new era of intelligence with {G}emini 3},
    url = {https://blog.google/products/gemini/gemini-3/},
    author = {{Gemini Team}},
    month = {November},
    year = {2025}
}

@misc{GPT-5,
    title = {Introducing {GPT}‑5},
    url = {https://openai.com/index/introducing-gpt-5/},
    author = {{OpenAI}},
    month = {August},
    year = {2025}
}

@misc{qwen3.6-27b,
    title  = {{Qwen3.6-27B}: Flagship-Level Coding in a {27B} Dense Model},
    author = {{Qwen Team}},
    month  = {April},
    year   = {2026},
    url    = {https://qwen.ai/blog?id=qwen3.6-27b}
}

@misc{openai2025gptoss120bgptoss20bmodel,
      title={gpt-oss-120b \& gpt-oss-20b Model Card}, 
      author={OpenAI},
      year={2025},
      eprint={2508.10925},
      archivePrefix={arXiv},
      primaryClass={cs.CL},
      url={https://arxiv.org/abs/2508.10925}, 
}

@misc{gemma4,
    author={Google},
    year = {2026},
	title = {{Gemma 4 Model Card}},
	url = {https://ai.google.dev/gemma/docs/core/model_card_4},
}

@inproceedings{van-der-lee-etal-2019-best,
    title = "Best practices for the human evaluation of automatically generated text",
    author = "van der Lee, Chris  and
      Gatt, Albert  and
      van Miltenburg, Emiel  and
      Wubben, Sander  and
      Krahmer, Emiel",
    editor = "van Deemter, Kees  and
      Lin, Chenghua  and
      Takamura, Hiroya",
    booktitle = "Proceedings of the 12th International Conference on Natural Language Generation",
    month = oct # "–" # nov,
    year = "2019",
    address = "Tokyo, Japan",
    publisher = "Association for Computational Linguistics",
    url = "https://aclanthology.org/W19-8643/",
    doi = "10.18653/v1/W19-8643",
    pages = "355--368"
}

@article{bender-friedman-2018-data,
    title = "Data Statements for Natural Language Processing: Toward Mitigating System Bias and Enabling Better Science",
    author = "Bender, Emily M.  and
      Friedman, Batya",
    editor = "Lee, Lillian  and
      Johnson, Mark  and
      Toutanova, Kristina  and
      Roark, Brian",
    journal = "Transactions of the Association for Computational Linguistics",
    volume = "6",
    year = "2018",
    address = "Cambridge, MA",
    publisher = "MIT Press",
    url = "https://aclanthology.org/Q18-1041/",
    doi = "10.1162/tacl_a_00041",
    pages = "587--604"
}

@inproceedings{joshi-etal-2020-state,
    title = "The State and Fate of Linguistic Diversity and Inclusion in the {NLP} World",
    author = "Joshi, Pratik  and
      Santy, Sebastin  and
      Budhiraja, Amar  and
      Bali, Kalika  and
      Choudhury, Monojit",
    editor = "Jurafsky, Dan  and
      Chai, Joyce  and
      Schluter, Natalie  and
      Tetreault, Joel",
    booktitle = "Proceedings of the 58th Annual Meeting of the Association for Computational Linguistics",
    month = jul,
    year = "2020",
    address = "Online",
    publisher = "Association for Computational Linguistics",
    url = "https://aclanthology.org/2020.acl-main.560/",
    doi = "10.18653/v1/2020.acl-main.560",
    pages = "6282--6293"
}

@misc{aclrollingreview_responsiblenlp, 
title={ACL Rolling Review A peer review platform for the Association for Computational Linguistics},
url={https://aclrollingreview.org/responsibleNLPresearch/}, 
journal={ACL Rolling Review}, 
author={Rolling Review, ACL},
year={2022}
}

@inproceedings{zeinert-etal-2021-annotating,
    title = "Annotating Online Misogyny",
    author = "Zeinert, Philine  and
      Inie, Nanna  and
      Derczynski, Leon",
    editor = "Zong, Chengqing  and
      Xia, Fei  and
      Li, Wenjie  and
      Navigli, Roberto",
    booktitle = "Proceedings of the 59th Annual Meeting of the Association for Computational Linguistics and the 11th International Joint Conference on Natural Language Processing (Volume 1: Long Papers)",
    month = aug,
    year = "2021",
    address = "Online",
    publisher = "Association for Computational Linguistics",
    url = "https://aclanthology.org/2021.acl-long.247/",
    doi = "10.18653/v1/2021.acl-long.247",
    pages = "3181--3197"
}

@inproceedings{zhang-etal-2021-emailsum,
    title = "{E}mail{S}um: Abstractive Email Thread Summarization",
    author = "Zhang, Shiyue  and
      Celikyilmaz, Asli  and
      Gao, Jianfeng  and
      Bansal, Mohit",
    editor = "Zong, Chengqing  and
      Xia, Fei  and
      Li, Wenjie  and
      Navigli, Roberto",
    booktitle = "Proceedings of the 59th Annual Meeting of the Association for Computational Linguistics and the 11th International Joint Conference on Natural Language Processing (Volume 1: Long Papers)",
    month = aug,
    year = "2021",
    address = "Online",
    publisher = "Association for Computational Linguistics",
    url = "https://aclanthology.org/2021.acl-long.537/",
    doi = "10.18653/v1/2021.acl-long.537",
    pages = "6895--6909"
}

@inproceedings{haber-etal-2023-improving,
    title = "Improving the Detection of Multilingual Online Attacks with Rich Social Media Data from {S}ingapore",
    author = {Haber, Janosch  and
      Vidgen, Bertie  and
      Chapman, Matthew  and
      Agarwal, Vibhor  and
      Lee, Roy Ka-Wei  and
      Yap, Yong Keong  and
      R{\"o}ttger, Paul},
    editor = "Rogers, Anna  and
      Boyd-Graber, Jordan  and
      Okazaki, Naoaki",
    booktitle = "Proceedings of the 61st Annual Meeting of the Association for Computational Linguistics (Volume 1: Long Papers)",
    month = jul,
    year = "2023",
    address = "Toronto, Canada",
    publisher = "Association for Computational Linguistics",
    url = "https://aclanthology.org/2023.acl-long.711/",
    doi = "10.18653/v1/2023.acl-long.711",
    pages = "12705--12721"
}

@inproceedings{doneva-etal-2024-neurotrialner,
    title = "{N}euro{T}rial{NER}: An Annotated Corpus for Neurological Diseases and Therapies in Clinical Trial Registries",
    author = "Doneva, Simona Emilova  and
      Ellendorff, Tilia  and
      Sick, Beate  and
      Goldman, Jean-Philippe  and
      Cannon, Amelia Elaine  and
      Schneider, Gerold  and
      Ineichen, Benjamin Victor",
    editor = "Al-Onaizan, Yaser  and
      Bansal, Mohit  and
      Chen, Yun-Nung",
    booktitle = "Proceedings of the 2024 Conference on Empirical Methods in Natural Language Processing",
    month = nov,
    year = "2024",
    address = "Miami, Florida, USA",
    publisher = "Association for Computational Linguistics",
    url = "https://aclanthology.org/2024.emnlp-main.1050/",
    doi = "10.18653/v1/2024.emnlp-main.1050",
    pages = "18868--18890"
}

@article{ziems-etal-2024-large,
    title = "Can Large Language Models Transform Computational Social Science?",
    author = "Ziems, Caleb  and
      Held, William  and
      Shaikh, Omar  and
      Chen, Jiaao  and
      Zhang, Zhehao  and
      Yang, Diyi",
    journal = "Computational Linguistics",
    volume = "50",
    number = "1",
    month = mar,
    year = "2024",
    address = "Cambridge, MA",
    publisher = "MIT Press",
    url = "https://aclanthology.org/2024.cl-1.8/",
    doi = "10.1162/coli_a_00502",
    pages = "237--291"
}

@inproceedings{zhao-etal-2024-comprehensive,
    title = "A Comprehensive Study of Gender Bias in Chemical Named Entity Recognition Models",
    author = "Zhao, Xingmeng  and
      Niazi, Ali  and
      Rios, Anthony",
    editor = "Duh, Kevin  and
      Gomez, Helena  and
      Bethard, Steven",
    booktitle = "Proceedings of the 2024 Conference of the North American Chapter of the Association for Computational Linguistics: Human Language Technologies (Volume 1: Long Papers)",
    month = jun,
    year = "2024",
    address = "Mexico City, Mexico",
    publisher = "Association for Computational Linguistics",
    url = "https://aclanthology.org/2024.naacl-long.245/",
    doi = "10.18653/v1/2024.naacl-long.245",
    pages = "4360--4374"
}

@misc{li2024contradocunderstandingselfcontradictionsdocuments,
      title={ContraDoc: Understanding Self-Contradictions in Documents with Large Language Models}, 
      author={Jierui Li and Vipul Raheja and Dhruv Kumar},
      year={2024},
      eprint={2311.09182},
      archivePrefix={arXiv},
      primaryClass={cs.CL},
      url={https://arxiv.org/abs/2311.09182}, 
}

@misc{belouadi2024detikzifysynthesizinggraphicsprograms,
      title={DeTikZify: Synthesizing Graphics Programs for Scientific Figures and Sketches with TikZ}, 
      author={Jonas Belouadi and Simone Paolo Ponzetto and Steffen Eger},
      year={2024},
      eprint={2405.15306},
      archivePrefix={arXiv},
      primaryClass={cs.CL},
      url={https://arxiv.org/abs/2405.15306}, 
}

@inproceedings{ide-etal-2025-coam,
    title = "{C}o{AM}: Corpus of All-Type Multiword Expressions",
    author = "Ide, Yusuke  and
      Tanner, Joshua  and
      Nohejl, Adam  and
      Hoffman, Jacob  and
      Vasselli, Justin  and
      Kamigaito, Hidetaka  and
      Watanabe, Taro",
    editor = "Che, Wanxiang  and
      Nabende, Joyce  and
      Shutova, Ekaterina  and
      Pilehvar, Mohammad Taher",
    booktitle = "Proceedings of the 63rd Annual Meeting of the Association for Computational Linguistics (Volume 1: Long Papers)",
    month = jul,
    year = "2025",
    address = "Vienna, Austria",
    publisher = "Association for Computational Linguistics",
    url = "https://aclanthology.org/2025.acl-long.1311/",
    doi = "10.18653/v1/2025.acl-long.1311",
    pages = "27004--27021",
    ISBN = "979-8-89176-251-0"
}

@misc{altemeyer2025argumentsummarizationevaluationera,
      title={Argument Summarization and its Evaluation in the Era of Large Language Models}, 
      author={Moritz Altemeyer and Steffen Eger and Johannes Daxenberger and Yanran Chen and Tim Altendorf and Philipp Cimiano and Benjamin Schiller},
      year={2025},
      eprint={2503.00847},
      archivePrefix={arXiv},
      primaryClass={cs.CL},
      url={https://arxiv.org/abs/2503.00847}, 
}

@misc{belouadi2025tikzerozeroshottextguidedgraphics,
      title={TikZero: Zero-Shot Text-Guided Graphics Program Synthesis}, 
      author={Jonas Belouadi and Eddy Ilg and Margret Keuper and Hideki Tanaka and Masao Utiyama and Raj Dabre and Steffen Eger and Simone Paolo Ponzetto},
      year={2025},
      eprint={2503.11509},
      archivePrefix={arXiv},
      primaryClass={cs.CL},
      url={https://arxiv.org/abs/2503.11509}, 
}

@misc{greisinger2026tikzillascalingtexttotikzhighquality,
      title={TikZilla: Scaling Text-to-TikZ with High-Quality Data and Reinforcement Learning}, 
      author={Christian Greisinger and Steffen Eger},
      year={2026},
      eprint={2603.03072},
      archivePrefix={arXiv},
      primaryClass={cs.AI},
      url={https://arxiv.org/abs/2603.03072}, 
}

@inproceedings{sai-b-etal-2023-indicmt,
    title = "{I}ndic{MT} Eval: A Dataset to Meta-Evaluate Machine Translation Metrics for {I}ndian Languages",
    author = "Sai B, Ananya  and
      Dixit, Tanay  and
      Nagarajan, Vignesh  and
      Kunchukuttan, Anoop  and
      Kumar, Pratyush  and
      Khapra, Mitesh M.  and
      Dabre, Raj",
    editor = "Rogers, Anna  and
      Boyd-Graber, Jordan  and
      Okazaki, Naoaki",
    booktitle = "Proceedings of the 61st Annual Meeting of the Association for Computational Linguistics (Volume 1: Long Papers)",
    month = jul,
    year = "2023",
    address = "Toronto, Canada",
    publisher = "Association for Computational Linguistics",
    url = "https://aclanthology.org/2023.acl-long.795/",
    doi = "10.18653/v1/2023.acl-long.795",
    pages = "14210--14228"
}

@inproceedings{agnew-etal-2023-mechanical,
    title = "The Mechanical Bard: An Interpretable Machine Learning Approach to {S}hakespearean Sonnet Generation",
    author = "Agnew, Edwin  and
      Qiu, Michelle  and
      Zhu, Lily  and
      Wiseman, Sam  and
      Rudin, Cynthia",
    editor = "Rogers, Anna  and
      Boyd-Graber, Jordan  and
      Okazaki, Naoaki",
    booktitle = "Proceedings of the 61st Annual Meeting of the Association for Computational Linguistics (Volume 2: Short Papers)",
    month = jul,
    year = "2023",
    address = "Toronto, Canada",
    publisher = "Association for Computational Linguistics",
    url = "https://aclanthology.org/2023.acl-short.140/",
    doi = "10.18653/v1/2023.acl-short.140",
    pages = "1627--1638"
}

@inproceedings{lin-etal-2023-argue,
    title = "Argue with Me Tersely: Towards Sentence-Level Counter-Argument Generation",
    author = "Lin, Jiayu  and
      Ye, Rong  and
      Han, Meng  and
      Zhang, Qi  and
      Lai, Ruofei  and
      Zhang, Xinyu  and
      Cao, Zhao  and
      Huang, Xuanjing  and
      Wei, Zhongyu",
    editor = "Bouamor, Houda  and
      Pino, Juan  and
      Bali, Kalika",
    booktitle = "Proceedings of the 2023 Conference on Empirical Methods in Natural Language Processing",
    month = dec,
    year = "2023",
    address = "Singapore",
    publisher = "Association for Computational Linguistics",
    url = "https://aclanthology.org/2023.emnlp-main.1039/",
    doi = "10.18653/v1/2023.emnlp-main.1039",
    pages = "16705--16720"
}

@inproceedings{mathur-etal-2025-social,
    title = "Social Genome: Grounded Social Reasoning Abilities of Multimodal Models",
    author = "Mathur, Leena  and
      Qian, Marian  and
      Liang, Paul Pu  and
      Morency, Louis-Philippe",
    editor = "Christodoulopoulos, Christos  and
      Chakraborty, Tanmoy  and
      Rose, Carolyn  and
      Peng, Violet",
    booktitle = "Proceedings of the 2025 Conference on Empirical Methods in Natural Language Processing",
    month = nov,
    year = "2025",
    address = "Suzhou, China",
    publisher = "Association for Computational Linguistics",
    url = "https://aclanthology.org/2025.emnlp-main.1264/",
    doi = "10.18653/v1/2025.emnlp-main.1264",
    pages = "24868--24891",
    ISBN = "979-8-89176-332-6"
}

@misc{ivetta2025heseiacommunitybaseddatasetevaluating,
      title={HESEIA: A community-based dataset for evaluating social biases in large language models, co-designed in real school settings in Latin America}, 
      author={Guido Ivetta and Marcos J. Gomez and Sofía Martinelli and Pietro Palombini and M. Emilia Echeveste and Nair Carolina Mazzeo and Beatriz Busaniche and Luciana Benotti},
      year={2025},
      eprint={2505.24712},
      archivePrefix={arXiv},
      primaryClass={cs.CL},
      url={https://arxiv.org/abs/2505.24712}, 
}

@misc{huang2025visbiasmeasuringexplicitimplicit,
      title={VisBias: Measuring Explicit and Implicit Social Biases in Vision Language Models}, 
      author={Jen-tse Huang and Jiantong Qin and Jianping Zhang and Youliang Yuan and Wenxuan Wang and Jieyu Zhao},
      year={2025},
      eprint={2503.07575},
      archivePrefix={arXiv},
      primaryClass={cs.CV},
      url={https://arxiv.org/abs/2503.07575}, 
}

@inproceedings{chen-etal-2025-benchmarking-llms,
    title = "Benchmarking {LLM}s for Translating Classical {C}hinese Poetry: Evaluating Adequacy, Fluency, and Elegance",
    author = "Chen, Andong  and
      Lou, Lianzhang  and
      Chen, Kehai  and
      Bai, Xuefeng  and
      Xiang, Yang  and
      Yang, Muyun  and
      Zhao, Tiejun  and
      Zhang, Min",
    editor = "Christodoulopoulos, Christos  and
      Chakraborty, Tanmoy  and
      Rose, Carolyn  and
      Peng, Violet",
    booktitle = "Proceedings of the 2025 Conference on Empirical Methods in Natural Language Processing",
    month = nov,
    year = "2025",
    address = "Suzhou, China",
    publisher = "Association for Computational Linguistics",
    url = "https://aclanthology.org/2025.emnlp-main.1678/",
    doi = "10.18653/v1/2025.emnlp-main.1678",
    pages = "33019--33036",
    ISBN = "979-8-89176-332-6"
}

@inproceedings{kostikova-etal-2024-fine,
    title = "Fine-Grained Detection of Solidarity for Women and Migrants in 155 Years of {G}erman Parliamentary Debates",
    author = {Kostikova, Aida  and
      Paassen, Benjamin  and
      Beese, Dominik  and
      P{\"u}tz, Ole  and
      Wiedemann, Gregor  and
      Eger, Steffen},
    editor = "Al-Onaizan, Yaser  and
      Bansal, Mohit  and
      Chen, Yun-Nung",
    booktitle = "Proceedings of the 2024 Conference on Empirical Methods in Natural Language Processing",
    month = nov,
    year = "2024",
    address = "Miami, Florida, USA",
    publisher = "Association for Computational Linguistics",
    url = "https://aclanthology.org/2024.emnlp-main.337/",
    doi = "10.18653/v1/2024.emnlp-main.337",
    pages = "5884--5907"
}

@misc{walsh2024sonnetnotbotpoetry,
      title={Sonnet or Not, Bot? Poetry Evaluation for Large Models and Datasets}, 
      author={Melanie Walsh and Anna Preus and Maria Antoniak},
      year={2024},
      eprint={2406.18906},
      archivePrefix={arXiv},
      primaryClass={cs.CL},
      url={https://arxiv.org/abs/2406.18906}, 
}

@inproceedings{chakrabarty-etal-2021-dont,
    title = "Don{'}t Go Far Off: An Empirical Study on Neural Poetry Translation",
    author = "Chakrabarty, Tuhin  and
      Saakyan, Arkadiy  and
      Muresan, Smaranda",
    editor = "Moens, Marie-Francine  and
      Huang, Xuanjing  and
      Specia, Lucia  and
      Yih, Scott Wen-tau",
    booktitle = "Proceedings of the 2021 Conference on Empirical Methods in Natural Language Processing",
    month = nov,
    year = "2021",
    address = "Online and Punta Cana, Dominican Republic",
    publisher = "Association for Computational Linguistics",
    url = "https://aclanthology.org/2021.emnlp-main.577/",
    doi = "10.18653/v1/2021.emnlp-main.577",
    pages = "7253--7265"
}

@inproceedings{zou-2025-bipro,
    title = "{BIP}ro: Zero-shot {C}hinese Poem Generation via Block Inverse Prompting Constrained Generation Framework",
    author = "Zou, Xu",
    editor = "Che, Wanxiang  and
      Nabende, Joyce  and
      Shutova, Ekaterina  and
      Pilehvar, Mohammad Taher",
    booktitle = "Proceedings of the 63rd Annual Meeting of the Association for Computational Linguistics (Volume 1: Long Papers)",
    month = jul,
    year = "2025",
    address = "Vienna, Austria",
    publisher = "Association for Computational Linguistics",
    url = "https://aclanthology.org/2025.acl-long.56/",
    doi = "10.18653/v1/2025.acl-long.56",
    pages = "1116--1134",
    ISBN = "979-8-89176-251-0"
}

@inproceedings{el-kheir-etal-2024-beyond,
    title = "Beyond Orthography: Automatic Recovery of Short Vowels and Dialectal Sounds in {A}rabic",
    author = "El Kheir, Yassine  and
      Mubarak, Hamdy  and
      Ali, Ahmed  and
      Chowdhury, Shammur",
    editor = "Ku, Lun-Wei  and
      Martins, Andre  and
      Srikumar, Vivek",
    booktitle = "Proceedings of the 62nd Annual Meeting of the Association for Computational Linguistics (Volume 1: Long Papers)",
    month = aug,
    year = "2024",
    address = "Bangkok, Thailand",
    publisher = "Association for Computational Linguistics",
    url = "https://aclanthology.org/2024.acl-long.711/",
    doi = "10.18653/v1/2024.acl-long.711",
    pages = "13172--13184"
}

@inproceedings{breit-etal-2021-wic,
    title = "{WiC-TSV}: {A}n Evaluation Benchmark for Target Sense Verification of Words in Context",
    author = "Breit, Anna  and
      Revenko, Artem  and
      Rezaee, Kiamehr  and
      Pilehvar, Mohammad Taher  and
      Camacho-Collados, Jose",
    editor = "Merlo, Paola  and
      Tiedemann, Jorg  and
      Tsarfaty, Reut",
    booktitle = "Proceedings of the 16th Conference of the European Chapter of the Association for Computational Linguistics: Main Volume",
    month = apr,
    year = "2021",
    address = "Online",
    publisher = "Association for Computational Linguistics",
    url = "https://aclanthology.org/2021.eacl-main.140/",
    doi = "10.18653/v1/2021.eacl-main.140",
    pages = "1635--1645"
}

@inproceedings{jiang-riloff-2021-learning,
    title = "Learning Prototypical Functions for Physical Artifacts",
    author = "Jiang, Tianyu  and
      Riloff, Ellen",
    editor = "Zong, Chengqing  and
      Xia, Fei  and
      Li, Wenjie  and
      Navigli, Roberto",
    booktitle = "Proceedings of the 59th Annual Meeting of the Association for Computational Linguistics and the 11th International Joint Conference on Natural Language Processing (Volume 1: Long Papers)",
    month = aug,
    year = "2021",
    address = "Online",
    publisher = "Association for Computational Linguistics",
    url = "https://aclanthology.org/2021.acl-long.540/",
    doi = "10.18653/v1/2021.acl-long.540",
    pages = "6941--6951"
}

@inproceedings{asai-choi-2021-challenges,
    title = "Challenges in Information-Seeking {QA}: Unanswerable Questions and Paragraph Retrieval",
    author = "Asai, Akari  and
      Choi, Eunsol",
    editor = "Zong, Chengqing  and
      Xia, Fei  and
      Li, Wenjie  and
      Navigli, Roberto",
    booktitle = "Proceedings of the 59th Annual Meeting of the Association for Computational Linguistics and the 11th International Joint Conference on Natural Language Processing (Volume 1: Long Papers)",
    month = aug,
    year = "2021",
    address = "Online",
    publisher = "Association for Computational Linguistics",
    url = "https://aclanthology.org/2021.acl-long.118/",
    doi = "10.18653/v1/2021.acl-long.118",
    pages = "1492--1504"
}

@inproceedings{wang-etal-2020-learning-efficient,
    title = "Learning Efficient Dialogue Policy from Demonstrations through Shaping",
    author = "Wang, Huimin  and
      Peng, Baolin  and
      Wong, Kam-Fai",
    editor = "Jurafsky, Dan  and
      Chai, Joyce  and
      Schluter, Natalie  and
      Tetreault, Joel",
    booktitle = "Proceedings of the 58th Annual Meeting of the Association for Computational Linguistics",
    month = jul,
    year = "2020",
    address = "Online",
    publisher = "Association for Computational Linguistics",
    url = "https://aclanthology.org/2020.acl-main.566/",
    doi = "10.18653/v1/2020.acl-main.566",
    pages = "6355--6365"
}

@inproceedings{jin-etal-2020-hooks,
    title = "Hooks in the Headline: Learning to Generate Headlines with Controlled Styles",
    author = "Jin, Di  and
      Jin, Zhijing  and
      Zhou, Joey Tianyi  and
      Orii, Lisa  and
      Szolovits, Peter",
    editor = "Jurafsky, Dan  and
      Chai, Joyce  and
      Schluter, Natalie  and
      Tetreault, Joel",
    booktitle = "Proceedings of the 58th Annual Meeting of the Association for Computational Linguistics",
    month = jul,
    year = "2020",
    address = "Online",
    publisher = "Association for Computational Linguistics",
    url = "https://aclanthology.org/2020.acl-main.456/",
    doi = "10.18653/v1/2020.acl-main.456",
    pages = "5082--5093"
}

@inproceedings{patro-etal-2018-multimodal,
    title = "Multimodal Differential Network for Visual Question Generation",
    author = "Patro, Badri Narayana  and
      Kumar, Sandeep  and
      Kurmi, Vinod Kumar  and
      Namboodiri, Vinay",
    editor = "Riloff, Ellen  and
      Chiang, David  and
      Hockenmaier, Julia  and
      Tsujii, Jun{'}ichi",
    booktitle = "Proceedings of the 2018 Conference on Empirical Methods in Natural Language Processing",
    month = oct # "-" # nov,
    year = "2018",
    address = "Brussels, Belgium",
    publisher = "Association for Computational Linguistics",
    url = "https://aclanthology.org/D18-1434/",
    doi = "10.18653/v1/D18-1434",
    pages = "4002--4012"
}

@inproceedings{jiang-etal-2020-novel,
    title = "A Novel Workflow for Accurately and Efficiently Crowdsourcing Predicate Senses and Argument Labels",
    author = "Jiang, Youxuan  and
      Zhu, Huaiyu  and
      Kummerfeld, Jonathan K.  and
      Li, Yunyao  and
      Lasecki, Walter",
    editor = "Cohn, Trevor  and
      He, Yulan  and
      Liu, Yang",
    booktitle = "Findings of the Association for Computational Linguistics: EMNLP 2020",
    month = nov,
    year = "2020",
    address = "Online",
    publisher = "Association for Computational Linguistics",
    url = "https://aclanthology.org/2020.findings-emnlp.38/",
    doi = "10.18653/v1/2020.findings-emnlp.38",
    pages = "415--421"
}

@inproceedings{vasilev-etal-2025-ruscode,
    title = "{R}us{C}ode: {R}ussian Cultural Code Benchmark for Text-to-Image Generation",
    author = "Vasilev, Viacheslav  and
      Agafonova, Julia  and
      Gerasimenko, Nikolai  and
      Kapitanov, Alexander  and
      Mikhailova, Polina  and
      Mironova, Evelina  and
      Dimitrov, Denis",
    editor = "Chiruzzo, Luis  and
      Ritter, Alan  and
      Wang, Lu",
    booktitle = "Findings of the Association for Computational Linguistics: NAACL 2025",
    month = apr,
    year = "2025",
    address = "Albuquerque, New Mexico",
    publisher = "Association for Computational Linguistics",
    url = "https://aclanthology.org/2025.findings-naacl.425/",
    doi = "10.18653/v1/2025.findings-naacl.425",
    pages = "7656--7672",
    ISBN = "979-8-89176-195-7"
}

@misc{kang2024humanintheloopsynthetictextdata,
      title={Human-in-the-Loop Synthetic Text Data Inspection with Provenance Tracking}, 
      author={Hong Jin Kang and Fabrice Harel-Canada and Muhammad Ali Gulzar and Violet Peng and Miryung Kim},
      year={2024},
      eprint={2404.18881},
      archivePrefix={arXiv},
      primaryClass={cs.HC},
      url={https://arxiv.org/abs/2404.18881}, 
}

@inproceedings{zelikman-etal-2023-generating,
    title = "Generating and Evaluating Tests for K-12 Students with Language Model Simulations: A Case Study on Sentence Reading Efficiency",
    author = "Zelikman, Eric  and
      Ma, Wanjing  and
      Tran, Jasmine  and
      Yang, Diyi  and
      Yeatman, Jason  and
      Haber, Nick",
    editor = "Bouamor, Houda  and
      Pino, Juan  and
      Bali, Kalika",
    booktitle = "Proceedings of the 2023 Conference on Empirical Methods in Natural Language Processing",
    month = dec,
    year = "2023",
    address = "Singapore",
    publisher = "Association for Computational Linguistics",
    url = "https://aclanthology.org/2023.emnlp-main.135/",
    doi = "10.18653/v1/2023.emnlp-main.135",
    pages = "2190--2205"
}

@inproceedings{weller-seppi-2019-humor,
    title = "Humor Detection: A Transformer Gets the Last Laugh",
    author = "Weller, Orion  and
      Seppi, Kevin",
    editor = "Inui, Kentaro  and
      Jiang, Jing  and
      Ng, Vincent  and
      Wan, Xiaojun",
    booktitle = "Proceedings of the 2019 Conference on Empirical Methods in Natural Language Processing and the 9th International Joint Conference on Natural Language Processing (EMNLP-IJCNLP)",
    month = nov,
    year = "2019",
    address = "Hong Kong, China",
    publisher = "Association for Computational Linguistics",
    url = "https://aclanthology.org/D19-1372/",
    doi = "10.18653/v1/D19-1372",
    pages = "3621--3625"
}

@inproceedings{zeng-etal-2020-meddialog,
    title = "{M}ed{D}ialog: Large-scale Medical Dialogue Datasets",
    author = "Zeng, Guangtao  and
      Yang, Wenmian  and
      Ju, Zeqian  and
      Yang, Yue  and
      Wang, Sicheng  and
      Zhang, Ruisi  and
      Zhou, Meng  and
      Zeng, Jiaqi  and
      Dong, Xiangyu  and
      Zhang, Ruoyu  and
      Fang, Hongchao  and
      Zhu, Penghui  and
      Chen, Shu  and
      Xie, Pengtao",
    editor = "Webber, Bonnie  and
      Cohn, Trevor  and
      He, Yulan  and
      Liu, Yang",
    booktitle = "Proceedings of the 2020 Conference on Empirical Methods in Natural Language Processing (EMNLP)",
    month = nov,
    year = "2020",
    address = "Online",
    publisher = "Association for Computational Linguistics",
    url = "https://aclanthology.org/2020.emnlp-main.743/",
    doi = "10.18653/v1/2020.emnlp-main.743",
    pages = "9241--9250"
}

@inproceedings{khot-etal-2021-text,
    title = "Text Modular Networks: Learning to Decompose Tasks in the Language of Existing Models",
    author = "Khot, Tushar  and
      Khashabi, Daniel  and
      Richardson, Kyle  and
      Clark, Peter  and
      Sabharwal, Ashish",
    editor = "Toutanova, Kristina  and
      Rumshisky, Anna  and
      Zettlemoyer, Luke  and
      Hakkani-Tur, Dilek  and
      Beltagy, Iz  and
      Bethard, Steven  and
      Cotterell, Ryan  and
      Chakraborty, Tanmoy  and
      Zhou, Yichao",
    booktitle = "Proceedings of the 2021 Conference of the North American Chapter of the Association for Computational Linguistics: Human Language Technologies",
    month = jun,
    year = "2021",
    address = "Online",
    publisher = "Association for Computational Linguistics",
    url = "https://aclanthology.org/2021.naacl-main.99/",
    doi = "10.18653/v1/2021.naacl-main.99",
    pages = "1264--1279"
}

@inproceedings{xiong-etal-2023-confidence,
    title = "A Confidence-based Partial Label Learning Model for Crowd-Annotated Named Entity Recognition",
    author = "Xiong, Limao  and
      Zhou, Jie  and
      Zhu, Qunxi  and
      Wang, Xiao  and
      Wu, Yuanbin  and
      Zhang, Qi  and
      Gui, Tao  and
      Huang, Xuanjing  and
      Ma, Jin  and
      Shan, Ying",
    editor = "Rogers, Anna  and
      Boyd-Graber, Jordan  and
      Okazaki, Naoaki",
    booktitle = "Findings of the Association for Computational Linguistics: ACL 2023",
    month = jul,
    year = "2023",
    address = "Toronto, Canada",
    publisher = "Association for Computational Linguistics",
    url = "https://aclanthology.org/2023.findings-acl.89/",
    doi = "10.18653/v1/2023.findings-acl.89",
    pages = "1375--1386"
}

@inproceedings{mousi-etal-2023-llms,
    title = "Can {LLM}s Facilitate Interpretation of Pre-trained Language Models?",
    author = "Mousi, Basel  and
      Durrani, Nadir  and
      Dalvi, Fahim",
    editor = "Bouamor, Houda  and
      Pino, Juan  and
      Bali, Kalika",
    booktitle = "Proceedings of the 2023 Conference on Empirical Methods in Natural Language Processing",
    month = dec,
    year = "2023",
    address = "Singapore",
    publisher = "Association for Computational Linguistics",
    url = "https://aclanthology.org/2023.emnlp-main.196/",
    doi = "10.18653/v1/2023.emnlp-main.196",
    pages = "3248--3268"
}

@inproceedings{hengle-etal-2024-intent,
    title = "Intent-conditioned and Non-toxic Counterspeech Generation using Multi-Task Instruction Tuning with {RLAIF}",
    author = "Hengle, Amey  and
      Padhi, Aswini  and
      Singh, Sahajpreet  and
      Bandhakavi, Anil  and
      Akhtar, Md Shad  and
      Chakraborty, Tanmoy",
    editor = "Duh, Kevin  and
      Gomez, Helena  and
      Bethard, Steven",
    booktitle = "Proceedings of the 2024 Conference of the North American Chapter of the Association for Computational Linguistics: Human Language Technologies (Volume 1: Long Papers)",
    month = jun,
    year = "2024",
    address = "Mexico City, Mexico",
    publisher = "Association for Computational Linguistics",
    url = "https://aclanthology.org/2024.naacl-long.374/",
    doi = "10.18653/v1/2024.naacl-long.374",
    pages = "6716--6733"
}

@article{he-etal-2024-exploring,
    title = "Exploring Human-Like Translation Strategy with Large Language Models",
    author = "He, Zhiwei  and
      Liang, Tian  and
      Jiao, Wenxiang  and
      Zhang, Zhuosheng  and
      Yang, Yujiu  and
      Wang, Rui  and
      Tu, Zhaopeng  and
      Shi, Shuming  and
      Wang, Xing",
    journal = "Transactions of the Association for Computational Linguistics",
    volume = "12",
    year = "2024",
    address = "Cambridge, MA",
    publisher = "MIT Press",
    url = "https://aclanthology.org/2024.tacl-1.13/",
    doi = "10.1162/tacl_a_00642",
    pages = "229--246"
}

@inproceedings{loginova-loguinova-2025-deep,
    title = "Deep Temporal Reasoning in Video Language Models: A Cross-Linguistic Evaluation of Action Duration and Completion through Perfect Times",
    author = "Loginova, Olga  and
      Loguinova, Sof{\'i}a Ortega",
    editor = "Che, Wanxiang  and
      Nabende, Joyce  and
      Shutova, Ekaterina  and
      Pilehvar, Mohammad Taher",
    booktitle = "Proceedings of the 63rd Annual Meeting of the Association for Computational Linguistics (Volume 1: Long Papers)",
    month = jul,
    year = "2025",
    address = "Vienna, Austria",
    publisher = "Association for Computational Linguistics",
    url = "https://aclanthology.org/2025.acl-long.1000/",
    doi = "10.18653/v1/2025.acl-long.1000",
    pages = "20472--20502",
    ISBN = "979-8-89176-251-0"
}

@inproceedings{popovic-belz-2021-reproduction,
    title = "A Reproduction Study of an Annotation-based Human Evaluation of {MT} Outputs",
    author = "Popovi{\'c}, Maja  and
      Belz, Anya",
    editor = "Belz, Anya  and
      Fan, Angela  and
      Reiter, Ehud  and
      Sripada, Yaji",
    booktitle = "Proceedings of the 14th International Conference on Natural Language Generation",
    month = aug,
    year = "2021",
    address = "Aberdeen, Scotland, UK",
    publisher = "Association for Computational Linguistics",
    url = "https://aclanthology.org/2021.inlg-1.31/",
    doi = "10.18653/v1/2021.inlg-1.31",
    pages = "293--300"
}

@misc{eger2026transformingsciencelargelanguage,
      title={Transforming Science with Large Language Models: A Survey on AI-assisted Scientific Discovery, Experimentation, Content Generation, and Evaluation}, 
      author={Steffen Eger and Yong Cao and Jennifer D'Souza and Andreas Geiger and Christian Greisinger and Stephanie Gross and Yufang Hou and Brigitte Krenn and Anne Lauscher and Yizhi Li and Chenghua Lin and Nafise Sadat Moosavi and Wei Zhao and Tristan Miller},
      year={2026},
      eprint={2502.05151},
      archivePrefix={arXiv},
      primaryClass={cs.CL},
      url={https://arxiv.org/abs/2502.05151}, 
}

@inproceedings{karamolegkou-etal-2025-ethical,
    title = "Ethical Concern Identification in {NLP}: A Corpus of {ACL} {A}nthology Ethics Statements",
    author = "Karamolegkou, Antonia  and
      Hansen, Sandrine Schiller  and
      Christopoulou, Ariadni  and
      Stamatiou, Filippos  and
      Lauscher, Anne  and
      S{\o}gaard, Anders",
    editor = "Chiruzzo, Luis  and
      Ritter, Alan  and
      Wang, Lu",
    booktitle = "Proceedings of the 2025 Conference of the Nations of the Americas Chapter of the Association for Computational Linguistics: Human Language Technologies (Volume 1: Long Papers)",
    month = apr,
    year = "2025",
    address = "Albuquerque, New Mexico",
    publisher = "Association for Computational Linguistics",
    url = "https://aclanthology.org/2025.naacl-long.580/",
    doi = "10.18653/v1/2025.naacl-long.580",
    pages = "11618--11635",
    ISBN = "979-8-89176-189-6"
}

@misc{xie2025bridgingaiscienceimplications,
      title={Bridging AI and Science: Implications from a Large-Scale Literature Analysis of AI4Science}, 
      author={Yutong Xie and Yijun Pan and Hua Xu and Qiaozhu Mei},
      year={2025},
      eprint={2412.09628},
      archivePrefix={arXiv},
      primaryClass={cs.AI},
      url={https://arxiv.org/abs/2412.09628}, 
}

@inproceedings{oketch-etal-2025-bridging,
    title = "Bridging the {LLM} Accessibility Divide? Performance, Fairness, and Cost of Closed versus Open {LLM}s for Automated Essay Scoring",
    author = "Oketch, Kezia  and
      Lalor, John P.  and
      Yang, Yi  and
      Abbasi, Ahmed",
    editor = "Arviv, Ofir  and
      Clinciu, Miruna  and
      Dhole, Kaustubh  and
      Dror, Rotem  and
      Gehrmann, Sebastian  and
      Habba, Eliya  and
      Itzhak, Itay  and
      Mille, Simon  and
      Perlitz, Yotam  and
      Santus, Enrico  and
      Sedoc, Jo{\~a}o  and
      Shmueli Scheuer, Michal  and
      Stanovsky, Gabriel  and
      Tafjord, Oyvind",
    booktitle = "Proceedings of the Fourth Workshop on Generation, Evaluation and Metrics (GEM{\texttwosuperior})",
    month = jul,
    year = "2025",
    address = "Vienna, Austria and virtual meeting",
    publisher = "Association for Computational Linguistics",
    url = "https://aclanthology.org/2025.gem-1.60/",
    pages = "655--669",
    ISBN = "979-8-89176-261-9"
}

\appendix
\section{\label{appx:data}Datasets}
The following 34 keywords were used to identify papers involving human annotation based on matches in titles and abstracts.

\begin{itemize}
\setlength{\itemsep}{2pt}
\setlength{\topsep}{1pt}

\item Amazon Mechanical Turk
\item annotator
\item crowdworker
\item crowdsourcer
\item crowdsourcing
\item expert annotation
\item expert evaluation
\item expert opinion
\item HIL
\item human annotation
\item human annotator
\item human benchmark
\item human dataset
\item human evaluation
\item human expert
\item human in the loop
\item human judgment
\item human judgement
\item human performance
\item human perception
\item human rating
\item human raters
\item human respondent
\item human study
\item human-in-the-loop
\item labeler
\item manual annotation
\item manual evaluation
\item manual rating
\item manually annotated
\item manually evaluated
\item MTurk
\item Prolific
\item student worker

\end{itemize}

\subsection{Random-sample validation of keyword retrieval}
\label{appx:random_validation}

The large-scale extraction set, \datasetsub{LLM}, is constructed using keyword-based retrieval over paper titles and abstracts. This strategy is designed to increase the concentration of papers likely to contain human annotation tasks, but it may also change the composition of the resulting sample. To assess this effect, we compare \datasetsub{LLM} with a stratified random sample drawn from the same venue-year scope.

The random sample contains 3,000 papers drawn across the same venues and publication years. Of these, 1,085 papers contain annotatable human annotation content, corresponding to a retention rate of 36\%. In contrast, keyword-based retrieval yields 1,603 retained papers out of 1,995 candidates, corresponding to a retention rate of 82\%. This confirms that keyword retrieval substantially improves the efficiency of identifying papers with human annotation content.

\begin{figure}[ht]
    \centering
    \resizebox{\columnwidth}{!}{%
        \includegraphics{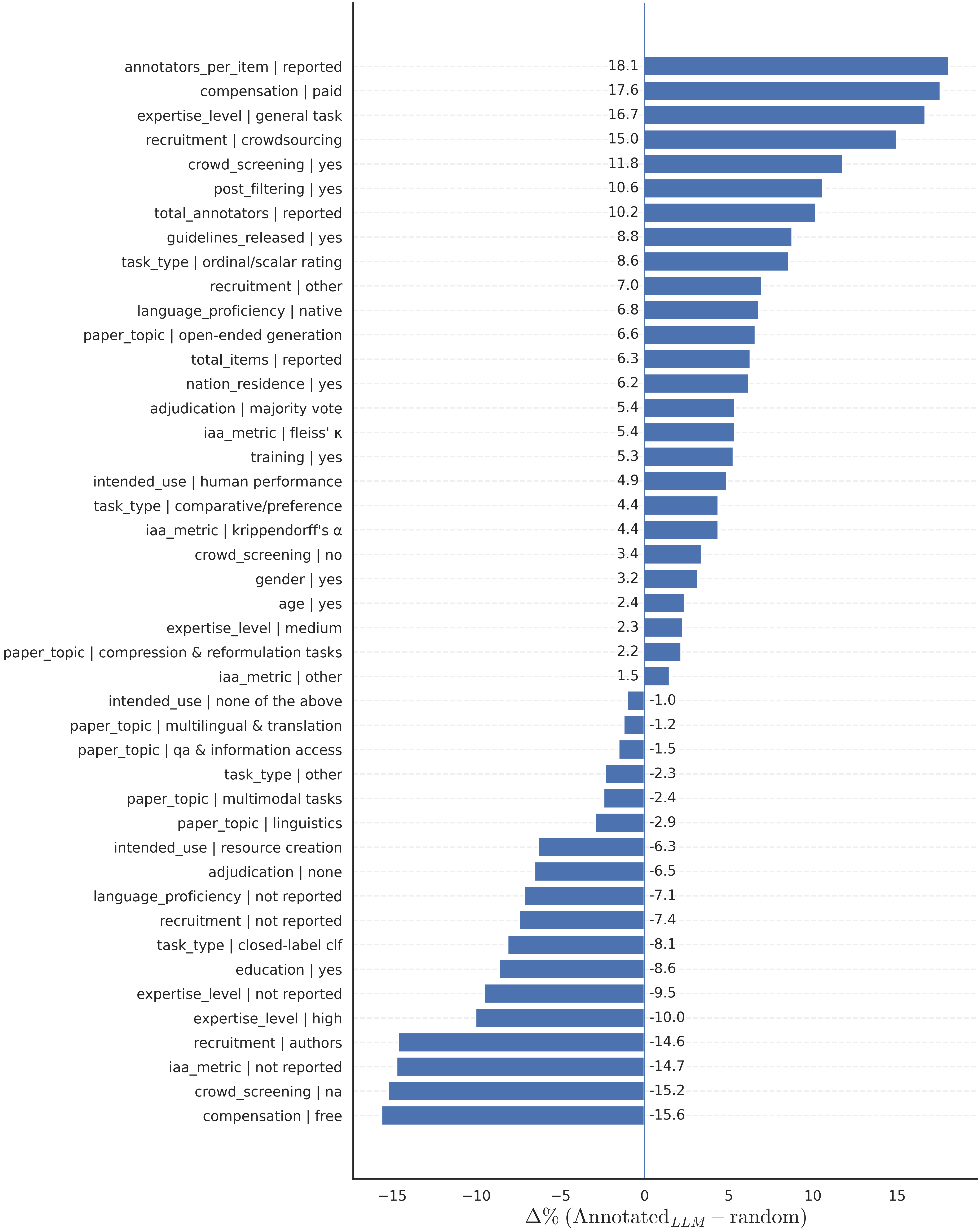}
    }
    \caption{\label{fig:cross_sample}Significant proportional differences ($\Delta$\%) between filtered and random samples across category–value pairs (\abc{c}hi-square test, p < 0.05). Bars are sorted by effect size and centered at zero to indicate direction (over- vs under-representation in the \abc{keyword-based sample}). For binary and binarized string fields, only the significant proportional deltas for the positive value (e.g., ``yes'', ``reported'') are shown to avoid duplication.}
    
\end{figure}

\mk{To check whether our sampling strategy introduced any bias, w}e compare the distributions of taxonomy-value pairs between the keyword-retrieved set and the random sample. For each category-value pair, we compute the normalized frequency in both samples and test the difference in proportions using a $\chi^2$ test. Figure~\ref{fig:cross_sample} shows only category-value pairs with statistically significant differences at $p < 0.05$. To avoid duplicate visualizations for binary variables, the figure reports only the positive or reported value where applicable.

The comparison shows that keyword retrieval introduces some statistically significant composition differences. However, these differences are generally modest: among statistically significant category-value pairs, the average absolute difference is 5.2 percentage points. 
\mk{Only a small number of category-value pairs exhibit substantial deviations, with extreme cases reaching $-16$ and $+18$ percentage points in opposite directions.}
In other words, the keyword-retrieved set is not simply higher or lower on all reporting dimensions. Instead, it over-represents some category values and under-represents others.

We therefore interpret \datasetsub{LLM} as an annotation-focused corpus rather than a fully representative sample of all ACL-venue papers. The random-sample comparison suggests that keyword retrieval changes sample composition, but does not reveal a single systematic directional bias across reporting categories. This supports our use of \datasetsub{LLM} for analyzing reporting patterns among papers likely to involve human annotation, while motivating caution in generalizing the exact proportions to \mk{all ACL Anthology papers containing human annotation tasks}.


\section{\label{appx:score}\metric{} Calculation}

The \metric{} is computed as the proportion of reported annotation attributes relative to the set of attributes applicable to a given annotation task. The full annotation schema contains 21 reporting categories grouped into:
\begin{itemize}
    \item 10 universal attributes expected for all studies (guidelines release, recruitment strategy, training procedures, level of expertise, language proficiency, educational level, number of annotators, number of annotated items, compensation, post-annotation filtering)
    \item 5 socio-demographic attributes describing annotator background (age, gender, nation of residence, nation of origin, political orientation), which are expected in tasks annotating subjective or socially-constructed language phenomena
    \item 6 attributes that can be excluded as inapplicable or redundant given other types of information reported in the paper for the task (inter-annotator metric and value, annotator per item, items per annotator, adjudication, crowdworker screening)
\end{itemize}

The denominator is adjusted dynamically to avoid penalizing papers for omitting inapplicable or redundant information\abc{, following a set of rules as follows}. 
Specifically, the papers belonging to \textit{Subjective language \& social meaning}, the denominator includes the five socio-demographic characteristics, for other papers, they are viewed as optional. 
Inter-annotator agreement (IAA) and adjudication are excluded from the denominator for single-annotator studies (i.e., when \texttt{total\_annotators = 1} or \texttt{annotators\_per\_item = 1}, which is often the case for text production tasks); the IAA value is excluded if no IAA metric is reported; crowdworker screening is excluded when recruitment strategy does not involve crowdsourcing; and adjudication is excluded for tasks without open-ended annotation.

To avoid double-counting, redundant fields \abc{are} excluded when they encode equivalent information. For example, if \texttt{total\_annotators} and \texttt{annotators\_per\_item} have the same number, one of them is omitted. The \texttt{items\_per\_annotator} \abc{is} excluded when recoverable from other reported fields. 
Missing information is treated as incomplete reporting; for instance, reporting an IAA metric without the corresponding value does not count as valid reportage. Fine-grained compensation extensions such as payment rate are treated as refinements rather than independent reporting dimensions and are excluded from scoring.

The final score is computed separately for each annotation task in a paper using the resulting set of applicable reporting categories.

\section{\label{appx:demogr}Annotator demographics}

\la{In total, 12 annotators contributed to the annotation process: 2 professors, 2 postdoctoral researchers, 6 PhD students, and 2 master’s students. Among the annotators, 5 identified as women and 7 as men. The annotator pool included individuals originating from Italy, Germany, India, China, and Russia. All annotators reported fluent English proficiency, as all annotated papers were written in English. At the time of annotation, most annotators were affiliated with institutions located in Germany and Austria. Annotation was carried out on a part-time basis alongside regular academic activities. Professors contributed approximately 10 hours in total, while postdoctoral researchers, PhD students, and master’s students each contributed approximately 50 hours overall. Based on the 2026 standardized academic personnel rates published by the Deutsche Forschungsgemeinschaft, the estimated total personnel cost of the annotation process was approximately 6,300 \abc{Euros}.}

\onecolumn
\section{\label{appx:humanno}Details on Human Annotation}

\begin{table*}[h]
\centering
\small
\begin{tabularx}{\textwidth}{@{}l c X c c@{}}
\toprule
\textbf{Paper ID} & \textbf{Tasks} & \textbf{Domain} & \textbf{Year} & \textbf{Venue} \\
\midrule
\citet{wang-etal-2020-learning-efficient} & 3 & Task-Oriented Dialogue Learning & 2020 & ACL \\
\citet{jin-etal-2020-hooks} & 3 & Headlines Generation with Controlled Styles & 2020 & ACL \\
\citet{zeinert-etal-2021-annotating} & 1 & Misogyny detection & 2021 & ACL \\
\citet{zhang-etal-2021-emailsum} & 3 & Text Summarization  & 2021 & ACL \\
\citet{jiang-riloff-2021-learning}& 1 & Prototypical function labeling of physical artifacts & 2021 & ACL \\
\citet{asai-choi-2021-challenges} & 2 & Information-Seeking QA & 2021 & ACL \\
\citet{haber-etal-2023-improving} & 3 & Detection of Multilingual Online Attacks & 2023 & ACL \\
\citet{sai-b-etal-2023-indicmt} & 1 & Meta-Evaluation of Machine Translation Metrics & 2023  & ACL \\
\citet{agnew-etal-2023-mechanical} & 2 & Poetry Generation & 2023 & ACL \\
\citet{xiong-etal-2023-confidence} & 2 & Crowd-Annotated Named Entity Recognition & 2023 & ACL \\
\citet{el-kheir-etal-2024-beyond} & 2 & Dialectal Sound and Vowelization Recovery & 2024 & ACL \\
\citet{ide-etal-2025-coam} & 1 & MWE Corpus Creation & 2025 & ACL \\
\citet{zou-2025-bipro} & 1 & Poetry Generation & 2025 & ACL\\
\citet{loginova-loguinova-2025-deep}& 2 & Deep Temporal Reasoning in Video Language Models & 2025 & ACL \\
\midrule
\citet{breit-etal-2021-wic} & 1 & Target Sense Verification & 2021 & EACL\\
\midrule
\citet{patro-etal-2018-multimodal} & 1 & Visual Question Generation & 2018 & EMNLP \\
\citet{weller-seppi-2019-humor} & 1 & Humor Detection & 2019 & EMNLP \\
\citet{jiang-etal-2020-novel} & 2 & Crowdsourced Semantic Role Labeling & 2020 & EMNLP \\
\citet{zeng-etal-2020-meddialog} & 1 & Large-scale Medical Dialogue Datasets & 2020 & EMNLP \\
\citet{chakrabarty-etal-2021-dont} & 1 & Neural Poetry Translation & 2021 & EMNLP \\
\citet{mousi-etal-2023-llms}& 3 & Concept-Based Model Interpretability & 2021 & EMNLP \\
\citet{lin-etal-2023-argue} & 3 & Counter-argument Generation & 2023 & EMNLP  \\
\citet{zelikman-etal-2023-generating} & 3 & Generating and Evaluating Educational Tests & 2023 & EMNLP \\
\citet{doneva-etal-2024-neurotrialner} & 1 & Biomedical NER Corpus Creation & 2024 & EMNLP \\
\citet{kostikova-etal-2024-fine} & 1 & Solidarity Detection & 2024 & EMNLP \\
\citet{walsh2024sonnetnotbotpoetry} & 1 & Poetry Evaluation & 2024 & EMNLP\\
\citet{altemeyer2025argumentsummarizationevaluationera} & 2 & Argument Summarization and Evaluation & 2025 & EMNLP \\
\citet{mathur-etal-2025-social} & 3 & Social Reasoning & 2025 & EMNLP \\
\citet{ivetta2025heseiacommunitybaseddatasetevaluating} & 1 & Community-Based Bias Evaluation & 2025 & EMNLP  \\
\citet{huang2025visbiasmeasuringexplicitimplicit} & 1 & Social Biases in Vision Language Models & 2025 & EMNLP \\
\citet{chen-etal-2025-benchmarking-llms} & 2 & Benchmarking for Translating Classical Chinese Poetry & 2025 & EMNLP \\
\midrule
\citet{belouadi2025tikzerozeroshottextguidedgraphics} & 1 & Text-to-Image Generation & 2025 & ICCV \\
\midrule
\citet{greisinger2026tikzillascalingtexttotikzhighquality} & 3 & Evaluation of TikZ Generated Code & 2026 & ICLR \\
\midrule
\citet{khot-etal-2021-text} & 1 & Question Answering & 2021 & NAACL \\
\citet{zhao-etal-2024-comprehensive} & 1 & Gender Bias in Chemical NER Models & 2024 & NAACL\\
\citet{li2024contradocunderstandingselfcontradictionsdocuments} & 2 & Contradiction Detection & 2024 & NAACL \\
\citet{kang2024humanintheloopsynthetictextdata} & 2 & Synthetic Text Data Inspection & 2024 & NAACL \\
\citet{hengle-etal-2024-intent} & 2 & Counterspeech Generation & 2024 & NAACL \\
\citet{vasilev-etal-2025-ruscode} & 2 & Text-to-Image Generation & 2025 & NAACL \\
\midrule
\citet{belouadi2024detikzifysynthesizinggraphicsprograms} & 1 & Text-to-Image Generation & 2024 & NIPS \\
\midrule
\citet{he-etal-2024-exploring} & 2 & Human-Like Translation Strategy & 2024 & TACL \\
\bottomrule
\end{tabularx}
\caption{\label{tab:papers}Double-annotated and adjudicated papers included in the gold dataset.}
\end{table*}
\clearpage
\begin{xltabular}{\textwidth}{c|p{2cm}p{2.9cm} X}

\toprule
\textbf{Aspect} & \textbf{Category} & \textbf{Values} & \textbf{Instruction} \\
\midrule
\endfirsthead

\textbf{Aspect} & \textbf{Category} & \textbf{Values} & \textbf{Instruction} \\
\midrule
\endhead
\multirow{11}{*}{%
  \rotatebox{90}{\parbox{13cm}{\textbf{General Description}}}%
}
& \multirow{2}{2cm}{Paper's topic} 
& & This category captures the core problem the paper introduces or solves, as stated in the abstract or introduction. A paper gets exactly one label, determined by the dominant topical focus of contribution, usually found in abstract. The 10 drop-down options are adopted from ACL categories. In case of hybrid papers, decide what is their primary focus and contribution. If in doubt, assign at most two categories.\\
& & Linguistics & Core linguistic representation, word to sentence levels (e.g. PoS, parsing, NER, semantic roles, word sense disambiguation, entity linking). \\
& &  Discourse/Pragmatics  & The study of cross-sentence meaning and contextual semantics, relations between sentences/clauses in a text and text-level linguistic phenomena (coreference resolution, discourse relations, implicature/ellipsis resolution, topic segmentation, dialogue act tagging without generation).\\
& & Information Extraction  & Advances in text to knowledge bases or structured schema conversion (knowledge graph construction, slot filling, relation/event/temporal prediction). \\
& & QA/Information access  & Answering questions from text or corpora (extractive, open-domain, multi-hop QA, both deterministic SQuAD and open-ended).\\
& & Semantics, inference \& similarity  & Judging meaning relations between texts (entailment, reasoning benchmarks, paraphrase-detection, STS, NLI). \\
& & Subjective language \& social meaning & Topics that explore social and cultural phenomena through language (sentiment, emotion, stance, humor, sarcasm, toxicity, bias, propaganda, authorship attribution). \\
& & Open-ended generation & Open-ended output production (dialogue/story/poetry/jokes/argument generation from a given topic or context, instruction following). \\
& & Rewriting tasks & Unlike generic generation, this task transforms a given text and is required to remain faithful to the input in some aspect (summarisation, simplification, paraphrase-generation, style transfer, rewriting under constraints). \\
& & {\small Multilingual/Translation}  & Cross-linguistic mapping (MT, cross-lingual transfer evaluation, code-switching, multilingual generation). \\
& & Multimodal tasks & Language from other modalities (image/graphics captioning, visual QA, video description, multimodal assistants). \\
\cmidrule(lr){2-4}
& \multirow{2}{2cm}{Task type} 
& & Select the type of task performed by human annotators. \\
& & Closed-label classification  & Tasks where annotators choose from predefined categories (binary, multi-class, multi-label, tagging with a fixed schema). \\
& & Ordinal/scalar rating   & Tasks where annotators pass judgments on a scale (Likert scale, 1-5 ratings, graded acceptability, quality scores). \\
& & Span-level or structural annotation  & Annotators mark parts of the input or define structure (NER spans, coreference links, constituency/dependency trees, alignment). \\
& & {\small Comparative/preference judgments} & Relative evaluation of two or more items (A/B choice, ranking, pairwise preference, best–worst scaling). \\
& & Open-ended generation or transformation  & Annotators produce new text or modify input (summarization, simplification, paraphrasing, translation, explanation writing). \\
& & Other & None of the above \\
\cmidrule(lr){2-4}
& Task identity  & [User input] & Record the specific name of each human annotation subtask as reported in the paper. This allows matching the order of subtasks within a paper.
Especially important for papers reporting up to three annotation studies, but apply to all papers.
How to extract: Use the exact name or description of the subtask as given in the paper. Do not create new names or abbreviations.\\
\cmidrule(lr){2-4}
& \multirow{2}{2cm}{Intended use}  
& & Categorizes types of human annotation studies by their purpose. Select Resource creation and Human performance when an annotated dataset is used as a human baseline, without collecting additional judgments.\\
& & Resource creation & Human judgment are used to build a dataset, a gold standard, a benchmark or a train/test set for potential further use, including for a subsequent evaluation of their own models in the same study. \\
& & Human performance & Humans solve the same task as the machine to provide the human upper-bound and to compare human-machine performances. \\
& & Model output evaluation & Collecting human judgment to evaluate automatically generated items. \\
& & None of the above &  A fallback option\\
\cmidrule(lr){2-4}
& \multirow{2}{2cm}{Guidelines release}  
& & Indicates whether the annotation guidelines released by the paper allow the experiment to be reproduced.\\
& & Yes & Guidelines are accessible (in the paper, appendix, via a working link or citation of another paper) and contain all information necessary to reproduce the annotation. Select Yes only if someone could realistically repeat the annotation using the provided materials. \\
& & No & Guidelines are not accessible, or do not contain enough information to reproduce the annotation. \\
\midrule

\multirow{2}{*}{%
  \rotatebox{90}{\parbox{3.4cm}{\textbf{Agreement Level}}}%
}
& IAA metric 
& Fleiss' $\kappa$, Cohen's $\kappa$, Krippendorff's $\alpha$, F1 agreement, Pearson, Spearman, Majority agreement, Kendall's $\tau$, Total agreement, Other, Not reported & Multiple selection dropdown. 
If more than one metric is reported, select several items in the order of appearance in the paper. 
The corresponding values should be entered in the IAA Value column in the same order separated by a pipe (|).
Other: some other IAA measure/method is explicitely reported. 
Not reported: either not applicable (one annotator) or not given in the paper.\\
\cmidrule(lr){2-4}
& IAA value & [User input] & A string field. Refrain from inferring values.
Separate multiple values (aligned with IAA metric names) with a pipe (|).
Enter a string if ranges are reported, or any other form of report is made (not one value).
NA: if Other metric is reported or if the value is not directly reported and needs to be inferred.
Not reported: either not applicable (one annotator) or not given in the paper.\\
\midrule

\multirow{4}{*}{%
  \rotatebox{90}{\parbox{2.2cm}{\textbf{Workload}}}%
} 
& Annotators/item & [User input] &  \multirow{4}{9cm}[-1ex]{A number or "Not reported" string. Type ``Not reported'' if this information is not explicitly given.}\\
\cmidrule(lr){2-3}
& Total annotators & [User input] &  \\
\cmidrule(lr){2-3}
& Total items  & [User input] &  \\
\cmidrule(lr){2-3}
& Items/annotator & [User input] &  \\
\midrule

\multirow{24}{*}{%
  \rotatebox{90}{\parbox{10cm}{\textbf{Recruitment and Qualifications}}}%
} 
&  \multirow{2}{2cm}{Recruitment} 
& & This category captures who performed the annotation, i.e., the source from which annotators were recruited.\\
&   & Crowdsourcing &  Annotators recruited via a crowdsourcing platform or described as crowdworkers.\\
&  & Authors &  One or more authors performed the annotation.\\
& & Other &  Annotators are a specified non-crowd group (e.g., students, experts, hired assistants, colleagues).\\
& & Mixed & Annotators include people from different specified groups.\\
& & Not reported & The source of annotators is not specified as a specific group.\\
\cmidrule(lr){2-4}

&  \multirow{2}{2cm}{Crowd qualifications} 
&  & Applies only to crowdsourced annotators. If the paper does not use crowdworkers, select NA.\\
& & Yes& The paper reports preliminary screening or filtering of crowdworkers to ensure annotation quality (beyond basic socio-demographics).\\
& & No& The paper does not provide any information on quality-related pre-selection or screening of crowdworkers beyond socio-demographics. Choose No if Recruitment is Not reported. \\
& & NA& There is an explicit mention of other-than-crowd sources of annotators (other and authors).\\
\cmidrule(lr){2-4}
 
&  \multirow{2}{2cm}{Annotator training} 
&  & Is any preliminary annotator training reported (e.g. calibration sessions, trial studies, hands-on prior experience with the annotation tool).\\
& & Yes& Some form of annotator training is reported.\\
& & No& Either not discussed or explicit mention of no training provided.\\

\cmidrule(lr){2-4}
&  \multirow{2}{2cm}{Language proficiency} 
&  & Capture the reported language proficiency of annotators with respect to the language of the annotation task.\\
& & Native& The annotators are directly described as native speakers. If annotators’ ethnicity is reported, and it aligns with the language of the annotation task, annotators should be considered native speakers. \\
& & Non-native& The authors report some level of proficiency in the language of the annotated data (fluent, B2, C1), but not native.\\
& & Mixed& Both native and non-native speakers were employed.\\
& & Not reported& Language proficiency is not directly reported.\\
\cmidrule(lr){2-4}

&  \multirow{2}{2cm}{Level of expertise} 
&  & Indicates the actual expertise of the employed annotators for tasks where specialized knowledge is relevant.
\\
& & High& Annotators have specialized domain knowledge relevant to the task (e.g., medical, legal, technical, linguistic experts).\\
& & Medium& Annotators have some relevant background, but are not deeply specialized (e.g., students in a related field).\\
& & Mixed& Reported levels of expertise vary (e.g. experts and students, or students and unfiltered crowd workers).\\
& & Not reported& The task could require specialized knowledge, but the paper does not specify the annotators' expertise.\\
& & General language task& The task does not require specialized knowledge.\\
\midrule

\multirow{8}{*}{%
  \rotatebox{90}{\parbox{5cm}{\textbf{Compensation}}}%
} 
& \multirow{2}{2cm}{Reported compensation} & & Indicates whether annotator compensation is described.\\
& & Paid& Some compensation is mentioned, including ``annotators were paid'', ``full-time employees''.\\
& & Free& Cases, where voluntary effort is reported (authors of the paper or colleagues).\\
& & Not reported& No mention of compensation.\\
\cmidrule(lr){2-4}

& \multirow{2}{2cm}{Payment rate} & & Provide specifics on how compensation is reported for papers where compensation is Paid in Reported compensation.\\
& & Specific numeric rate& A concrete payment is given (e.g., hourly rate, per-task payment, total amount).\\
& & General mention& Compensation is mentioned, but no exact amount is provided (e.g., ``annotators were paid'', ``full-time employees'').\\
& & NA& Select NA for cases where the compensation is ``Free'' or ``Not reported'' in Reported compensation.\\
\midrule

\multirow{6}{*}{%
  \rotatebox{90}{\parbox{4cm}{\textbf{Socio-Demographics}}}%
} 
& Age& Yes; No & Is age reported?\\
\cmidrule(lr){2-4}

& Gender& Yes; No & Is gender of annotators reported?\\
\cmidrule(lr){2-4}

& Nation origin& Yes; No & Is nationality of annotators discussed? Nationality is interpreted as the country of birth.\\
\cmidrule(lr){2-4}

& Nation residence& Yes; No & Is the nation of residence at the time of annotation discussed? It can be different from the nation of origin.\\
\cmidrule(lr){2-4}

& Education& Yes; No & Is educational level or field of study reported?\\
\cmidrule(lr){2-4}

& Political orientation& Yes; No & Is political orientation discussed?\\
\midrule
\pagebreak
\multirow{2}{*}{%
  \rotatebox{90}{\parbox{7cm}{\textbf{Quality Control}}}%
} 
& \multirow{3}{2cm}{Post-filtering}&  & Does the paper report any quality control or filtering of annotations after collection?\\
& &  Yes& Some form of control and post-annotation filtering is reported or they explicitly report disqualifying experts due to lack of guidelines understanding or removing items due to issues revealed during annotation. \\
& & No& Nothing is reported about filtering the annotation results.\\
\cmidrule(lr){2-4}

& \multirow{2}{2cm}{Adjudication}&  & Captures how annotation disagreements are handled to produce final labels. This refers to explicit procedures for resolving conflicting judgments at the item level (not simply aggregating scores across all annotated items).\\
& & Majority vote & The most common label among annotators is selected.\\
& & Expert adjudication & A designated expert or third-party judge reviews disagreements and assigns the final label.\\
& & Third annotator & Another annotator from the annotator pull steps in for arbitration between diverging opinions. \\
& & Consensus discussion &  Annotators discuss and agree on a final label.  \\
& & Soft labels & Disagreement is preserved as a label distribution rather than resolved. \\
& & Weighted voting & Votes are weighted based on annotator reliability or expertise. \\
& & Other / mixed & Any combination or alternative method. \\
& & None & No explicit disagreement resolution; annotator outputs are used as-is (e.g., simple averaging or single annotator per item). \\
\bottomrule
\caption{\label{tab:anno_guide}Annotation categories, values and instructions by aspect.}
\end{xltabular}
\twocolumn
\clearpage 

\clearpage
\onecolumn
\begin{table*}[!h]
\centering
\caption{Inter-annotator agreement by category and aspect. Human--Human (N=71) vs Human--LLM (N=64).}
\label{tab:iaa_full}
\setlength{\tabcolsep}{6pt}
\resizebox{\textwidth}{!}{
\begin{tabular}{llcc|cc}
\toprule
\textbf{Aspect} & \textbf{Category} & \multicolumn{2}{c|}{\textbf{Human--Human}} & \multicolumn{2}{c}{\textbf{Human--LLM}} \\
 &  & Agree\% & $\alpha$ & Agree\% & $\alpha$ \\
\midrule

\multirow{4}{*}{General Description}
 & Paper's topic        & 77.5 & 0.677 & 70.3 & 0.290 \\
 & Task type            & 73.2 & 0.660 & 71.9 & 0.638 \\
 & Intended use         & 81.7 & 0.535 & 79.7 & 0.592 \\
 & Guidelines released  & 66.2 & 0.328 & 71.9 & 0.433 \\
\cmidrule{2-6}
 & \textit{Avg}         & \textbf{74.6} & \textbf{0.550} & \textbf{73.4} & \textbf{0.488} \\
\midrule

\multirow{2}{*}{Agreement Level}
 & IAA metric           & 88.7 & 0.815 & 84.4 & 0.735 \\
 & IAA value            & 93.0 & 0.897 & 75.0 & 0.603 \\
\cmidrule{2-6}
 & \textit{Avg}         & \textbf{90.8} & \textbf{0.856} & \textbf{79.7} & \textbf{0.669} \\
\midrule

\multirow{4}{*}{Workload}
 & Annotators/item      & 67.6 & 0.541 & 68.8 & 0.559 \\
 & Total annotators     & 73.2 & 0.693 & 71.9 & 0.681 \\
 & Total items          & 49.3 & 0.453 & 57.8 & 0.546 \\
 & Items/annotator      & 69.0 & 0.246 & 76.6 & 0.253 \\
\cmidrule{2-6}
 & \textit{Avg}         & \textbf{64.8} & \textbf{0.483} & \textbf{68.8} & \textbf{0.510} \\
\midrule

\multirow{5}{*}{Recruitment \& Qualif.}
 & Recruitment          & 66.2 & 0.518 & 59.4 & 0.384 \\
 & Crowd screening      & 87.3 & 0.671 & 87.5 & 0.586 \\
 & Training             & 83.1 & 0.625 & 87.5 & 0.693 \\
 & Language proficiency & 77.5 & 0.549 & 84.4 & 0.641 \\
 & Expertise level      & 52.1 & 0.353 & 53.1 & 0.399 \\
\cmidrule{2-6}
 & \textit{Avg}         & \textbf{73.2} & \textbf{0.543} & \textbf{74.4} & \textbf{0.540} \\
\midrule

\multirow{2}{*}{Compensation}
 & Compensation         & 84.5 & 0.741 & 81.2 & 0.681 \\
 & Payment rate         & 95.8 & 0.909 & 90.6 & 0.784 \\
\cmidrule{2-6}
 & \textit{Avg}         & \textbf{90.1} & \textbf{0.825} & \textbf{85.9} & \textbf{0.733} \\
\midrule

\multirow{6}{*}{Demographics}
 & Age                  & 98.6 & 0.949 & 96.9 & 0.858 \\
 & Gender               & 97.2 & 0.893 & 96.9 & 0.872 \\
 & Nation origin        & 88.7 & -0.052 & 96.9 & 0.653 \\
 & Nation residence     & 84.5 & 0.336 & 89.1 & 0.529 \\
 & Education            & 79.7 & 0.580 & 79.7 & 0.597 \\
 & Political orient.    & 100.0 & 1.000 & 100.0 & 1.000 \\
\cmidrule{2-6}
 & \textit{Avg}         & \textbf{91.3} & \textbf{0.618} & \textbf{93.2} & \textbf{0.751} \\
\midrule

\multirow{2}{*}{Quality Control}
 & Post-filtering       & 66.2 & 0.194 & 85.9 & 0.602 \\
 & Adjudication         & 78.9 & 0.520 & 79.7 & 0.539 \\
\cmidrule{2-6}
 & \textit{Avg}         & \textbf{72.5} & \textbf{0.357} & \textbf{82.8} & \textbf{0.570} \\
\midrule

\textbf{Overall} 
 & All categories (macro avg) 
 & \textbf{79.2} & \textbf{0.585} 
 & \textbf{79.9} & \textbf{0.606} \\
\bottomrule
\end{tabular}}
\end{table*}

\twocolumn



\clearpage
\section{LLM Extraction Prompt}
\label{app:llm_extraction_prompt}
The following prompt was used to guide the Large Language Model in extracting structured data from the scientific papers according to our defined taxonomy.

\begin{lstlisting}[style=promptstyle]
You are extracting structured data about HUMAN ANNOTATION experiments from a scientific paper.
Your task is to fill in a rigid taxonomy. You MUST use ONLY the exact allowed values specified below.
Do NOT paraphrase, do NOT add explanations, do NOT use synonyms.

## Annotation Taxonomy -- EXACT Allowed Values

You MUST output values EXACTLY as listed below -- no paraphrasing, no synonyms, no extra words.

### 1. paper_experiment_id
Format: {filename}-1, {filename}-2, etc.

### 2. paper_topic -- LIST of one or more, each EXACTLY one of:
  * "Linguistics"
  * "Discourse & Pragmatics"
  * "Information Extraction"
  * "QA & information access"
  * "Semantics, inference & similarity"
  * "Subjective language & social meaning"
  * "Open-ended generation"
  * "Compression & reformulation tasks"
  * "Multilingual & translation"
  * "Multimodal tasks"
  Classify the paper's topic(s) based on its main contributions. Use a list with one or more values.

### 3. human_annotation_type -- PICK ONE exactly:
  * "Closed-label classification"
  * "Ordinal / scalar rating"
  * "Span-level or structural annotation"
  * "Comparative / preference judgments"
  * "Open-ended generation or transformation"
  * "Other"
Decision rules:
  - Binary / multi-class / multi-label / tagging with fixed schema -> "Closed-label classification"
  - Likert scales, 1-5 ratings, quality scores, graded judgments -> "Ordinal / scalar rating"
  - NER spans, coreference, constituency/dependency trees, sequence labeling, alignment -> "Span-level or structural annotation"
  - A/B choice, ranking, pairwise preference, best-worst scaling -> "Comparative / preference judgments"
  - Summarization, paraphrasing, translation, simplification, free-text generation -> "Open-ended generation or transformation"
  - Anything that does not fit the above five -> "Other"

### 4. subtask_name
Short descriptive name of the HUMAN ANNOTATION subtask -- what the annotators were asked to label, rate, or judge.
If one experiment covers multiple subtasks that share the same annotator pool and annotation round, combine them with " | " separator.
  [!] Name the ANNOTATION task, NOT the downstream NLP modelling task:
      * Write "hate speech annotation", NOT "hate speech classifier"
      * Write "stance labeling", NOT "stance detection model"
      * Write "named entity tagging", NOT "NER model training"
  The name must reflect what humans were asked to DO, not the downstream prediction objective.

  [!] Be SPECIFIC -- mirror the paper's own wording, not a generic category:
      CORRECT (specific):   "binary online attack classification"
      WRONG (too generic):  "attack annotation"
      CORRECT (specific):   "clinical named entity recognition (NER)"
      WRONG (too generic):  "NER annotation"
      CORRECT (specific):   "Pairwise Comparison of Model-Generated Summaries"
      WRONG (too generic):  "evaluation"
      CORRECT (specific):   "rating arguments on a Likert scale across 5 dimensions"
      WRONG (too generic):  "argument rating"
      CORRECT (specific):   "evaluate generated headlines for relevance, attractiveness and language fluency"
      WRONG (too generic):  "headline evaluation"

  [!] Use " | " ONLY for subtasks sharing the SAME annotator pool and annotation round.
     If tasks differ in annotator group, annotation unit, instructions, or objective -> create SEPARATE experiments.
     Example of correct " | " use: "Misogyny detection - Abusive language detection | Misogyny detection - Target identification | Misogyny detection - Hate speech categorization"
       (all three done by the same annotators in the same study)

  [!] Papers routinely contain 2-3 distinct annotation experiments. Common patterns:
      * Different label spaces on the same data (e.g. hate speech -> then target type -> then severity)
      * Same task repeated by two different annotator groups (e.g. expert gold-standard AND crowd annotation)
      * A text creation task PLUS a separate rating/evaluation task on the model output
      * Multiple rounds: pilot annotation + main annotation with different guidelines or annotators, if IAA is reported separately
      Each of these MUST be a separate experiment with its own row.

### 5. intended_use -- LIST of one or more, each EXACTLY one of:
  * "Resource creation"    (the annotation is primarily for building a dataset/resource)
  * "Human performance"    (the annotation measures human performance on a task)
  * "Model output evaluation" (the annotators evaluate model-generated outputs)
  * "None of the above"    (does not fit any of the above categories)
  Use a list with one or more values. Multiple uses are common (e.g. ["Resource creation", "Model output evaluation"]).
  - DEFAULT: ["None of the above"]

### 6. guidelines_released -- "Yes" or "No"
  - "Yes" IF any of the following are true:
      * The paper includes a dedicated appendix section with detailed annotation instructions (not just a brief task description in the main text)
      * A URL / hyperlink to guidelines, an annotation manual, or a codebook is provided (even if the link may now be broken)
      * The paper cites another paper whose guidelines were re-used and those guidelines are thus accessible via that citation
      * Supplementary material, a shared GitHub/OSF repository, or a dataset release specifically includes the annotation guidelines or schema
      * The paper explicitly states that guidelines were released or made available
  - "No" if guidelines are only described informally in the main text without a dedicated document, appendix, cited source, or link; OR if no guidelines information is given
  - DEFAULT: "No"
  [!] Common mistake: Do NOT mark "Yes" just because the paper describes the annotation task in a paragraph. There must be an accessible, standalone guidelines document or appendix section, an annotator interface screenshot, allowing reproduction.

### 7. iaa_metric_name -- LIST of one or more, each EXACTLY one of:
  * "Fleiss' kappa"
  * "Cohen's kappa"
  * "Krippendorff's alpha"
  * "F1 agreement"
  * "Pearson"
  * "Spearman"
  * "Majority agreement"
  * "Kendall's tau"
  * "Total agreement"     (also called percent agreement, percentage agreement, observed agreement, raw agreement, exact agreement)
  * "Other"               (any explicit metric not in the list above, e.g. ICC, Scott's pi, Gwet's AC1, BLEU, Kappa)
  * "Not reported"        (either not applicable because there is only one annotator, or no IAA metric is given in the paper)
IMPORTANT: If a metric is stated but you cannot identify which one, use "Other".
If no metric is mentioned, use "Not reported".
If multiple metrics are reported, list them in the order they appear in the paper.
[!] INTERDEPENDENCY: If annotators_per_item == "1" OR the task is open-ended free-text generation -> force ["Not reported"].

### 8. iaa_value -- LIST aligned 1-to-1 with iaa_metric_name
  - Numeric values as strings (e.g. "0.58", "85.3%", "0.09 to 0.23") for any real metric, or "NA" if the paper does not directly state the value. Refrain from inferring values.
  - If ranges are reported or any non-single-value form is used, enter as a string (e.g. "0.09 to 0.23").
  - "NA" if iaa_metric_name[i] == "Other" (regardless of whether a number is given)
  - "Not reported" if iaa_metric_name[i] == "Not reported"
  
  - Keep the same number of elements as iaa_metric_name
[!] INTERDEPENDENCY: iaa_value[i] is FULLY DETERMINED by iaa_metric_name[i] -- see Field Interdependencies rule A above.
[!] IMPORTANT: If you identified an IAA metric, ACTIVELY search for its numeric value in: (a) the main text near where the metric is mentioned, (b) ALL tables (especially agreement/statistics tables), (c) the appendix. IAA values are very often reported in tables rather than prose. Only output "NA" after checking tables.

### 9-12. Workload fields -- integer string or "Not reported"
  - **annotators_per_item**: annotators who labeled EACH item. Integer or "Not reported". Ranges like "2-3" are acceptable.
  - **total_annotators**: total number of distinct annotators. Integer or "Not reported".
  - **total_annotated_items**: total number of items/instances annotated IN THIS SPECIFIC EXPERIMENT. Integer or "Not reported". If given with K suffix (e.g. "27.9K"), keep as-is. If given with commas keep as-is (e.g. "2,864").
    [!] Extract only the item count for the annotation experiment described. An annotated item is a sample under evaluation, not the number of annotated spans or the total tokens/lines across annotated items (e.g. the number of poems, sentences, documents), NOT the total corpus/dataset size that may be larger. If the paper provides multiple numbers (e.g. a larger pre-filtered pool and a smaller annotated subset), use the smaller annotated subset count.
  - **items_per_annotator**: how many items each annotator processed. Integer or "Not reported".
  - DEFAULT: "Not reported" -- do NOT guess or calculate from other fields.

### 13. human_evaluated_sample -- free-text string
  - Describe the size/scope of the human-evaluated sample (e.g. "100", "all", "200 randomly chosen tasks from the training data", "500", "random 50")
  - If the paper describes a specific number of items evaluated by humans, report that number or description
  - report cross-annotated sample, if different to the total number of annotated items  - "Not reported" if no human-evaluated sample is described or the size is not specified
  - "all" if all items were human-evaluated, especially if the intended use of the human labour is to create a resource
  - DEFAULT: "Not reported"

### 14. recruitment -- PICK ONE exactly:
  * "Crowdsourcing"  (annotators recruited via a crowdsourcing platform or described as crowdworkers)
  * "Authors"        (one or more authors performed the annotation)
  * "Other"          (annotators are a specified non-crowd group, e.g. students, experts, hired assistants, colleagues)
  * "Mixed"          (annotators include people from different specified groups)
  * "Not reported"   (the source of annotators is not specified as a specific group)
  - DEFAULT: "Not reported"

### 15. crowd_qualifications_screening -- "Yes", "No", or "NA"
  - "Yes" if any pre-screening, qualification test, or performance record is described for crowdworkers, typically: 95% acceptance rate in previous jobs (not  socio-demographic filter)
  - "No" if crowdsourcing was used but no screening mentioned
  - "NA" if NOT a crowdsourcing study
  - DEFAULT: "NO" when not crowdsourcing; "No" when crowdsourcing 
[!] INTERDEPENDENCY (set before checking paper content):
  - recruitment == "Crowdsourcing"              -> "Yes" or "No" only
  - recruitment == "Authors" / "Other" / "Mixed" / "Not reported" -> MUST be "NO"

### 16. annotator_training -- "Yes" or "No"
  - "Yes" if training, practice rounds, trial tasks, calibration sessions, or hands-on prior experience with the annotation tool are mentioned
  - "No" if no training is described, or if there is an explicit mention of no training provided
  - DEFAULT: "No"

### 17. language_proficiency -- PICK ONE exactly:
  * "Native"          (native speakers explicitly stated; OR annotators' ethnicity is reported and it aligns with the language of the annotation task)
  * "Non-Native"      (authors report some level of proficiency, e.g. fluent, B2, C1, but not native)
  * "Mixed"           (both native and non-native speakers were employed)
  * "Medium"          (annotators have moderate language proficiency, e.g. advanced learners)
  * "Not reported"    (language proficiency not directly reported)
  - Do NOT infer from dataset language or annotator country of residence alone.
  - DEFAULT: "Not reported"

### 18. expertise_level -- PICK ONE exactly:
  * "High"         (annotators have specialised domain knowledge relevant to the task, e.g. medical professionals, legal experts, trained linguists, NLP researchers, annotating linguistic phenomena)
  * "Medium"       (annotators have some relevant background but are not deeply specialised, e.g. students in a related field, people with self-reported relevant experience)
  * "Mixed"        (reported levels of expertise vary, e.g. experts and students, or students and crowd workers used together)
  * "Not reported" (the task COULD require specialised knowledge, but the paper does not specify annotators' expertise level)
  * "General Language/Knowledge task" (the task is straightforward and DOES NOT REQUIRE any specialised knowledge -- e.g. rating text fluency/relevance/acceptability using only common language sense, evaluating whether a headline is attractive, binary preference judgments on everyday content)
  - DEFAULT: "Not reported"

### 19. reported_compensation -- PICK ONE exactly:
  * "Paid"         (any compensation is mentioned: hourly rate, per-task payment, total amount, gift cards, course credit used as incentive, being described as a paid/professional annotator or employee)
  * "Free"         (annotation was done without payment: authors annotated the data themselves, annotators were volunteers or labelled as unpaid, students completed it as a course assignment with no monetary reward)

  * "Not reported" (the paper says nothing about how or whether annotators were compensated)
  - DEFAULT: "Not reported"

### 20. payment_rate -- PICK ONE exactly:
  * "Specific numeric rate" (a concrete payment is given: hourly rate like "$12/hr", per-task like "$0.05/HIT", total amount like "paid $500 total", or any explicit monetary figure)
  * "General mention"       (compensation is confirmed but no exact amount: "annotators were paid", "workers received payment", "full-time employees", "compensated for their time")

  * "NA"                    (reported_compensation is "Free" or "Not reported" -- payment information not available)
  - DEFAULT: "NA"
  [!] Check tables and appendices for payment figures -- amounts are often reported there, not in the main text.

### 21-26. Socio-Demographics -- ALL are "Yes" or "No" ONLY
  - **age_reported**: "Yes" ONLY if actual age data (mean age, age range, age distribution) is reported WITH values. "No" otherwise.
  - **gender_reported**: "Yes" ONLY if actual gender data (counts, percentages, distribution) is reported WITH values. "No" otherwise.
  - **nation_origin_reported**: "Yes" ONLY if annotators' nationality/country of origin is reported WITH values. "No" otherwise.
  - **nation_residence_reported**: "Yes" ONLY if annotators' country of residence is reported WITH values. "No" otherwise. Indirect references to the residence of participants (e.g. poets from a local club, faculty employed at an American university department of X, English-speaking country) should be treated as Nation of residence = Yes.
  - **education_reported**: "Yes" ONLY if annotators' education level or field of study is reported WITH values. "No" otherwise.
    [!] INTERDEPENDENCY: If expertise_level == "High", set education_reported = "Yes" (explicit expert labelling implies education was assessed).
  - **political_orientation_reported**: "Yes" ONLY if political orientation is reported WITH values. "No" otherwise.
  *** CRITICAL DEFAULT: "No" for ALL demographic fields. ***
  *** "No" does NOT mean the paper says "we did not collect demographics". ***
  *** "No" simply means the paper does NOT report that specific demographic data. ***
  *** NEVER use "Not reported" for these fields -- only "Yes" or "No". ***

### 27. quality_control_filtering -- "Yes" or "No" ONLY
  - "Yes" if: quality control measures are described (related to discarding annotations or deboarding annotators: attention checks, gold or repeat questions, filtering bad annotators, removing low-quality annotations)
  - "No" in ALL other cases -- including when experts were employed and their judgments trusted without filtering, or when no QC details are given
  - *** NEVER output "NA" for this field -- "NA" is treated as "No", use "No" directly ***
  - DEFAULT: "No"

### 28. disagreement_resolution_adjudication -- PICK ONE exactly:
  * "Majority vote"      (final label determined by majority voting among annotators)
  * "Expert adjudication" (an expert resolves disagreements)
  * "Third annotator"     (a third annotator breaks ties)
  * "Consensus discussion" (annotators discuss to reach consensus)
  * "Soft labels / Distributional labels" (all annotations kept as a distribution, no single label chosen)
  * "Weighted voting"     (votes weighted by annotator reliability/expertise)
  * "Other / mixed"       (other resolution method or combination of methods)
  * "None"               (no disagreement resolution described: all opinions used as-is or averaged)
  - DEFAULT: "None"

## Extraction Rules -- READ CAREFULLY

### Experiment Identification

[!] **ANNOTATION EXPERIMENTS ONLY** -- Only extract experiments where humans actively create, label, rate, rank, or judge data. Do NOT extract model training runs, automatic evaluation steps, or system-based experiments that do not involve human annotators as a core part of the process.

**SPLIT into separate experiments when ANY of the following differ between annotation tasks:**
  1. **Different annotation objective** -- distinct label spaces or annotation goals (e.g. hate speech detection vs. stance detection)
  2. **Different annotator pools** -- different groups used for different tasks (e.g. experts for one subtask, crowd for another)
  3. **Different recruitment / compensation** -- one task paid, another unpaid; or recruited via different channels
  4. **Different annotation units** -- sentence-level vs document-level; word-level vs sentence-level
  5. **Separate dedicated annotation rounds** -- the paper describes multiple distinct rounds, each with its own guidelines, annotators, or IAA

Real-paper SPLIT examples:
    * binary attack classification / language classification / attack target classification -> 3 experiments (different label spaces)
    * summarization generation / summarization rating -> 2 experiments (generation task + evaluation task)
    * Expert Gold-Standard NER / Crowd-Based NER -> 2 experiments (different annotator pools)
    * poetry generation by crowd / poetry generation by experts -> 2 experiments (different recruitment)
    * Is label Acceptable/Unacceptable / If Acceptable, is it Precise/Imprecise -> 2 experiments (different annotation objectives)

**MERGE into one experiment when:**
  - The same annotation was applied to multiple datasets with the same label scheme and annotators
  - Multiple IAA metrics are reported for the same annotation task
  - Minor sub-tasks that share annotators, and annotated items, type of human judgment, and process differ only in topic or domain

[!] DEFAULT BIAS -- **ALWAYS err toward SPLITTING, not merging.** Under-splitting is the most common error. A paper that runs hate speech labeling and then target-type labeling on the same dataset has TWO experiments, not one. Different label spaces = different experiments.

[!] **SELF-AUDIT before writing JSON**: After identifying your first experiment, explicitly ask:
  1. Does the paper describe any OTHER annotation task (different labels, different annotators, different purpose in a different paper section)?
  2. Are there separate sections titled "Experiment 2", "Human Evaluation", "Quality Annotation", "Pilot Study"?
  3. Did any OTHER group of humans label, rate, or judge data in this paper?
  4. Is there a text production task AND a separate evaluation/rating task performed by humans?
  If the answer to ANY of these is yes -> add another experiment.

### CRITICAL: Avoid These Common Mistakes
1. **Do NOT use "Not reported" for demographic fields** -- use "No". These fields ONLY accept "Yes" or "No".
2. **disagreement_resolution_adjudication must be one of the specific methods** -- "Majority vote", "Expert adjudication", "Third annotator", "Consensus discussion", "Soft labels / Distributional labels", "Weighted voting", "Other / mixed", or "None".
3. **Do NOT infer language proficiency** from the dataset language or annotator country alone (exception: ethnicity matching task language -> "Native").
4. **Do NOT infer compensation amounts** -- but DO infer "Free" when authors annotate their own data, or mention voluntary effort.
5. **Do NOT assume guidelines are released** just because the paper describes the task in prose -- there must be a dedicated document, appendix, URL, or cited source.
6. **Do NOT calculate workload fields** from other numbers -- only extract explicitly stated values.
7. **For recruitment**, use the broad categories: "Crowdsourcing", "Authors", "Other", "Mixed", "Not reported" -- do NOT use platform-specific names.
8. **For compensation, do NOT use "NA"** -- use "Not reported" if compensation is unmentioned.
9. **Do NOT confuse expertise_level "General Language/Knowledge task" and "Not reported"**: "General Language/Knowledge task" = task needs no specialization at all; "Not reported" = task could need specialization but paper doesn't specify annotators' expertise.
10. **For IAA values, check tables** -- numeric values are very often in tables, not the prose. Do not output "NA" for a real metric until you have checked all tables.
11. **For total_annotated_items, use the annotated subset size**, not the total corpus/dataset size -- these can differ substantially.
12. **subtask_name must name the ANNOTATION task**, not the downstream NLP modelling task -- "sentiment rating" not "sentiment classifier"; "NER tagging" not "NER model".
13. **For every numerical value, record evidence** in the `numerical_evidence` object -- do NOT leave it empty when you extracted a non-"Not reported" value.

### Field Interdependencies -- APPLY THESE BEFORE FINALISING OUTPUT
These conditional rules take PRECEDENCE over all other guidance. Check each one after filling every field.

**A. iaa_value is fully determined by iaa_metric_name (apply per list position):**
  - iaa_metric_name[i] == "Other"        -> iaa_value[i] = "NA"
  - iaa_metric_name[i] == "Not reported" -> iaa_value[i] = "Not reported"
  - iaa_metric_name[i] is any real metric -> iaa_value[i] = the numeric value reported (or "NA" if value is absent)

**B. iaa_metric_name / iaa_value <- annotators_per_item or task type:**
  - annotators_per_item == "1" (single annotator)                           -> iaa_metric_name = ["Not reported"], iaa_value = ["Not reported"]
  - task is open-ended text generation (humans produce free text, no fixed label set) -> iaa_metric_name = ["Not reported"], iaa_value = ["Not reported"]

**C. crowd_qualifications_screening <- recruitment:**
  - recruitment == "Crowdsourcing"   -> crowd_qualifications_screening = "Yes" OR "No" (based on paper content)
  - recruitment == "Authors"         -> crowd_qualifications_screening = "NO"
  - recruitment == "Other"           -> crowd_qualifications_screening = "NO"
  - recruitment == "Mixed"           -> crowd_qualifications_screening = "NO" (unless paper explicitly mentions crowd screening for the crowd sub-group)
  - recruitment == "Not reported"    -> crowd_qualifications_screening = "NO"

**D. education_reported <- expertise_level:**
  - expertise_level == "High" (experts)  -> education_reported = "Yes"
    (Rationale: explicitly labelling annotators as domain experts implies their educational/professional qualification was assessed and reported.)
  - All other expertise_level values     -> apply standard "Yes"/"No" rule

### Default values summary (when information is absent from the paper):
  - guidelines_released -> "No"
  - iaa_metric_name -> ["Not reported"]
  - iaa_value -> ["Not reported"]
  - workload fields -> "Not reported"
  - recruitment -> "Not reported"
  - crowd_qualifications_screening -> "NA" (non-crowd) or "No" (crowd)
  - annotator_training -> "No"
  - language_proficiency -> "Not reported"
  - expertise_level -> "Not reported"
  - reported_compensation -> "Not reported"
  - payment_rate -> "NA"
  - human_evaluated_sample -> "Not reported"
  - ALL demographics -> "No"
  - quality_control_filtering -> "No" (NEVER "NA")
  - disagreement_resolution_adjudication -> "None"

### Numerical Evidence
For every numerical value you extract, you MUST populate the `numerical_evidence` object with the verbatim text (sentence fragment, table cell, or figure caption) from the paper that supports that number.
  - **Fields that require evidence**: `annotators_per_item`, `total_annotators`, `total_annotated_items`, `items_per_annotator`, each element of `iaa_value` (keyed as `iaa_value_0`, `iaa_value_1`, etc.), and `payment_rate` when it is "Specific numeric rate".
  - If a value was inferred rather than directly stated, write: `"field_name": "inferred from: [brief reason]"`
  - Omit a key from `numerical_evidence` only when the corresponding field value is "Not reported" or "NA".


## Approach -- Two-step extraction

**STEP 1 -- Locate annotation sections BEFORE filling any field.**
Annotation metadata is typically found in sections titled: "Annotation", "Data Collection",
"Annotators", "Human Evaluation", "Experimental Setup", "Crowdsourcing", "Dataset Construction",
"Annotation Procedure", "Annotation Scheme", or appendix sections.
[!] Only extract HUMAN ANNOTATION experiments (humans labeling/rating/judging data). Ignore model training runs, automatic evaluation steps, or any experiment that does not involve human annotators.
Scan those sections and mentally note:
  * How many DISTINCT annotation experiments? (different tasks or different annotation units = separate experiments; otherwise merge)
  * Who are the annotators -- crowd workers, authors, domain experts, students?
  * How many annotators per item? Total annotators count?
  * What IAA metric(s) and numeric value(s) are reported?
  * How were annotators recruited and compensated?
  * Any demographics WITH VALUES, training, guidelines, or quality control described?
  * For each numerical value found, note the EXACT sentence or table entry it came from (for `numerical_evidence`).
**Multi-experiment scan**: Papers with 2-3 distinct annotation experiments are COMMON. After finding your first experiment, continue scanning:
  - Other sections / appendices for additional annotation tasks
  - Whether the paper reports a "pilot" or "validation" annotation with different setup
  - Whether different tasks/label-spaces were used (e.g., first hate-speech, then target classification)
  - Whether two different annotator groups (crowd + expert) independently annotated the same or related data
  Do NOT stop at the first annotation section you find.

**STEP 2 -- Fill the experiments array** using only what you found in the paper.
When information is absent, apply the defaults in TAXONOMY_SCHEMA (typically "Not reported", "No", or "NA").

## WORKED EXAMPLE

Hypothetical paper: MTurk hate speech annotation with Fleiss' kappa, residence filter, majority-vote adjudication.

{{
  "paper_id": "example-paper",
  "skip": false,
  "experiments": [
    {{
      "paper_experiment_id": "example-paper-1",
      "papers_topic": ["Subjective language & social meaning"],
      "nlp_task_type": "Closed-label classification",
      "subtask_name": "hate speech detection",
      "intended_use": ["Resource creation"],
      "guidelines_released": "No",
      "iaa_metric_name": ["Fleiss' kappa"],
      "iaa_value": ["0.62"],
      "annotators_per_item": "3",
      "total_annotators": "50",
      "total_annotated_items": "10000",
      "items_per_annotator": "Not reported",
      "human_evaluated_sample": "Not reported",
      "recruitment": "Crowdsourcing",
      "crowd_qualifications_screening": "Yes",
      "annotator_training": "No",
      "language_proficiency": "Native",
      "expertise_level": "General Language/Knowledge task",
      "reported_compensation": "Paid",
      "payment_rate": "Specific numeric rate",
      "age_reported": "No",
      "gender_reported": "No",
      "nation_origin_reported": "No",
      "nation_residence_reported": "Yes",
      "education_reported": "No",
      "political_orientation_reported": "No",
      "quality_control_filtering": "Yes",
      "disagreement_resolution_adjudication": "Majority vote",
      "numerical_evidence": {{
        "annotators_per_item": "3/item",
        "total_annotators": "50 annotators from MTurk",
        "total_annotated_items": "collected 10,000 posts",
        "iaa_value_0": "Fleiss' kappa = 0.62"
      }}
    }}
  ]
}}

## Output Format
Output ONLY valid JSON matching this schema:
{json_schema}

## Paper Content
{paper_content}

## REMINDER before you write your JSON:
- **analysis**: write this FIRST -- 1-2 sentences: N experiments, annotator type, IAA metric+value, compensation
- Demographics (age/gender/nation_origin/
/education/political_orientation): default is "No", ONLY "Yes" if actual values WITH NUMBERS are reported
- disagreement_resolution_adjudication: use specific method name ("Majority vote", "Expert adjudication", "Third annotator", "Consensus discussion", "Soft labels / Distributional labels", "Weighted voting", "Other / mixed", "None")
- **IAA values**: check ALL tables -- values are very often in a table, not the prose
- recruitment: "Crowdsourcing" only if platform named or "crowd workers" used; "Authors" if authors themselves annotated; "Other" for students/experts/assistants; else "Not reported"
- **compensation**: authors annotating = reported_compensation "Free"; any pay mentioned = "Paid"; nothing mentioned = "Not reported"; "NA" if not applicable
- **payment_rate**: "Specific numeric rate" / "General mention" / "No" (if Free) / "NA" (if Not reported or NA) -- check tables for exact amounts
- **expertise_level**: "General Language/Knowledge task" if task needs no specialization (rating fluency/attractiveness/preference); "Not reported" if specialization COULD be relevant but paper doesn't specify
- **guidelines_released**: "Yes" if appendix with detailed instructions, URL, cited guidelines paper, or supplementary material -- NOT just because the paper describes the task
- **total_annotated_items**: use the annotated subset count, NOT the full corpus/dataset size
- **subtask_name**: use SPECIFIC wording mirroring the paper (e.g. "binary online attack classification", NOT "attack annotation"; "Pairwise Comparison of Model-Generated Summaries", NOT "evaluation")
- **papers_topic**: LIST of one or more topic areas
- **intended_use**: LIST of one or more intended uses
- **human_evaluated_sample**: free-text describing sample size/scope that was selected for human inspection or cross-annotated, or "Not reported"
- **Multiple experiments**: papers with 2-3 distinct annotation experiments are common -- re-read the FULL paper before settling on 1 experiment. Different label spaces, different annotator groups, generation + rating, pilot + main -> each is a SEPARATE experiment row
\end{lstlisting}

\onecolumn
\clearpage

\twocolumn

\twocolumn
\section{\label{appx:analysis}Additional Analysis Details}

\begin{figure}[!htbp]
     \centering
     \resizebox{\columnwidth}{!}{%
        \includegraphics{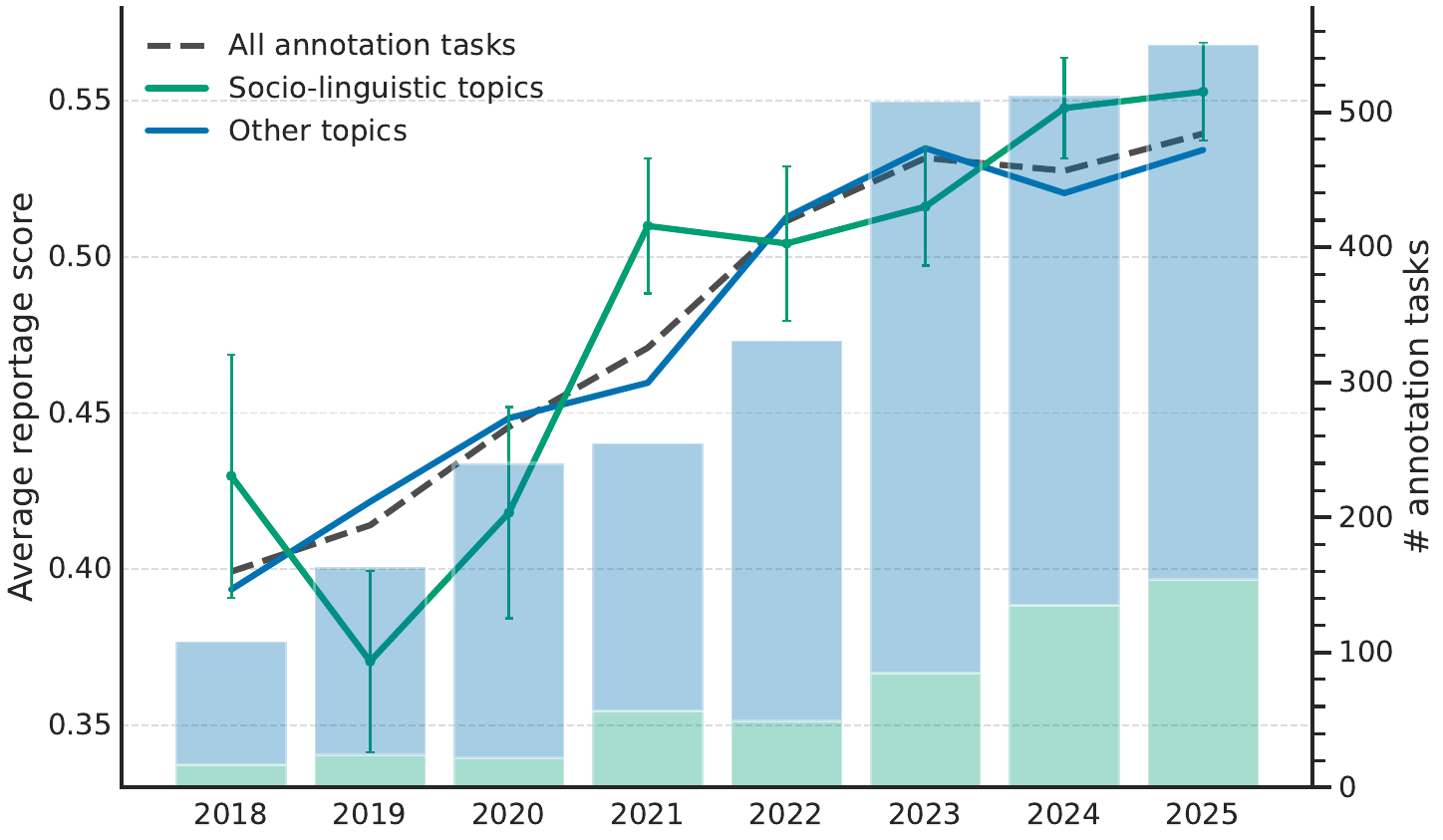}
    }
\caption{\label{fig:ss_vs_rest}Trends in reporting quality for socio-linguistic vs.\ other topics. Lines show reporting improvement over time; stacked bars indicate the number of papers per category. Error bars represent standard errors of the mean (SEM) for the focus group only. \abc{Stacked bars in the background reflect the frequency of the socially grounded papers in \datasetsub{extract}}.}
\end{figure}

\begin{figure}[ht]
    \centering
    \resizebox{0.8\columnwidth}{!}{%
        \includegraphics{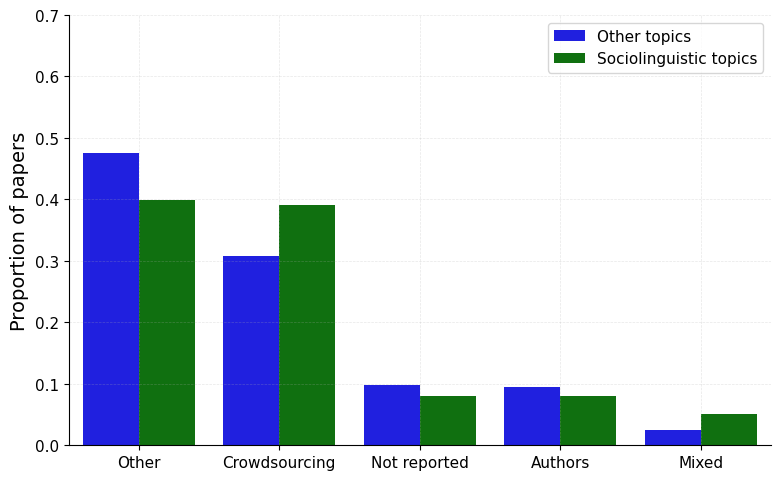}
    }
    \caption{\label{fig:recruit_ss} Distribution of recruitment categories in the socially grounded subset (green) versus the rest of the dataset (blue), showing higher use of crowdsourced and mixed annotator pools and slightly more frequent reporting of recruitment information in socially grounded papers.}
    
\end{figure}

\paragraph{RQ1: Additional analysis}
Papers on subjective and socio-linguistic phenomena do not exhibit systematically stronger reporting practices than the rest of the sample  (see Figure~\ref{fig:ss_vs_rest}).
Regarding recruitment and qualification practices, \abc{socially grounded} papers report annotator recruitment information slightly more often, suggesting more detailed documentation of annotator setup. These papers are also more likely to rely on crowdsourced and mixed annotator pools, and less likely to employ authors as annotators (see Figure~\ref{fig:recruit_ss}).

\begin{figure}[ht]
    \centering
    \resizebox{0.8\columnwidth}{!}{%
        \includegraphics{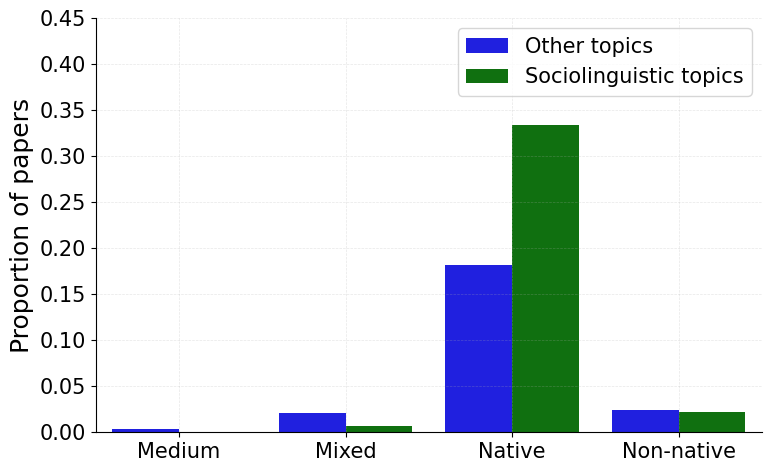}
    }
    \caption{\label{fig:nativeness}Language proficiency reporting rates, with higher reporting of nativeness for the socially grounded subset (green) compared to the rest of the dataset (blue).}
\end{figure}

A\mk{s} shown \mk{in} Figure~\ref{fig:nativeness}, with respect to language proficiency, papers on subjective language and social meaning topics report annotator nativeness more frequently than the rest of the dataset. A chi-square test indicates a statistically significant association between subset membership and native language reporting ($\chi^2 = 20.50, p < 0.001$), although the effect size is small (Cramer's $V = 0.088$), suggesting a weak association. \mk{At the same time,} native speaker status is explicitly reported in fewer than 35\% of socially grounded papers overall.

\begin{figure}[ht]
    \centering
    \resizebox{0.4\columnwidth}{!}{%
        \includegraphics{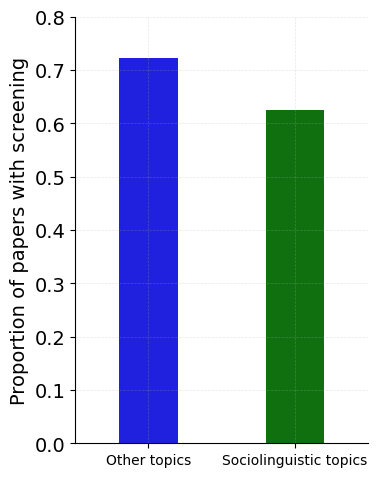}
    }
    \caption{\label{fig:crowd_screen} Distribution of quality control measures in crowd-annotated papers, defined as post-annotation filtering or validation of annotations, showing a higher rate in the socially grounded subset compared to the rest of the dataset (72\% vs. 62.5\% respectively).}
    
\end{figure}

It was observed that \abc{papers studying social phenomena} are more likely to rely on crowdsourced and mixed annotator pools. Building on this higher use of crowdsourcing, crowd-annotated papers are also shown to exhibit a higher rate of quality control measures in the social subset compared to the rest of the dataset (see Figure~\ref{fig:crowd_screen}). 

\begin{figure}[ht]
    \centering
    \resizebox{0.6\columnwidth}{!}{%
        \includegraphics{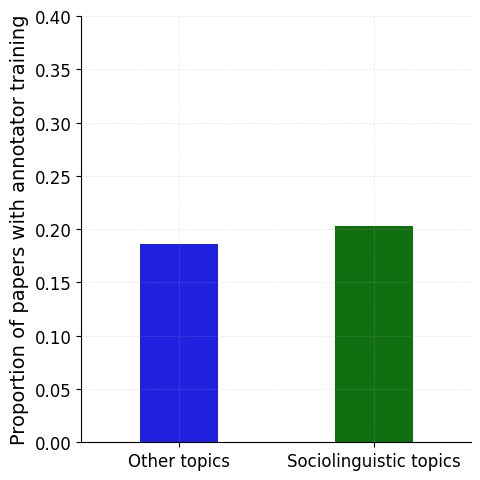}
    }
    \caption{\label{fig:annot_train} Annotator training reporting rates in \abc{socially grounded} papers versus the full dataset, showing a slightly higher frequency in the socially grounded subset (20\% vs. 18\%), with an overall difference of approximately 2\%.}
    
\end{figure}

\begin{figure}[ht]
    \centering
    \resizebox{0.8\columnwidth}{!}{%
        \includegraphics{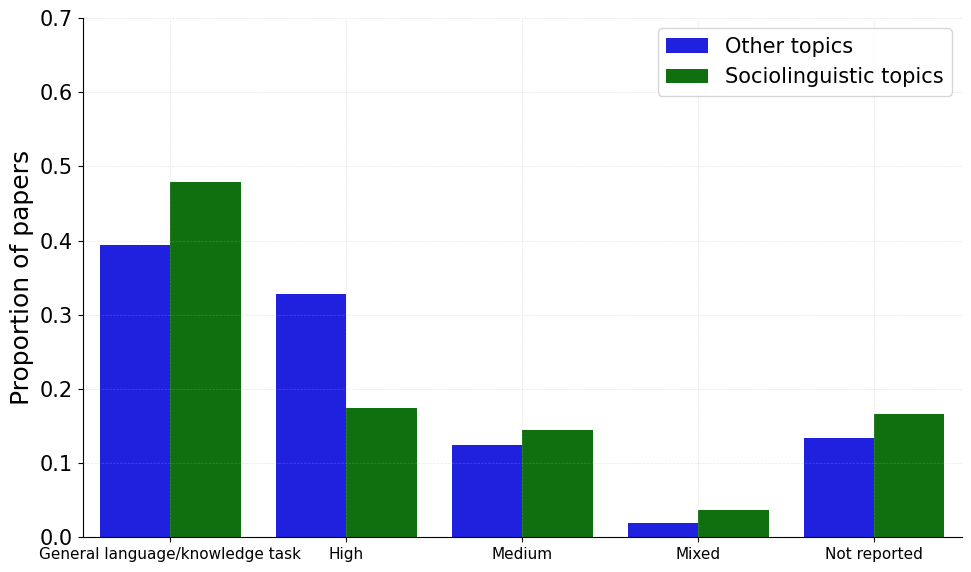}
    }
    \caption{\label{fig:annot_exp} Distribution of annotator expertise in socially grounded papers versus the rest of the dataset, showing no higher use of expert annotators in the social subset, but a higher share of general language or knowledge-based tasks and more cases where annotator expertise is not reported.}
    
\end{figure}

This \abc{includes} more frequent screening or filtering of crowdworkers in these studies. As shown in Figure~\ref{fig:annot_train}, annotator training is also slightly more frequently reported in the social subset. This difference is small but consistent, particularly when annotators belong to explicitly defined non-crowd groups (e.g., students, experts, hired assistants, or colleagues), or when multiple annotator groups are involved (see Figure~\ref{fig:annot_exp}).

\begin{figure}[ht]
    \centering
    \resizebox{0.9\columnwidth}{!}{%
        \includegraphics{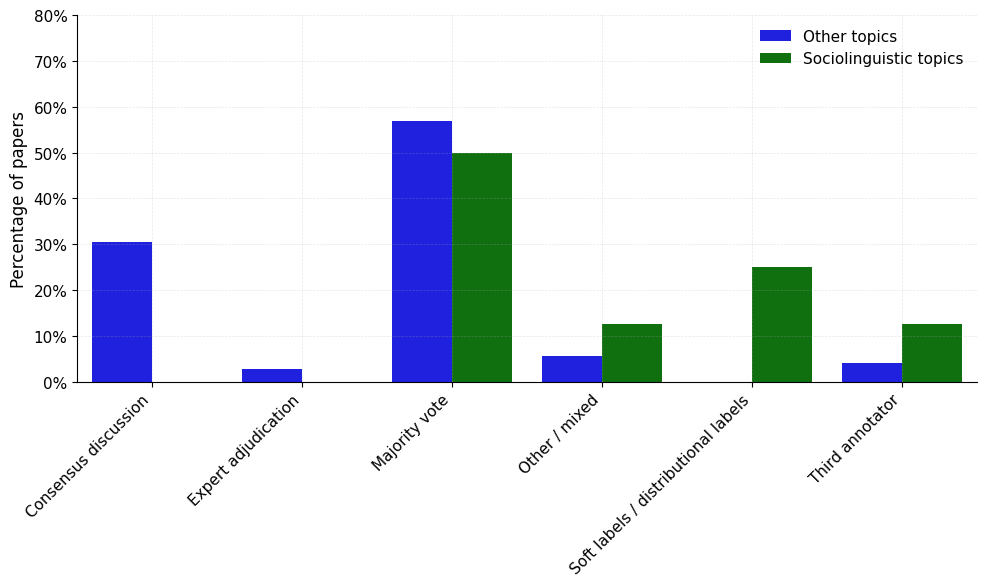}
    }
    \caption{\label{fig:disag_rating} Distribution of adjudication strategies in ordinal and scalar rating tasks, showing that socially grounded papers more often use majority voting (50.0\%) or label distributions (25.0\%) that preserve disagreement, while rarely relying on expert adjudication or consensus-based discussion.}
    
\end{figure}

Considering the category related to annotators\mk{'} expertise, no evidence is found to support the hypothesis that papers addressing socio-linguistic topics rely more on expert annotators. In fact, these studies include a higher proportion of tasks classified as general language\mk{-} or knowledge-based, and more frequently do not report annotator expertise, as illustrated in Figure~\ref{fig:annot_exp}. This contrasts with the general pattern observed for other annotator-related dimensions, where \mk{papers on socio-linguistic topics} tend to provide more detailed reporting.  A further difference emerges in how annotation disagreements are handled in ordinal or scalar rating tasks. As shown in Figure~\ref{fig:disag_rating}, within the socially grounded subset, adjudication is more commonly performed through majority voting, followed by approaches that preserve disagreement as label distributions. In contrast, methods such as expert adjudication or consensus-based discussion are rarely employed. This suggests that, in addition to more detailed reporting, socially grounded studies tend to adopt aggregation strategies that retain or simplify annotator variation rather than resolving it through expert intervention.

Overall, socially oriented NLP studies do not exhibit substantially different overall reporting quality, but they do show some distinctive tendencies in annotator profiles and documentation practices, including lower reliance on authors as annotators, greater use of crowdworkers, and more frequent reporting of native-speaker status. More broadly, these studies tend to document annotator characteristics, particularly annotator sourcing and language background, more systematically and in greater detail. \abc{However, reporting remains relatively low even within this subset, suggesting that socially grounded papers do not consistently document the broader methodological details needed to assess annotation quality, despite the importance of annotator-related variables for the phenomena they study.}

    

\paragraph{RQ2: Venue-specific interrupted time series analysis}

\abc{As a heterogeneity check, we repeated the interrupted time series analysis in Section~\ref{ssec:trends} separately for the three venues with the largest representation in our dataset: ACL, EMNLP, and NAACL. 
The interrupted time series analysis uses ordinary least squares (OLS), as implemented in the \texttt{statsmodels} library (formula: \texttt{score \textasciitilde{} time + post\_2022 + time:post\_2022}). 
The shaded bands in Figure~\ref{fig:checklist} represent 95\% confidence intervals for the mean \metric{}, computed using the conventional OLS covariance matrix and Student's $t$-distribution.}

\begin{figure}[!t]
    \centering
    \resizebox{.9\columnwidth}{!}{%
        \includegraphics{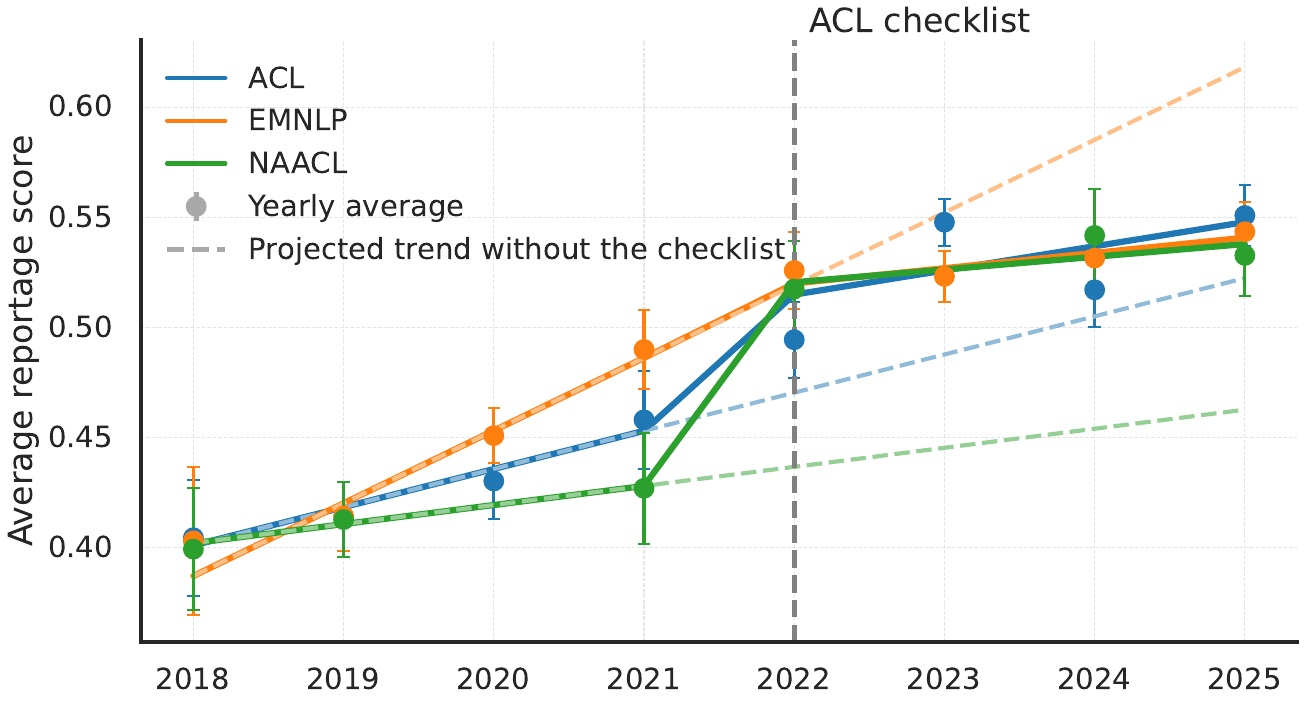}
    }
    \caption{\label{fig:venues_converge}
Interrupted time series analysis of reportage across major NLP venues (ACL, EMNLP, NAACL). Solid colored lines show model-estimated trends for each venue, while dots with whiskers indicate observed yearly means with standard errors. Dashed lines represent counterfactual trajectories under continuation of the pre-2022 trend.
}
\end{figure}

Figure~\ref{fig:venues_converge} shows venue-specific fitted trends, observed yearly reportage scores, and counterfactual continuations of the pre-2022 trajectories.

The three venues exhibit distinct pre-2022 reporting trajectories. EMNLP shows comparatively high \metric{} values throughout the observation period, with no pronounced discontinuity around 2022 beyond the existing upward trend. In contrast, ACL and especially NAACL begin from substantially lower baseline levels and display stronger upward trends prior to 2022, gradually converging toward the EMNLP level.

Following 2022, the previously observed upward trajectories flatten across all three venues. Rather than accelerating post-intervention improvements, reportage scores stabilize at a more similar level across venues, reducing between-venue variation. This pattern suggests that the ACL Responsible NLP Checklist may have had a standardizing effect on reporting practices, even though we do not observe evidence for a strong immediate increase in reportage quality after its introduction.

Because the checklist was introduced during a broader transition period involving ARR adoption and venue-specific rollout schedules, the intervention year should be interpreted as an approximate temporal reference point rather than a perfectly synchronized policy change across all venues.

\begin{figure}[ht]
    \centering
    
    \begin{subfigure}[t]{0.48\columnwidth}
        \centering
        \includegraphics[width=\linewidth]{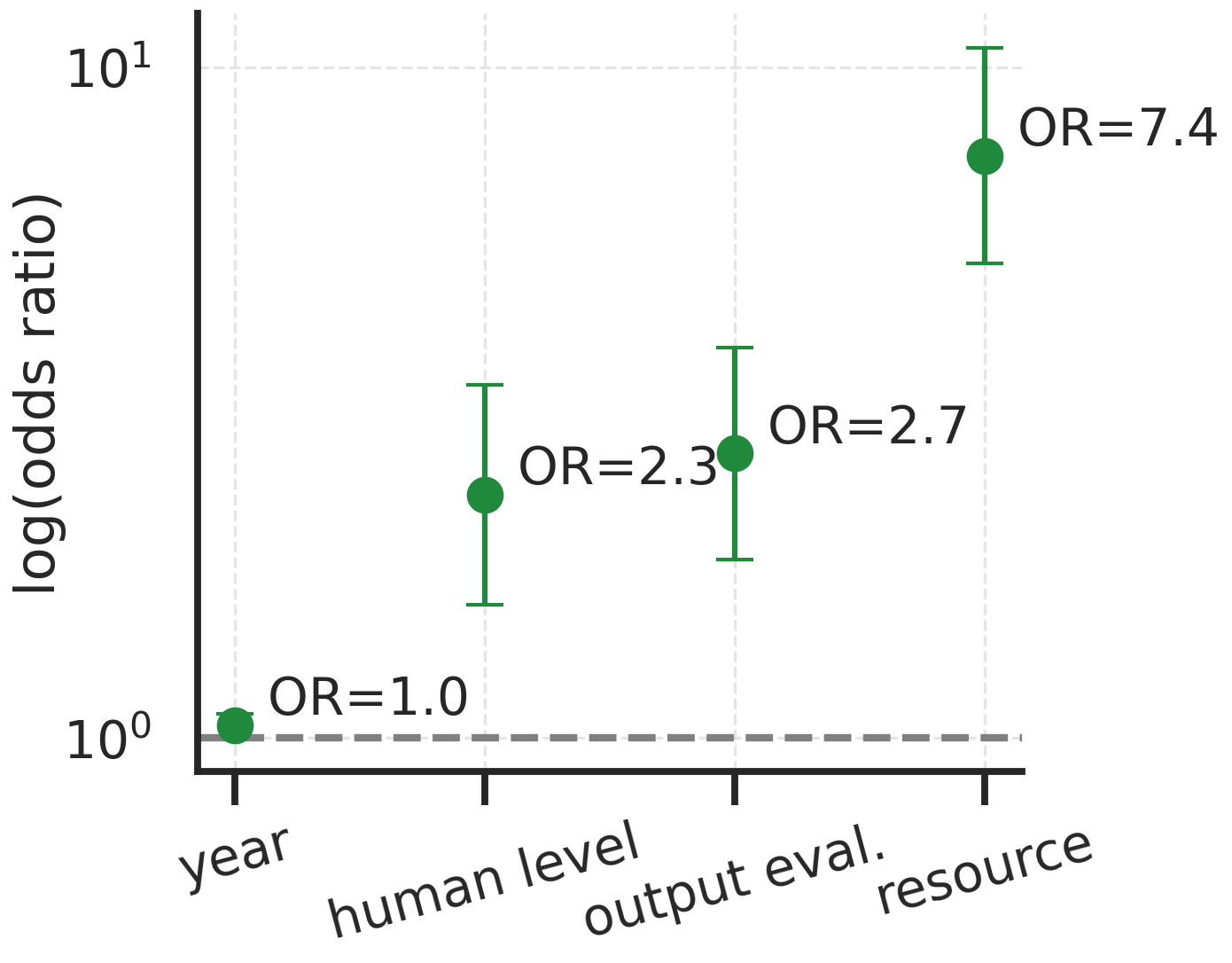}
        \caption{\label{fig:qc}{\small QC measures combined}}
    \end{subfigure}
    \hfill
    \begin{subfigure}[t]{0.48\columnwidth}
        \centering
        \includegraphics[width=\linewidth]{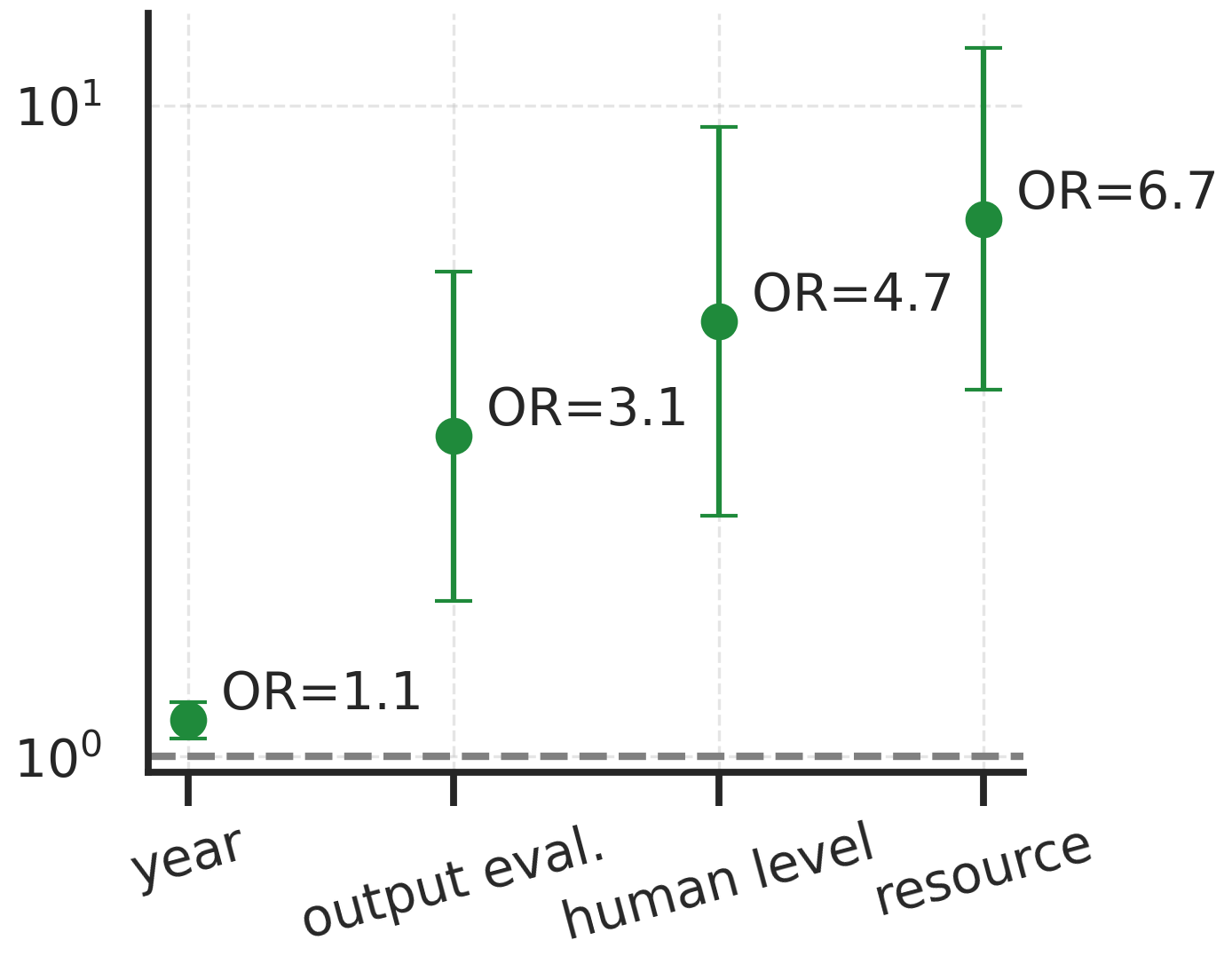}
        \caption{\label{fig:recruit}{\small recruitment+comp.}}
    \end{subfigure}

    \caption{\label{fig:qc_by_use}
    Logistic regression of quality-control (QC) usage by annotation purpose and year.
    Points show odds ratios (ORs) with 95\% CIs on a log scale.
    \abc{The dashed horizontal line at OR = 1 indicates no association.
    (a) QC usage indicates the presence of at least one of the following:
    adjudication, post-filtering, annotator training, or an IAA metric;
    (b) recruitment source and compensation are analyzed separately.}}
\end{figure}

\paragraph{RQ3: Additional analysis} 
\abc{Figure~\ref{fig:qc_by_use} visualizes the logistic regression estimates discussed in the main text. We fitted binary logistic regression models using the \texttt{statsmodels} library to predict the reporting of two sets of quality-related attributes. For panel (a), the binary outcome indicates whether a task reports at least one of the following: adjudication, post-annotation filtering, annotator training, or an IAA metric. Panel (b) analogously considers recruitment source and compensation. The predictors indicate whether the task is intended for resource creation, model-output evaluation, or benchmarking human performance, with publication year included as a control (formula: \texttt{has\_qc \textasciitilde{} is\_resource + is\_model + is\_human + year\_num}). Parameters were estimated by maximum likelihood using the Newton--Raphson optimizer, and inference was based on two-sided Wald tests.}

\abc{In both subplots, annotation tasks aimed at resource creation (resource) show the strongest positive association with reporting procedural details, with odds ratios substantially above 1 (dashed line) across all measures considered. Tasks aimed at benchmarking human performance (human level) occupy an intermediate position, whereas model-output evaluation tasks (output eval.) consistently show the weakest reporting patterns.}

The similarity between the two panels suggests that these differences are not limited to formal quality control procedures alone (a), but extend more broadly to annotator transparency and documentation practices (b) -- who are the annotators? Notably, the comparatively weak reporting of recruitment source and compensation in model evaluation studies raises the possibility that authors acting as annotators may remain systematically underreported in this setting. Temporal effects are comparatively modest in both models, indicating that annotation purpose is a substantially stronger predictor of reporting behavior than publication year.



\paragraph{Sample Annotated Data}
The following figures present excerpts from \citet{haber-etal-2023-improving}, included as an example of a manually annotated paper with comprehensive reporting practices. The excerpts illustrate reporting of annotator recruitment, demographic and linguistic characteristics, workload distribution, inter-annotator agreement (IAA), and disagreement resolution procedures.

\begin{figure}[htbp]
    \centering

    \includegraphics[width=0.9\columnwidth]{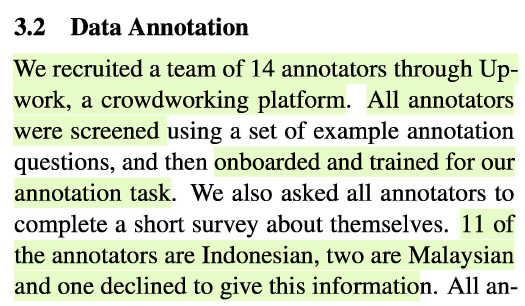}

    \caption{
    Example excerpt from \citet{haber-etal-2023-improving} illustrating comprehensive reporting of annotator-related information, including the number of annotators, recruitment procedures, and annotators' nation of origin.
    }
    \label{fig:paper_exc1}

    \vspace{2mm}
    
\includegraphics[width=0.9\columnwidth]{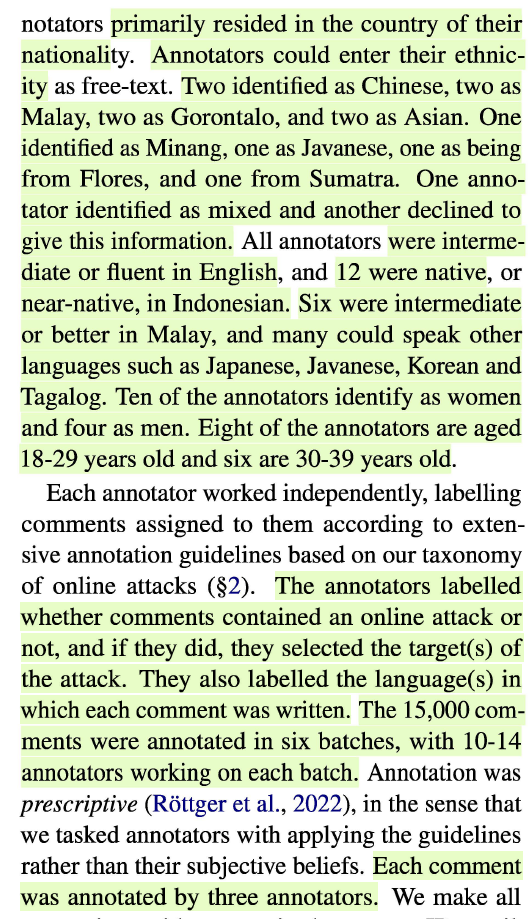}

\caption{
Example excerpt from \citet{haber-etal-2023-improving} providing extensive reporting of annotation methodology and annotator characteristics, including annotators' nation of origin, language proficiency, gender and age, as well as total number of annotated items and number of annotators per item.
}
\label{fig:paper_exc2}

    \vspace{2mm}
    
\includegraphics[width=0.9\columnwidth]{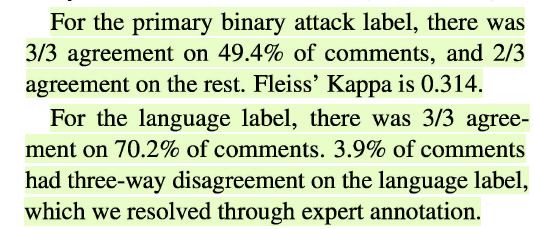}

\caption{
Example excerpt from \citet{haber-etal-2023-improving} providing detailed reporting of inter-annotator agreement (IAA) metrics, scores, and disagreement resolution procedures.
}
\label{fig:paper_exc3}

\end{figure}

\onecolumn
\clearpage

\begin{xltabular}{\textwidth}{@{}p{2cm} X rrrrr@{}}

\toprule
category                                  & value                                 & 2018-2021 \# & 2022-2025 \# & 2018-2021 \% & 2022-2025 \% & p\textless{}0.05 \\
\midrule
\endfirsthead

category                                  & value                                 & 2018-2021 \# & 2022-2025 \# & 2018-2021 \% & 2022-2025 \% & p\textless{}0.05 \\ 
\midrule
\endhead

\multirow{10}{*}{paper\_topic}          & open-ended generation                 & 349          & 805          & 28.4         & 25.1         & *                \\
                                        & semantics, inference \& similarity    & 232          & 453          & 18.9         & 14.1         & *                \\
                                        & subjective language \& social meaning & 120          & 423          & 9.8          & 13.2         & *                \\
                                        & discourse \& pragmatics               & 116          & 242          & 9.4          & 7.6          & *                \\
                                        & qa \& information access              & 108          & 265          & 8.8          & 8.3          &                  \\
                                        & information extraction                & 103          & 302          & 8.4          & 9.4          &                  \\
                                        & compression \& reformulation tasks    & 95           & 201          & 7.7          & 6.3          &                  \\
                                        & multimodal tasks                      & 44           & 247          & 3.6          & 7.7          & *                \\
                                        & multilingual \& translation           & 39           & 174          & 3.2          & 5.4          & *                \\
                                        & linguistics                           & 24           & 93           & 2.0          & 2.9          & NA               \\
\midrule
\multirow{6}{*}{task\_type}             & closed-label clf                      & 253          & 632          & 33.0         & 33.2         &                  \\
                                        & ordinal/scalar rating                 & 229          & 551          & 29.9         & 29.0         &                  \\
                                        & open-ended generation                 & 103          & 200          & 13.4         & 10.5         & *                \\
                                        & comparative/preference                & 91           & 262          & 11.9         & 13.8         &                  \\
                                        & span-level/structural                 & 68           & 156          & 8.9          & 8.2          &                  \\
                                        & other                                 & 22           & 100          & 2.9          & 5.3          & NA               \\
\midrule
\multirow{4}{*}{intended\_use}          & model output evaluation               & 380          & 1043         & 47.1         & 50.5         &                  \\
                                        & resource creation                     & 329          & 773          & 40.8         & 37.5         &                  \\
                                        & human performance                     & 86           & 219          & 10.7         & 10.6         &                  \\
                                        & none of the above                     & 11           & 29           & 1.4          & 1.4          & NA               \\
\midrule
\multirow{2}{*}{guidelines}   & no                                    & 612          & 1145         & 79.9         & 60.2         & *                \\
                                        & yes                                   & 154          & 756          & 20.1         & 39.8         & *                \\
\midrule
\multirow{11}{*}{iaa\_metric}           & not reported                          & 502          & 1103         & 55.7         & 48.3         & *                \\
                                        & fleiss' $\kappa$                             & 116          & 272          & 12.9         & 11.9         &                  \\
                                        & cohen's $\kappa$                             & 79           & 268          & 8.8          & 11.7         & *                \\
                                        & krippendorff's $\alpha$                      & 68           & 289          & 7.5          & 12.7         & *                \\
                                        & other                                 & 42           & 140          & 4.7          & 6.1          &                  \\
                                        & total agreement                       & 37           & 124          & 4.1          & 5.4          &                  \\
                                        & spearman                              & 21           & 23           & 2.3          & 1.0          & NA               \\
                                        & f1 agreement                          & 15           & 24           & 1.7          & 1.1          & NA               \\
                                        & pearson                               & 13           & 19           & 1.4          & 0.8          & NA               \\
                                        & kendall's $\tau$                         & 5            & 19           & 0.6          & 0.8          & NA               \\
                                        & majority agreement                    & 3            & 3            & 0.3          & 0.1          & NA               \\
\midrule
\multirow{2}{*}{iaa\_value}             & reported                              & 765          & 1894         & 99.9         & 99.6         &                  \\
                                        & not\_reported                         & 1            & 7            & 0.1          & 0.4          & NA               \\
\midrule
\multirow{2}{*}{annotators/item}  & reported                              & 528          & 1201         & 68.9         & 63.2         & *                \\
                                        & not\_reported                         & 238          & 700          & 31.1         & 36.8         & *                \\
\midrule
\multirow{2}{*}{total\_annotators}      & not\_reported                         & 387          & 601          & 50.5         & 31.6         & *                \\
                                        & reported                              & 379          & 1300         & 49.5         & 68.4         & *                \\
\midrule
\multirow{2}{*}{total\_items}           & reported                              & 673          & 1621         & 87.9         & 85.3         &                  \\
                                        & not\_reported                         & 93           & 280          & 12.1         & 14.7         &                  \\
\midrule
\multirow{2}{*}{items/annotator}  & not\_reported                         & 640          & 1421         & 83.6         & 74.8         & *                \\
                                        & reported                              & 126          & 480          & 16.4         & 25.2         & *                \\
\midrule
\multirow{5}{*}{recruitment}            & crowdsourcing                         & 358          & 476          & 46.7         & 25.0         & *                \\
                                        & other                                 & 247          & 1009         & 32.2         & 53.1         & *                \\
                                        & not reported                          & 100          & 157          & 13.1         & 8.3          & *                \\
                                        & authors                               & 51           & 200          & 6.7          & 10.5         & *                \\
                                        & mixed                                 & 10           & 59           & 1.3          & 3.1          & NA               \\
\midrule
\multirow{3}{*}{crowd\_screening}       & no                                    & 492          & 1244         & 64.2         & 65.4         &                  \\
                                        & yes                                   & 186          & 341          & 24.3         & 17.9         & *                \\
                                        & na                                    & 88           & 316          & 11.5         & 16.6         & *                \\
\midrule
\multirow{2}{*}{training}               & no                                    & 667          & 1502         & 87.1         & 79.0         & *                \\
                                        & yes                                   & 99           & 399          & 12.9         & 21.0         & *                \\
\midrule
\multirow{5}{*}{lang.\_proficiency}  & not reported                          & 631          & 1403         & 82.4         & 73.8         & *                \\
                                        & native                                & 116          & 390          & 15.1         & 20.5         & *                \\
                                        & non-native                            & 11           & 54           & 1.4          & 2.8          & NA               \\
                                        & mixed                                 & 5            & 48           & 0.7          & 2.5          & NA               \\
                                        & medium                                & 3            & 6            & 0.4          & 0.3          & NA               \\
\midrule
\multirow{5}{*}{expertise\_level}       & general task                          & 448          & 613          & 58.5         & 32.2         & *                \\
                                        & high                                  & 147          & 708          & 19.2         & 37.2         & *                \\
                                        & not reported                          & 108          & 253          & 14.1         & 13.3         &                  \\
                                        & medium                                & 56           & 280          & 7.3          & 14.7         & *                \\
                                        & mixed                                 & 7            & 47           & 0.9          & 2.5          & NA               \\
\midrule
\multirow{3}{*}{compensation}           & not reported                          & 479          & 697          & 62.5         & 36.7         & *                \\
                                        & paid                                  & 218          & 913          & 28.5         & 48.0         & *                \\
                                        & free                                  & 69           & 291          & 9.0          & 15.3         & *                \\
\midrule
\multirow{3}{*}{payment\_rate}          & na                                    & 548          & 988          & 71.5         & 52.0         & *                \\
                                        & specific numeric rate                 & 139          & 605          & 18.1         & 31.8         & *                \\
                                        & general mention                       & 79           & 308          & 10.3         & 16.2         & *                \\
\midrule
\multirow{2}{*}{age}                    & no                                    & 760          & 1763         & 99.2         & 92.7         & *                \\
                                        & yes                                   & 6            & 138          & 0.8          & 7.3          & NA               \\
\midrule
\multirow{2}{*}{gender}                 & no                                    & 758          & 1744         & 99.0         & 91.7         & *                \\
                                        & yes                                   & 8            & 157          & 1.0          & 8.3          & NA               \\
\midrule
\multirow{2}{*}{nation\_origin}         & no                                    & 762          & 1845         & 99.5         & 97.1         & *                \\
                                        & yes                                   & 4            & 56           & 0.5          & 2.9          & NA               \\
\midrule
\multirow{2}{*}{nation\_residence}      & no                                    & 675          & 1621         & 88.1         & 85.3         &                  \\
                                        & yes                                   & 91           & 280          & 11.9         & 14.7         &                  \\
\midrule
\multirow{2}{*}{education}              & no                                    & 596          & 1007         & 77.8         & 53.0         & *                \\
                                        & yes                                   & 170          & 894          & 22.2         & 47.0         & *                \\
\midrule
\multirow{2}{*}{political\_orient.} & no                                    & 765          & 1887         & 99.9         & 99.3         &                  \\
                                        & yes                                   & 1            & 14           & 0.1          & 0.7          & NA               \\
\midrule
\multirow{2}{*}{post\_filtering}        & no                                    & 551          & 1447         & 71.9         & 76.1         & *                \\
                                        & yes                                   & 215          & 454          & 28.1         & 23.9         & *                \\
\midrule
\multirow{8}{*}{adjudication}           & none                                  & 593          & 1437         & 77.4         & 75.6         &                  \\
                                        & majority vote                         & 108          & 228          & 14.1         & 12.0         &                  \\
                                        & consensus discussion                  & 23           & 105          & 3.0          & 5.5          & NA               \\
                                        & expert adjudication                   & 14           & 42           & 1.8          & 2.2          & NA               \\
                                        & other/mixed                           & 14           & 49           & 1.8          & 2.6          & NA               \\
                                        & third annotator                       & 8            & 27           & 1.0          & 1.4          & NA               \\
                                        & soft labels                           & 3            & 11           & 0.4          & 0.6          & NA               \\
                                        & weighted voting                       & 3            & 2            & 0.4          & 0.1          & NA               \\ \bottomrule 

\caption{\label{tab:binned}
Impact of the ACL Responsible NLP Checklist. Distributions are shown for 2018–2021 vs.\ 2022–2025. Multilabel fields are expanded into individual categories; free-text fields are binarized (reported vs.\ not reported). Statistical significance (*) reflects differences in proportions between periods, assessed using a $\chi^2$ test ($p<0.05$); results are omitted (NA) when counts fall below 30 in either period. Total observations (tasks) : 2,669.}

\end{xltabular}


\end{document}